\documentclass[11pt]{article}

\usepackage[preprint]{acl}

\usepackage{times}
\usepackage{latexsym}

\usepackage[T1]{fontenc}
\usepackage[utf8]{inputenc}

\usepackage{microtype}

\usepackage{inconsolata}

\usepackage{graphicx}
\usepackage{mathtools}
\usepackage{CJKutf8}
\usepackage[table]{xcolor}
\usepackage{tabularx}
\usepackage{listings}
\usepackage{multirow} 
\usepackage{multicol} 
\usepackage{url}
\usepackage{subcaption}
\usepackage{booktabs} % for professional tables

\usepackage{hyperref}
\usepackage{amsmath, amssymb, amsthm}
\usepackage{float}
\usepackage{caption}
\usepackage{placeins}
\usepackage{setspace}
\usepackage[ruled,vlined,linesnumbered]{algorithm2e}
\usepackage{pgfplots}
\usepackage{xcolor}
\pgfplotsset{compat=1.18}
\usepgfplotslibrary{groupplots}

\title{ChampionKV: Head-Wise Adaptive Budget Allocation for KV Cache Compression}

\author{%
  Yuechuan Chen\textsuperscript{1},
  Yanda Guo\textsuperscript{1},
  Xiaoxi Liu\textsuperscript{1},
  Weilin Ruan\textsuperscript{2},
  Huanbo Xu\textsuperscript{3},
  Ruyi Wang\textsuperscript{4},
  Wei Zhu\textsuperscript{5}\\
  \textsuperscript{1}Sun Yat-sen University\\
  \textsuperscript{2}The Chinese University of Hong Kong\\
  \textsuperscript{3}University of Electronic Science and Technology of China\\
  \textsuperscript{4}Tongji University \\
  \textsuperscript{5}Hong Kong University \\
  \href{mailto:chenych227@mail2.sysu.edu.cn}{chenych227@mail2.sysu.edu.cn} \\
}

\begin{document}
\maketitle

\begin{abstract}
Some observation-window-based KV cache compressors use prompt attention to identify important tokens but allocate uniform or predefined budgets across heads, overlooking input-dependent differences in budget demand. We propose \textsc{ChampionKV}, an input-adaptive head-wise budget allocator for long-context KV cache compression. It uses the maximum aggregated observation-window attention score as a lightweight empirical proxy for attention concentration, assigning larger budgets to heads with dispersed attention and smaller budgets to concentrated heads. An abnormal-head mask and a floor-budget mechanism improve robustness under weak attention responses. \textsc{ChampionKV} plugs into compressors such as SnapKV and PyramidKV without changing their token-selection procedures. On LongBench and RULER, both Champion variants consistently improve over their SnapKV and PyramidKV base compressors; the improvements are largest and statistically significant on Llama3.1-70B-Instruct, especially under tight budgets, whereas results on the tested 7B/8B models remain competitive but mixed. \textsc{ChampionKV} does not uniformly outperform Ada-KV, whose global cross-head Top-$K$ selection retains an advantage in several settings. Controlled analysis further shows improved retained attention mass over the corresponding base compressors at negligible overhead. Code is available at \url{https://github.com/YuechuanChen/Champion_KV}

\end{abstract}

% \keywords{Large Language Models, Multi-Agent Systems, Dynamic Collaboration, Communication Graph Learning, Reinforcement Learning, Policy Optimization, Query-Adaptive Topology, Structured Reasoning}

\section{Introduction}

Large language models (LLMs) are increasingly used in long-context scenarios
such as long-document question answering, code completion, multi-turn dialogue,
and retrieval-augmented generation
~\citep{team2024gemini,guo2025deepseek,yang2025qwen3}.
During autoregressive decoding, Transformer decoders store historical key and
value states in a key-value (KV) cache. Because the cache grows linearly with
context length, batch size, model depth, and KV-head dimensionality, it becomes
a major memory bottleneck for long-context inference
~\citep{zhang2023h2o,li2024snapkv}. KV cache compression addresses this problem
by retaining only the KV entries most useful for future generation.

Most KV cache compression methods identify important tokens using attention
scores. H2O retains heavy-hitter and recent tokens
~\citep{zhang2023h2o}, while SnapKV selects historical tokens using attention
from an observation window near the end of the prompt
~\citep{li2024snapkv}. Recent work also shows that cache requirements vary
across model components: PyramidKV allocates non-uniform budgets across layers
~\citep{cai2024pyramidkv}, Ada-KV performs online global cross-head Top-$K$
selection~\citep{feng2026ada}, and HeadKV uses offline head-importance
estimates~\citep{fu2025not}. We therefore investigate whether a lightweight
per-head statistic available during prefilling can support input-adaptive
budget allocation without changing the underlying token-selection procedure.

\begin{figure*}[t]
    \centering
    \includegraphics[width=1\textwidth]{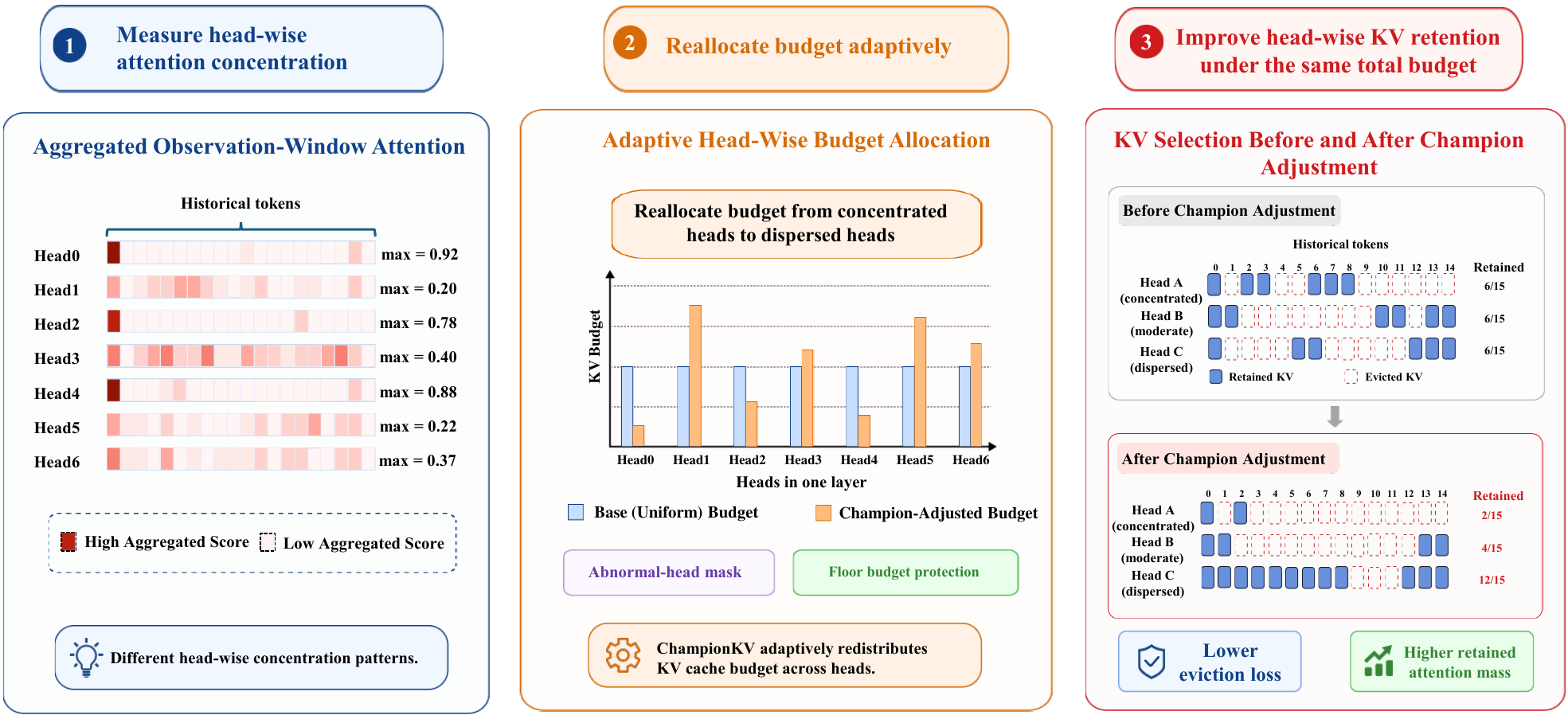}
    \caption{
    Overview of \textsc{ChampionKV}. Given observation-window attention, \textsc{ChampionKV} estimates head-wise attention concentration using the maximum aggregated attention score and adaptively reallocates KV cache budget across heads. Dispersed heads receive larger budgets, while concentrated heads receive smaller budgets, improving retained attention mass without modifying the underlying token-selection procedure.
    }
    \label{fig:championkv_overview}
\end{figure*}

We propose \textsc{ChampionKV}, an input-adaptive head-wise budget allocator
for observation-window-based KV cache compression. As shown in
Figure~\ref{fig:championkv_overview}, it uses the maximum aggregated
observation-window attention score as a lightweight empirical proxy for
attention concentration. Heads with dispersed attention receive larger
budgets, whereas concentrated heads are compressed more aggressively.
An abnormal-head mask and a floor-budget mechanism improve robustness under
weak or noisy attention responses. \textsc{ChampionKV} can be combined with
compressors such as SnapKV and PyramidKV without modifying their
token-selection procedures.

We evaluate \textsc{ChampionKV} on LongBench and RULER using
Llama3.1-70B-Instruct, Llama3.1-8B-Instruct, and
Mistral-7B-Instruct-v0.3. Both Champion variants consistently outperform
their SnapKV and PyramidKV base compressors, with statistically significant
gains on the 70B model under tight budgets; on the tested 7B/8B models the
results are competitive but mixed, and neither variant uniformly surpasses
Ada-KV. We further examine retained attention mass, hyper-parameter
sensitivity, and hardware efficiency.

Our contributions are threefold: (a) a lightweight attention-concentration
proxy for ranking head-wise cache demand; (b) an inverse-max allocation rule
with abnormal-head masking and floor-budget protection; and (c) an evaluation
of its downstream performance, retained attention mass, and hardware
efficiency under constrained cache budgets.

\section{Related Work}

\subsection{Taxonomy of KV Cache Compression}
KV cache compression aims to reduce the memory footprint of autoregressive LLM inference by retaining only the KV entries that are most useful for future generation. Existing methods can be broadly organized into three categories. First, observation-window-based methods use attention scores from recent prompt tokens to estimate the importance of historical tokens before decoding. For example,~\citet{li2024snapkv} propose SnapKV, which aggregates attention from an observation window at the end of the prompt and performs per-head Top-$K$ token selection. PyramidInfer also exploits recent-token attention consistency to identify pivotal context in advance~\citep{yang2024pyramidinfer}. Second, layer-wise methods study non-uniform cache requirements across Transformer layers.~\citet{cai2024pyramidkv} observe a pyramidal information funneling phenomenon, where lower layers tend to require larger budgets to aggregate global information, while higher layers can often operate with smaller budgets due to more concentrated attention. Third, head-wise methods further exploit heterogeneity among attention heads, assigning different compression strategies or cache budgets to different heads. This category is most closely related to our work. Overall, the methods mainly improve token selection, layer-wise methods improve inter-layer budget allocation, and head-wise methods improve intra-layer budget allocation. 

\subsection{Head-Wise KV Cache Compression}
Head-wise KV cache compression is motivated by the broader observation that attention heads are heterogeneous in both importance and function.~\citet{michel2019sixteen} show that many heads can be removed at inference time with limited performance degradation, suggesting unequal contributions across heads. Mechanistic studies further identify specialized head types, such as induction heads related to pattern completion and in-context learning~\citep{olsson2022context}, while long-context analyses show that a small number of retrieval heads can be crucial for factual retrieval from long inputs~\citep{wu2025retrieval}; see also the systematic discussion by~\citet{zheng2024attention}. In KV cache compression, FastGen~\citep{ge2024model} profiles attention heads and assigns different cache strategies to structural patterns such as local, special-token, punctuation, frequency-based, and broad attention heads. However, it mainly selects among discrete compression strategies rather than continuously allocating budgets among heads. Ada-KV~\citep{feng2026ada} formulates KV eviction from the perspective of retained attention mass and uses input-dependent attention scores for global cross-head Top-$K$ selection within each layer, enabling online adaptive budget allocation. HeadKV~\citep{fu2025not} estimates head importance using offline retrieval-reasoning examples and allocates KV cache budgets according to these precomputed importance scores.

\textsc{ChampionKV} follows the observation-window token-selection paradigm and can be combined with layer-wise allocation. Like Ada-KV, it adapts budgets to the current input during prefilling, but uses a scalar concentration proxy and inverse-max weighting instead of global cross-head token ranking. Unlike HeadKV, it does not require offline head-importance estimates. We next examine whether the maximum aggregated observation-window attention score can serve as this proxy.

\section{Pilot Experiment}

\subsection{Head-Wise Distribution Heterogeneity}
\label{subsec:Head-Wise_Distribution_Heterogeneity}

Following SnapKV~\cite{li2024snapkv}, we use the last $W$ tokens of the input sequence as an observation window to estimate which historical tokens are likely to be important for subsequent generation. 

For the $h$-th head in the $l$-th layer, the observation-window attention is computed as:

\[
\begin{aligned}
{A}^{l,h}
&=\operatorname{softmax}\!\left(
\frac{
Q^{l,h}_{\mathrm{obs}}
\left(K^{l,h}_{\mathrm{all}}\right)^\top
}{
\sqrt{d_{\mathrm{key}}}
}
\right)\\
&\qquad {A}^{l,h}\in\mathbb{R}^{W\times L}
\end{aligned}
\]

Here, \(Q^{l,h}_{\mathrm{obs}}\) denotes the query states of the last \(W\) tokens, and \(K^{l,h}_{\mathrm{all}}\) denotes the key states of the whole prompt. We then keep the historical-token part of the attention matrix,
\(A_{\mathrm{hist}}^{l,h}\in\mathbb{R}^{W\times (L-W)}\),
by excluding the last \(W\) observation-window positions.
The aggregated observation-window attention score is:
\[
a_i^{l,h}
=
\frac{1}{W}
\sum_{t=0}^{W-1} A_{\mathrm{hist},t,i}^{l,h},
\quad
i=0,\ldots,L-W-1.
\]

This vector, termed the \textbf{\textit{aggregated observation-window attention score}}, is used to estimate the importance of historical KV entries.

\begin{figure}[ht!]
    \centering
    \includegraphics[width=1\linewidth]{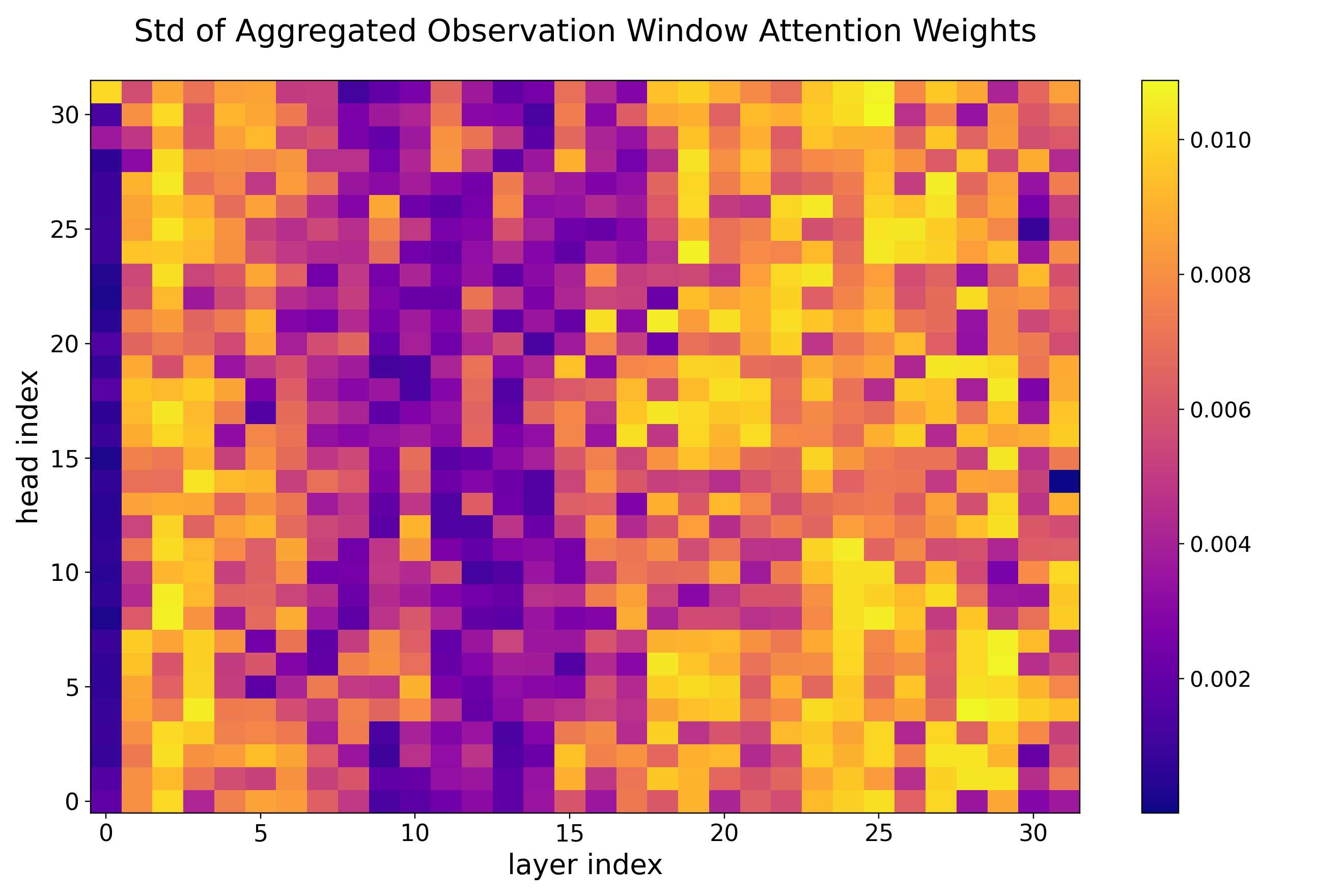}
    \caption{Standard deviation of aggregated observation-window attention scores across layers and heads in Llama3.1-8B-Instruct on NarrativeQA. The layer--head variation suggests heterogeneous attention concentration.}
    \label{fig:std}
\end{figure}

Prior studies on attention heads show that different heads may play different functional roles~\citep{michel2019sixteen,olsson2022context,wu2025retrieval,zheng2024attention}. We further observe that, under observation-window-based KV compression, different heads also exhibit different attention distribution shapes. As shown in Figure~\ref{fig:std}, the standard deviation of aggregated observation-window attention scores varies clearly across layers and heads, indicating heterogeneous attention concentration.

Specifically, we identify two typical distribution patterns. In a \textbf{\textit{concentrated head}}, most attention mass is assigned to only a few historical tokens, such as \texttt{<bos>}~\cite{xiao2024efficient} or task-relevant evidence tokens. Such a head can cover its major attention mass with a small number of KV entries. In contrast, in a \textbf{\textit{dispersed head}}, attention scores are spread over many historical tokens, making the attention mass harder to cover under a small cache budget. This motivates assigning larger KV cache budgets to dispersed heads than to concentrated heads. Detailed visual examples are provided in Appendix~\ref{app:Head-Wise_Distribution_Heterogeneity}.

\subsection{Motivation for Champion-Based Budget Allocation}
\label{subsec:champion_motivation}

Ada-KV~\cite{feng2026ada} analyzes KV eviction through attention-output perturbation and gives the following upper bound for the eviction loss within one layer:

\[
\mathcal{L}_{Eviction}
\leq
2N_hC
-
2C
\sum_{h=0}^{N_h-1}
\sum_{i=0}^{L-W-1}
I_i^{l,h} a_i^{l,h}
\]

where \(N_h\) is the number of attention heads, \(I_i^{l,h}\in\{0,1\}\) indicates whether the \(i\)-th historical KV entry in head \(h\) is retained, and \(C\) is a norm-related constant determined by value states and output projections. Since \(N_h\) and \(C\) are fixed for a given layer, minimizing this upper bound is equivalent to maximizing the retained attention mass:

\[
\sum_{h=0}^{N_h -1}
\sum_{i=0}^{L-W-1}
I_i^{l,h} a_i^{l,h}
\]

Given a head-wise budget $b^{l,h}$, selecting the Top-$b^{l,h}$ entries inside each head maximizes the retained attention mass for that head. Therefore, the key problem becomes how to allocate the limited layer-wise budget across heads.

The standard deviation of $a^{l,h}$ provides an empirical indication of attention concentration in the distributions studied here. However, directly computing and using second-order statistics introduces additional operations. We therefore use the maximum aggregated observation-window attention score:

\[
a_{\max}^{l,h}=\max_i a_i^{l,h}
\]

as a lightweight proxy for attention concentration. The historical-token score vector is non-negative and has a sum bounded by one. When its second moment is dominated by the largest entry and the squared mean is negligible, the following heuristic relation is motivated:

\[
\sigma^{l,h} \propto \frac{1}{\sqrt{L-W}} a_{\max}^{l,h}.
\]

We therefore use \(a_{\max}^{l,h}\) as a ranking-level proxy, rather than a strict estimator, for head-wise attention concentration. Among heads not flagged by the abnormal-head mask, smaller \(a_{\max}^{l,h}\) values lead to larger adaptive budgets. Appendix~\ref{appsubsec:Validation_of_the_Max-Attention_Proxy} provides empirical support across the tested model scales and window sizes, while also identifying low-response outliers for which the proxy can be unreliable.

\section{Methodology}
\label{sec:methodology}

Given the aggregated observation-window attention scores \(a^{l,h}\) defined in Section~\ref{subsec:Head-Wise_Distribution_Heterogeneity}, \textsc{ChampionKV} first computes the champion score, the maximum aggregated observation-window attention score \(a_{\max}^{l,h}\) defined in Section \ref{subsec:champion_motivation}.

\paragraph{Floor budget.}
\textit{For clarity, we describe the procedure based on the SnapKV pipeline.} Let \(B_{\mathrm{base}}\) denote the predefined KV cache budget for each head, and let \(W\) denote the observation window size. The per-head historical-token budget after excluding the observation window is:

\[
b_{\mathrm{hist}} = B_{\mathrm{base}} - W
\]

For each layer, the total historical KV cache budget is therefore:

\[
B_{\mathrm{hist}} = N_h \cdot b_{\mathrm{hist}}
\]

where \(N_h\) is the number of attention heads in the layer. We assign each head a fixed floor budget:

\[
b_{\mathrm{floor}}
=
\left\lfloor
\rho \cdot b_{\mathrm{hist}}
\right\rfloor
\]

where \(\rho\) is the floor-budget ratio, which controls the trade-off between guaranteed historical-token retention and adaptive redistribution. Increasing \(\rho\) gives each head a larger guaranteed budget but leaves less budget for adaptive allocation. We use \(\rho=0.2\) as a practical default; the preferred balance can vary across models and workloads, as examined in our hyperparameter analysis.

\paragraph{Abnormal-head mask.}
We construct the abnormal-head mask according to \(a_{\max}^{l,h}\):

\[
M_{\mathrm{abn}}^{l,h}
=
\mathbb{I}\left(a_{\max}^{l,h}<\tau\right)
\]

where \(\tau\) is the threshold for identifying abnormal heads. Equivalently:

\[
M_{\mathrm{abn}}^{l,h}
=
\begin{cases}
1, & a_{\max}^{l,h}<\tau,\\
0, & a_{\max}^{l,h}\ge \tau.
\end{cases}
\]

Heads with \(M_{\mathrm{abn}}^{l,h}=1\) are designated as abnormal heads under this threshold-based rule. They retain only the floor budget and are excluded from adaptive budget competition, preventing very small champion scores from receiving disproportionately large inverse-max weights. The threshold \(\tau=0.01\) is an empirical default rather than a universal boundary between attention patterns; its suitability may depend on the model and context length.

\paragraph{Inverse-max budget allocation.}
After floor allocation, the remaining budget is:

\[
B_{\mathrm{rem}}
=
N_h\cdot(b_{\mathrm{hist}}
-
 b_{\mathrm{floor}}).
\]

The remaining budget is allocated only within the non-abnormal head set:

\[
\mathcal{H}^l_{\mathrm{valid}}
=
\{h \mid M_{\mathrm{abn}}^{l,h}=0\}
\]

For each \(h\in\mathcal{H}^l_{\mathrm{valid}}\), we compute the inverse-max allocation weight:

\[
w^{l,h}
=
\frac{
\left(a_{\max}^{l,h}+\epsilon\right)^{-1}
}{
\sum_{j\in\mathcal{H}^l_{\mathrm{valid}}}
\left(a_{\max}^{l,j}+\epsilon\right)^{-1}
},
\]

where \(\epsilon\) is a small constant for numerical stability. The historical KV cache budget \(b^{l,h}\) of the \(h\)-th head is then given by:

\[
b^{l,h}=\begin{cases} b_{\mathrm{floor}}, & M_{\mathrm{abn}}^{l,h}=1,\\ \widehat{b}^{l,h}, & M_{\mathrm{abn}}^{l,h}=0,\end{cases}
\]

\[ \widehat{b}^{l,h} = \min\!\Bigl( b_{\mathrm{floor}} + \bigl\lfloor w^{l,h} B_{\mathrm{rem}} \bigr\rfloor, L-W \Bigr) \]

If unused budget remains due to rounding or the maximum allocatable length of an individual head, this budget is placed into a shared residual budget pool and further assigned according to the weight order of non-abnormal heads until the budget is exhausted or all allocatable heads reach their upper bound. This avoids waste caused by budget overflow and ensures that the limited KV cache budget is allocated to effective heads as much as possible.

\paragraph{Token selection.}
Finally, within each head, we perform Top-\(b^{l,h}\) selection based on the aggregated observation-window attention scores. To reduce the influence of local noise, we apply local pooling smoothing to \(a^{l,h}\)~\citep{li2024snapkv,cai2024pyramidkv}, obtaining the score vector:

\[
s^{l,h}
=
\operatorname{Pool}_{\kappa}(a^{l,h}),
\]

where \(\kappa\) is the kernel size. The selected historical token indices are:

\[
\mathcal{I}^{l,h}
=
\operatorname{TopK}(s^{l,h}, b^{l,h}).
\]

Meanwhile, we always retain the recent tokens:
\[
\mathcal{R}
=
\{L-W,\ldots,L-1\}.
\]

Therefore, the final retained KV cache index set for the \(h\)-th head is:

\[
\mathcal{T}^{l,h}
=
\mathcal{I}^{l,h}
\cup
\mathcal{R}.
\]

Indexing \(K^{l,h}\) and \(V^{l,h}\) using \(\mathcal{T}^{l,h}\) yields the compressed KV cache. The complete procedure is shown in Algorithm~\ref{alg:champion_adjusted_short}.

\begin{algorithm}[t]
\normalsize
\caption{\textsc{ChampionKV}}
\label{alg:champion_adjusted_short}
\DontPrintSemicolon
\SetAlgoLined
\SetAlgoNlRelativeSize{-1}
\SetAlCapSkip{0.4em}
\SetKwInput{KwIn}{Input}
\SetKwInput{KwOut}{Output}

\KwIn{$K^{l,h},V^{l,h},Q^{l,h}$; historical length $L$; heads $N_h$; per-layer
  budget $b_{\mathrm{hist}}$; window $W$;}
\KwOut{Compressed cache $\widetilde{K}^{l,h},\widetilde{V}^{l,h}$.}

$A^{l,h}=\mathrm{softmax}\!\left(Q^{l,h}_{L-W:L}(K^{l,h})^{\top}/\sqrt{d}\right)$\;
$a_i^{l,h}=\tfrac{1}{W}\sum_{t=L-W}^{L-1}A^{l,h}_{t,i}$\;
$s^{l,h}=\operatorname{Pool}(a^{l,h})$\;
$a_{\max}^{l,h}=\max_i s_i^{l,h}$;
$M_{\mathrm{abn}}^{l,h}=\mathbb{I}(a_{\max}^{l,h}<\tau)$\;
$b_{\mathrm{floor}}=\lfloor\rho\, b_{\mathrm{hist}}\rfloor$;
$B_{\mathrm{rem}}= N_h (b_{\mathrm{hist}}-b_{\mathrm{floor}})$\;
$\mathcal{H}^l_{\mathrm{valid}}=\{h\mid M_{\mathrm{abn}}^{l,h}=0\}$\;
$w^{l,h}=\dfrac{(a_{\max}^{l,h}+\epsilon)^{-1}}{\sum_{j\in\mathcal{H}^l_{\mathrm{valid}}}(a_{\max}^{l,j}+\epsilon)^{-1}}$\;
\ForEach{$h\in\mathcal{H}^l_{\mathrm{valid}}$}{
    $\widehat{b}^{l,h}=\min\!\bigl(b_{\mathrm{floor}}+\lfloor w^{l,h}B_{\mathrm{rem}}\rfloor,\,L-W\bigr)$\;
    $b^{l,h}\leftarrow\widehat{b}^{l,h}$\;
}
$b^{l,h}\leftarrow b_{\mathrm{floor}}$ for $h\notin\mathcal{H}^l_{\mathrm{valid}}$\;
Assign leftover units to eligible heads in descending $w^{l,h}$ order\;

\ForEach{head $h$}{
    $\mathcal{I}^{l,h}=\operatorname{TopK}(s^{l,h},b^{l,h})$\;
    $\mathcal{T}^{l,h}=\mathcal{I}^{l,h}\cup\{L-W,\ldots,L-1\}$\;
    $\widetilde{K}^{l,h}= K^{l,h}_{\mathcal{T}^{l,h}},\quad\widetilde{V}^{l,h}= V^{l,h}_{\mathcal{T}^{l,h}}$\;
}
\Return{$\widetilde{K}^{l,h},\widetilde{V}^{l,h}$}\;

\end{algorithm}

\section{Experiments}
\subsection{Experimental Setup}

\noindent \textbf{Benchmarks.} We evaluate on LongBench~\citep{bai2024LongBench} and RULER~\citep{hsieh2024ruler}. LongBench covers real-world long-context tasks such as QA, summarization, few-shot learning, synthetic retrieval, and code completion. RULER provides controllable synthetic tasks for evaluating effective context length. Detailed task descriptions are provided in Appendix~\ref{sec:detailed_benchmarks}.

\noindent \textbf{Base models.} We use three models for experiments: Llama3.1-8B-Instruct, Llama3.1-70B-Instruct~\cite{grattafiori2024Llama}, and Mistral-7B-Instruct-v0.3~\cite{chaplot2023albert}.

\noindent \textbf{Baselines.} We use SnapKV~\cite{li2024snapkv} and PyramidKV~\cite{cai2024pyramidkv} as base compressors and compare them with their Champion-adjusted variants, denoted as Champion-Snap and Champion-Pyramid, respectively. These comparisons isolate the effect of head-wise budget reallocation while preserving each compressor's token-selection procedure. We additionally include Ada-KV~\cite{feng2026ada} as a directly relevant online adaptive allocation baseline. FullKV serves as the uncompressed reference.

\noindent \textbf{Hyperparameters.} For PyramidKV, the layer-wise allocation hyperparameter is set to $\beta=20$~\cite{cai2024pyramidkv}. Unless otherwise specified in the hyperparameter analysis, we use a floor-budget ratio of $\rho=0.2$ and an abnormal-head threshold of $\tau=0.01$ for \textsc{ChampionKV} across the evaluation.

\begin{table*}[!t]
    \centering
    \normalsize
    \setlength{\tabcolsep}{2.4pt}
    \renewcommand{\arraystretch}{1.0}
    \resizebox{\textwidth}{!}{
    \begin{tabular}{l ccc ccc ccc ccc cc cc c}
    \toprule
    & \multicolumn{3}{c}{Single-Doc QA}
    & \multicolumn{3}{c}{Multi-Doc QA}
    & \multicolumn{3}{c}{Summarization}
    & \multicolumn{3}{c}{Few-shot Learning}
    & \multicolumn{2}{c}{Synthetic}
    & \multicolumn{2}{c}{Code}
    & \\
    \cmidrule(lr){2-4}
    \cmidrule(lr){5-7}
    \cmidrule(lr){8-10}
    \cmidrule(lr){11-13}
    \cmidrule(lr){14-15}
    \cmidrule(lr){16-17}
    \raisebox{1em}{Method}
    & \rotatebox[origin=c]{35}{NrtvQA}
    & \rotatebox[origin=c]{35}{Qasper}
    & \rotatebox[origin=c]{35}{MF-en}
    & \rotatebox[origin=c]{35}{HotpotQA}
    & \rotatebox[origin=c]{35}{2WikiMQA}
    & \rotatebox[origin=c]{35}{Musique}
    & \rotatebox[origin=c]{35}{GovReport}
    & \rotatebox[origin=c]{35}{QMSum}
    & \rotatebox[origin=c]{35}{MultiNews}
    & \rotatebox[origin=c]{35}{TREC}
    & \rotatebox[origin=c]{35}{TriviaQA}
    & \rotatebox[origin=c]{35}{SAMSum}
    & \rotatebox[origin=c]{35}{PCount}
    & \rotatebox[origin=c]{35}{PRe}
    & \rotatebox[origin=c]{35}{LCC}
    & \rotatebox[origin=c]{35}{RB-P}
    & \rotatebox[origin=c]{35}{Avg.} \\
    \midrule

    \multicolumn{18}{c}{KV Budget = Full} \\
    \midrule
    FullKV
    & 27.41 & 46.02 & 53.84
    & 53.96 & 59.35 & 33.42
    & 34.48 & 24.79 & 27.89
    & 68.00 & 91.33 & 47.18
    & 14.00 & 66.00
    & 64.08 & 54.92 & 47.92 \\

    \midrule
    \multicolumn{18}{c}{KV Budget = 64} \\
    \midrule
    Ada-KV
    & \textbf{27.17} & \textbf{42.69} & \textbf{52.08}
    & \textbf{44.56} & \textbf{52.41} & \textbf{24.65}
    & \textbf{23.01} & \textbf{22.53} & \textbf{22.37}
    & \textbf{67.00} & \textbf{91.27} & 44.25
    & \textbf{14.00} & \textbf{66.00}
    & 32.98 & \textbf{56.83}  & \textbf{42.74} \\
    
    SnapKV
    & 25.48 & 33.80 & 47.02
    & 42.36 & 44.46 & 22.83
    & 20.70 & 21.35 & 21.09
    & 60.00 & 88.26 & 43.15
    & \textbf{14.00} & \textbf{66.00}
    & 36.98 & 53.45 & 40.06 \\

    Champion-Snap
    & 26.69 & 35.58 & 49.30
    & 43.26 & 49.87 & 24.19
    & 22.00 & 21.65 & 21.13
    & 65.00 & 90.23 & 44.23
    & \textbf{14.00} & \textbf{66.00}
    & \textbf{40.94} & 54.36  & 41.78 \\

    PyramidKV
    & 25.15 & 32.28 & 48.10
    & 43.56 & 50.14 & 24.06
    & 21.88 & 22.33 & 21.87
    & 58.00 & 88.73 & \textbf{44.89}
    & \textbf{14.00} & \textbf{66.00}
    & 40.57 & 53.56   & 40.95 \\

    Champion-Pyramid
    & 26.79 & 38.64 & 48.98
    & 43.84 & 50.13 & 22.13
    & 22.49 & 22.21 & 22.25
    & 63.00 & 89.59 & 44.14
    & \textbf{14.00} & \textbf{66.00}
    & 40.92 & 54.65 & 41.86 \\

    \midrule
    \multicolumn{18}{c}{KV Budget = 128} \\
    \midrule
    Ada-KV
    & \textbf{27.27} & \textbf{43.41} & \textbf{53.15}
    & 50.22 & \textbf{53.36} & \textbf{26.61}
    & 24.49 & 23.23 & 23.32
    & \textbf{68.00} & \textbf{91.43} & \textbf{48.18}
    & \textbf{14.00} & \textbf{66.00}
    & 43.79 & \textbf{60.97}  & \textbf{44.84} \\
    
    SnapKV
    & 26.30 & 38.98 & 50.99
    & 44.24 & 51.57 & 24.88
    & 23.42 & 22.41 & 23.07
    & \textbf{68.00} & 90.67 & 45.86
    & \textbf{14.00} & \textbf{66.00}
    & 52.49 & 54.32 & 43.58 \\

    Champion-Snap
    & 24.17 & 40.81 & 49.91
    & 45.92 & 52.26 & 25.72
    & 23.92 & 23.09 & 23.38
    & 67.00 & \textbf{91.43} & 47.54
    & \textbf{14.00} & \textbf{66.00}
    & \textbf{54.35} & 57.03  & 44.16 \\

    PyramidKV
    & 27.00 & 39.63 & 52.81
    & 46.72 & 51.20 & 25.05
    & 24.21 & 23.34 & 23.18
    & 67.00 & 89.42 & 45.99
    & \textbf{14.00} & \textbf{66.00}
    & 45.35 & 55.14  & 43.50 \\

    Champion-Pyramid
    & 24.92 & 41.86 & 52.41
    & \textbf{51.33} & 52.57 & 26.24
    & \textbf{24.80} & \textbf{23.47} & \textbf{24.11}
    & 67.00 & 90.23 & 47.86
    & \textbf{14.00} & \textbf{66.00}
    & 42.77 & 58.63  & 44.26 \\

    \midrule
    \multicolumn{18}{c}{KV Budget = 512} \\
    \midrule
    Ada-KV
    & 27.81 & \textbf{45.56} & 54.05
    & \textbf{54.89} & 58.30 & 31.13
    & 27.13 & \textbf{24.17} & 25.45
    & \textbf{68.00} & \textbf{91.43} & \textbf{48.16}
    & \textbf{14.00} & \textbf{66.00}
    & 61.26 & 62.61  & 47.50 \\
    
    SnapKV
    & 27.23 & 43.10 & 53.99
    & 53.29 & 57.60 & 30.64
    & 27.83 & 23.62 & 25.59
    & \textbf{68.00} & \textbf{91.43} & 47.89
    & \textbf{14.00} & \textbf{66.00}
    & 62.96 & 61.74  & 47.18 \\

    Champion-Snap
    & 27.54 & 45.11 & 53.81
    & 53.75 & \textbf{59.44} & 31.04
    & 27.74 & 23.69 & \textbf{26.13}
    & \textbf{68.00} & \textbf{91.43} & 48.09
    & \textbf{14.00} & \textbf{66.00}
    & 63.02 & \textbf{62.95} & \textbf{47.61} \\

    PyramidKV
    & 27.41 & 43.31 & 52.74
    & 52.02 & 57.68 & 29.98
    & 27.41 & 23.99 & 25.65
    & \textbf{68.00} & 90.23 & 47.32
    & \textbf{14.00} & \textbf{66.00}
    & 62.00 & 57.74  & 46.59 \\

    Champion-Pyramid
    & \textbf{27.96} & 45.53 & \textbf{54.25}
    & 54.37 & 58.35 & \textbf{33.42}
    & \textbf{28.34} & 23.93 & 25.75
    & \textbf{68.00} & 89.73 & 47.40
    & \textbf{14.00} & \textbf{66.00}
    & \textbf{63.57} & 57.84 & 47.40 \\

    \midrule
    \multicolumn{18}{c}{KV Budget = 2048} \\
    \midrule
    Ada-KV
    & \textbf{27.73} & 46.04 & 54.19
    & 52.93 & \textbf{59.84} & 33.28
    & 31.02 & 24.14 & 27.54
    & \textbf{68.00} & \textbf{91.93} & 47.65
    & \textbf{14.00} & \textbf{66.00}
    & 62.13 & \textbf{56.85} & 47.70 \\
    
    SnapKV
    & 27.55 & 44.86 & 53.62
    & 53.22 & 58.95 & \textbf{33.65}
    & 31.67 & \textbf{24.76} & 27.62
    & \textbf{68.00} & \textbf{91.93} & \textbf{47.81}
    & \textbf{14.00} & \textbf{66.00}
    & 65.32 & 56.13 & 47.82 \\

    Champion-Snap
    & 27.72 & \textbf{46.16} & 54.37
    & \textbf{54.00} & 59.33 & 33.44
    & \textbf{32.51} & 24.16 & 27.86
    & \textbf{68.00} & \textbf{91.93} & 47.39
    & \textbf{14.00} & \textbf{66.00}
    & \textbf{66.83} & 56.55 & \textbf{48.14} \\

    PyramidKV
    & 27.52 & 44.76 & \textbf{54.58}
    & 52.31 & 58.88 & 32.91
    & 31.95 & 24.45 & \textbf{28.13}
    & \textbf{68.00} & \textbf{91.93} & 47.68
    & \textbf{14.00} & \textbf{66.00}
    & 63.79 & 55.80 & 47.67 \\

    Champion-Pyramid
    & 27.58 & 45.56 & 54.21
    & 52.96 & 59.03 & 33.17
    & 32.29 & 24.72 & 27.87
    & \textbf{68.00} & \textbf{91.93} & 47.08
    & \textbf{14.00} & \textbf{66.00}
    & 63.40 & 56.59 & 47.77 \\

    \bottomrule
    \end{tabular}
    }
    \caption{LongBench results on Llama3.1-70B-Instruct. Avg. is the mean over all 16 tasks. FullKV is the uncompressed reference; bold marks the best compressed result within each budget.}
    \label{tab:LongBench_Llama31_70b}
\end{table*}

\subsection{Main results}

\paragraph{Results for LongBench} We consider a group of KV cache budgets, $B_{base}\in\{64,128,512,2048\}$, and use $W=32$ throughout.

Table~\ref{tab:LongBench_Llama31_70b} reports the LongBench results on Llama3.1-70B-Instruct. The comparison is mixed: Ada-KV has the highest 16-task average at budgets 64 and 128 (42.74 and 44.84), while Champion-Snap is strongest at 512 and 2048 (47.61 and 48.14), narrowly above Ada-KV (47.50 and 47.70). Task-level results are complementary: Ada-KV is strong under tight budgets, and the Champion variants lead on subsets at moderate and larger budgets; PCount and PRe are saturated and do not distinguish the methods. Against the base compressors, however, both Champion variants improve their 16-task averages over the corresponding SnapKV or PyramidKV allocation at the majority of budgets, with task-level wins concentrated at the tight end (paired analysis in \S\ref{sec:paired_consistency}); relative to Ada-KV's global cross-head Top-$K$ selection, the margin is small and budget-dependent, so \textsc{ChampionKV} should be viewed as an upgrade to observation-window compressors rather than a uniformly stronger allocator.

Complete task-level LongBench results on Llama3.1-8B-Instruct and Mistral-7B-Instruct-v0.3, including Ada-KV, are in Appendix~\ref{app:experiments}. Ada-KV leads the 16-task average at three of four budgets on Llama3.1-8B-Instruct and at all four on Mistral-7B-Instruct-v0.3. The Champion variants remain competitive on individual tasks but do not lead the average, indicating that the 70B gains do not transfer uniformly to a smaller model.

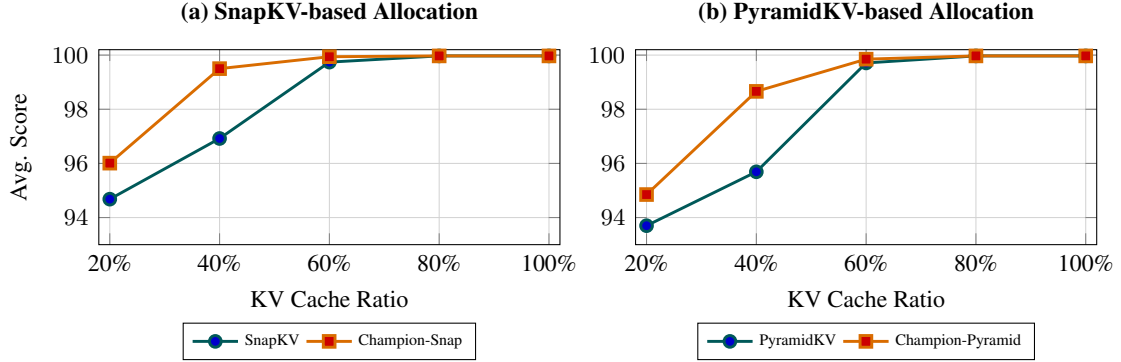
\begin{figure*}[!t]
    \centering
    \begin{tikzpicture}
        \begin{groupplot}[
            group style={
                group size=2 by 1,
                horizontal sep=1.0cm
            },
            width=0.48\textwidth,
            height=0.26\textwidth,
            xlabel={KV Cache Ratio},
            ylabel={Avg. Score},
            xmin=0.18,
            xmax=1.02,
            ymin=93,
            ymax=100.2,
            xtick={0.2,0.4,0.6,0.8,1.0},
            xticklabels={20\%,40\%,60\%,80\%,100\%},
            grid=both,
            grid style={
                line width=.1pt,
                draw=gray!20
            },
            major grid style={
                line width=.2pt,
                draw=gray!35
            },
            tick label style={font=\small},
            label style={font=\small},
            title style={font=\small\bfseries},
            legend style={
                font=\tiny,
                at={(0.5,-0.40)},
                anchor=north,
                legend columns=2
            },
            every axis plot/.append style={
                line width=1.1pt,
                mark size=2.2pt
            }
        ]

            \nextgroupplot[title={(a) SnapKV-based Allocation}]
                \addplot+[mark=*, teal!70!black] coordinates {
                    (0.2,94.68)
                    (0.4,96.92)
                    (0.6,99.74)
                    (0.8,99.97)
                    (1.0,99.97)
                };
                \addlegendentry{SnapKV}

                \addplot+[mark=square*, orange!85!black] coordinates {
                    (0.2,96.01)
                    (0.4,99.50)
                    (0.6,99.94)
                    (0.8,99.97)
                    (1.0,99.97)
                };
                \addlegendentry{Champion-Snap}

            \nextgroupplot[
                title={(b) PyramidKV-based Allocation},
                ylabel={}
            ]
                \addplot+[mark=*, teal!70!black] coordinates {
                    (0.2,93.70)
                    (0.4,95.69)
                    (0.6,99.71)
                    (0.8,99.97)
                    (1.0,99.97)
                };
                \addlegendentry{PyramidKV}

                \addplot+[mark=square*, orange!85!black] coordinates {
                    (0.2,94.85)
                    (0.4,98.66)
                    (0.6,99.85)
                    (0.8,99.97)
                    (1.0,99.97)
                };
                \addlegendentry{Champion-Pyramid}

        \end{groupplot}
    \end{tikzpicture}
   \caption{RULER performance on Llama3.1-70B-Instruct under different KV cache ratios. }
    \label{fig:ruler_Llama31_70b}
\end{figure*}

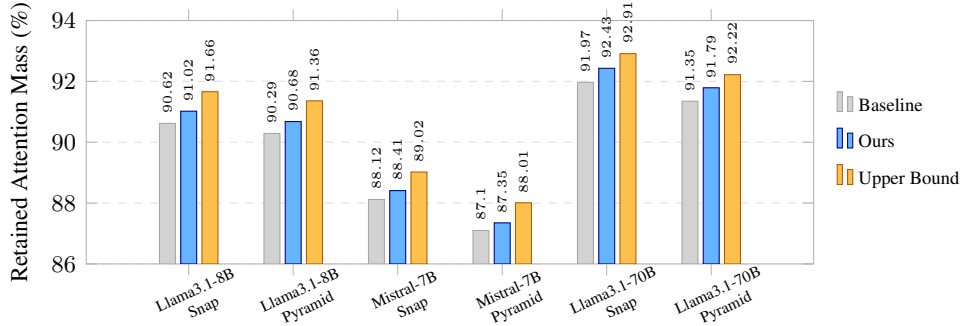
\begin{figure*}[!t]
    \centering
    \begin{tikzpicture}
        \begin{axis}[
            width=0.7\textwidth,
            height=0.30\textwidth,
            ybar,
            bar width=6pt,
            ymin=86,
            ymax=94,
            ylabel={Retained Attention Mass (\%)},
            xmin=0.5,
            xmax=6.5,
            xtick={1,2,3,4,5,6},
            xticklabels={
                \shortstack{Llama3.1-8B\\Snap},
                \shortstack{Llama3.1-8B\\Pyramid},
                \shortstack{Mistral-7B\\Snap},
                \shortstack{Mistral-7B\\Pyramid},
                \shortstack{Llama3.1-70B\\Snap},
                \shortstack{Llama3.1-70B\\Pyramid}
            },
            x tick label style={
               font=\fontsize{6pt}{6pt}\selectfont,
                align=center,
                rotate=25,
                anchor=north
            },
            ymajorgrids=true,
            grid style={dashed, gray!25},
            axis line style={gray!60},
            tick style={gray!60},
            ylabel style={font=\small},
            tick label style={font=\small},
            enlarge x limits=0.08,
            legend style={
                at={(1.02,0.5)},
                anchor=west,
                legend columns=1,
                draw=none,
                fill=none,
                font=\scriptsize,
                row sep=0.25em,
                cells={anchor=west},
                legend cell align=left
            },
            legend image post style={xscale=0.65, yscale=0.85},
            nodes near coords,
            every node near coord/.append style={
                font=\fontsize{5pt}{5pt}\selectfont,
                anchor=west,
                rotate=90,
            },
            nodes near coords align={vertical},
        ]

        \addplot[
            fill=gray!35,
            draw=gray!70
        ] coordinates {
            (1,90.62)
            (2,90.29)
            (3,88.12)
            (4,87.10)
            (5,91.97)
            (6,91.35)
        };

        \addplot[
            fill=blue!50!cyan!60,
            draw=blue!55!black
        ] coordinates {
            (1,91.02)
            (2,90.68)
            (3,88.41)
            (4,87.35)
            (5,92.43)
            (6,91.79)
        };

        \addplot[
            fill=orange!70!yellow!70,
            draw=orange!70!black
        ] coordinates {
            (1,91.66)
            (2,91.36)
            (3,89.02)
            (4,88.01)
            (5,92.91)
            (6,92.22)
        };

        \legend{Baseline, Ours, Upper Bound}

        \end{axis}
    \end{tikzpicture}
    \caption{Retained attention mass on randomly sampled LongBench prompts with KV cache budget 512 and observation window size \(W=32\). Baseline denotes the original SnapKV or PyramidKV allocation, while Ours denotes the same compressor after applying Champion head-wise reallocation.}
    \label{fig:retained_attention_mass_LongBench}
\end{figure*}

\paragraph{Results for RULER}

On Llama3.1-70B-Instruct, Figure~\ref{fig:ruler_Llama31_70b} shows that Champion-Snap improves over SnapKV by 1.33, 2.58, and 0.20 points at 20\%, 40\%, and 60\% retention and ties at 80\%; it also exceeds Ada-KV at 20--60\% (96.01 vs.\ 94.68, 99.50 vs.\ 97.40, 99.94 vs.\ 99.76) and ties at 80\% (99.97). The largest separation is at 40\% retention, and the methods converge as retention approaches 80\%.

Complete RULER results for all models and ratios are in Appendix~\ref{app:experiments}. The smaller-model comparisons are mixed: Champion-Snap exceeds Ada-KV at 20--60\% retention on Llama3.1-8B-Instruct and ties at 80\%, while Ada-KV is lower than the strongest tested method at all ratios on Mistral-7B-Instruct-v0.3. The tested 70B model gives the clearest downstream gains, a pattern that should not be generalized across families.

\paragraph{Hardware efficiency.}
We profile Llama3.1-8B-Instruct with a KV cache budget of 1024 on \textbf{NVIDIA H200}, using a 32,100-token input and 15 generated tokens (Table~\ref{tab:hardware_efficiency_main_app}). Champion-Snap and Champion-Pyramid use 0.375 GB and 0.365 GB less peak GPU memory than Ada-KV with comparable total generation time (4.66 ms and 53.31 ms faster), while SnapKV retains the lowest latency and total time. This supports a modest memory reduction under the tested configuration rather than a universal speed advantage; profiling at additional input lengths is in Appendix~\ref{app:experiments}.

\subsection{Further Analysis}

\subsubsection{Retained Attention Mass Coverage}
We evaluate retained observation-window attention mass with a total cache budget of 512 and $W=32$, changing only the head-wise allocation while keeping the underlying token-selection procedure fixed. Figure~\ref{fig:retained_attention_mass_LongBench} shows that Champion allocation improves retained mass for all six model--compressor pairs and moves closer to the upper-bound reference. Because Ada-KV uses global cross-head Top-$K$ selection, this controlled proxy comparison does not establish superiority over Ada-KV; direct downstream comparisons are reported in the LongBench and RULER results. Because the token-selection procedure is held fixed, the retained-mass gain can be attributed to the head-wise allocation, and its consistency across models and compressors foreshadows the downstream paired gains reported in the follwing section\S\ref{sec:paired_consistency}.

\subsubsection{Robustness Analysis of Hyper-parameters}
We sweep $\rho$ and $\tau$ on one task from each of the six LongBench categories. Preferred settings vary with model and budget: on Llama3.1-8B the best values differ across budgets, while $(\rho,\tau)=(0.2,0.01)$ gives the highest six-task average on Mistral-7B-Instruct-v0.3 (38.93). We therefore use this pair as a robustness-oriented default rather than a universal optimum; details are in Appendix~\ref{app:experiments}.

\paragraph{Prevalence of abnormal heads.}
At $\tau=0.01$, abnormal heads account for 0.57--1.66\% of layer--head instances on Llama3.1-8B and 0.81--1.50\% on Mistral-7B-Instruct-v0.3. Thus, the mask affects only a small minority of heads under the tested protocol; detailed counts and prompt-level ranges are provided in Appendix~\ref{app:experiments}. Under inverse-max allocation an under-concentrated head would otherwise capture an outsized share of the budget, so masking such heads preempts this at negligible cost; the low observed prevalence indicates this guard is rarely triggered under the tested protocol.

\paragraph{Stability across input lengths.}
Across 8k--32k inputs, six-task averages vary by at most 0.55 points on Llama3.1-8B and 0.32 on Mistral-7B-Instruct-v0.3, indicating limited sensitivity within this fixed protocol. However, this slice does not replace evaluation of generation-heavy, long-chain-of-thought, or multi-turn use; complete results are provided in Appendix~\ref{app:experiments}.

\subsubsection{Paired Consistency Analysis}
\label{sec:paired_consistency}
All scores use greedy decoding (deterministic per task), so mean differences alone cannot confirm consistent advantages. We conduct two-sided Wilcoxon signed-rank tests on paired deltas (LongBench: 16 tasks × 4 budgets; RULER: 11 subtasks × 4 ratios; Table~\ref{tab:paired_delta_stats_app}).

For Llama3.1-70B, gains over both baselines are significant on LongBench ($p<0.001$; win rates 68.8\% and 60.9\%), and Champion-Snap yields a significant RULER gain with zero losing pairs ($p=0.016$). On 7B/8B models, LongBench deltas are insignificant (win rates 39–44\%). Benefits are strongest where head-wise heterogeneity is highest (70B) and mixed for smaller models. Decomposition shows 70B gains peak under tight budgets (Champion-Snap: +1.72 at budget 64 vs. +0.32 at 2048) and low-to-moderate RULER retention (20–40\%), vanishing at 80\% as methods saturate. No such monotonic trend appears for 7B/8B models, where isolated per-budget gains do not aggregate to significance.

\subsubsection{Discussion}
Taken together, the analyses delineate when head-wise reallocation is worthwhile. The champion-score proxy requires no offline estimation and introduces negligible overhead, and because token selection is never altered, \textsc{ChampionKV} can be enabled alongside any observation-window compressor. Its practical role is therefore to upgrade existing observation-window compressors under scarce-budget, heterogeneous-head regimes, rather than to replace the underlying compressor or global cross-head selection.

\section{Conclusion}

This paper presents \textsc{ChampionKV}, an input-adaptive head-wise allocator
for observation-window KV cache compression. It combines an
attention-concentration proxy with inverse-max allocation, abnormal-head
masking, and floor-budget protection without changing token selection. The
proxy uses only the maximum aggregated observation-window attention score,
computed during prefilling with no offline estimation or retraining. On
LongBench and RULER, Champion-Snap and Champion-Pyramid consistently improve
over their SnapKV and PyramidKV base compressors, with statistically
significant gains on 70B model under tight budgets and moderate
retention; on the tested 7B/8B models they remain competitive but do
not consistently outperform. Gains concentrate where heads are most
heterogeneous: tight budgets and low retention favor reallocating capacity
from concentrated to dispersed heads, preserving the attention mass behind
downstream accuracy. Controlled analysis confirms improved retained attention
mass over matched baselines. Because allocation is decoupled
from token scoring, \textsc{ChampionKV} composes with any
observation-window compressor and complements global cross-head selection.
Overall, \textsc{ChampionKV} is a practical plug-in that upgrades
observation-window KV cache compressors.

\clearpage

\section*{Limitations}

The maximum-attention score is an empirical proxy, and abnormal-head masking cannot prevent every unreliable case; fixed $\rho$ and $\tau$ may not transfer across models or workloads, motivating adaptive calibration. Evaluation covers limited models and tasks; we report paired per-task significance (\S\ref{sec:paired_consistency}) but not repeated-trial sampling-based significance tests, and the 8k--32k slice does not assess generation-heavy, long-chain-of-thought, or multi-turn settings. Finally, because \textsc{ChampionKV} reallocates budgets without changing token scoring, it inherits the token-selection limitations of the underlying compressor.

\section*{Declaration for Using AI Assistants}

During the preparation of this work, AI-assisted tools were used to support language polishing, grammar correction, and improvement of the clarity and readability of the text. The AI assistants were not used to generate original research ideas, perform data analysis, draw conclusions, or replace the authors' own critical thinking and academic judgment. All AI-generated suggestions were carefully reviewed, revised, and verified by the authors. The authors take full responsibility for the accuracy, integrity, and originality of the content presented in this work.


\bibliography{custom}

\clearpage
\appendix

\section{Appendix: Pilot experiments}

\subsection{Head-Wise Distribution Heterogeneity}
\label{app:Head-Wise_Distribution_Heterogeneity}

We analyze the distributional differences of aggregated observation-window attention across layers and heads during prefilling. Specifically, we use Llama3.1-8B-Instruct~\cite{grattafiori2024Llama} as the analysis model and NarrativeQA~\cite{kovcisky2018narrativeqa} as a representative dataset, and observe the aggregated observation-window attention distributions during the prefilling stage of model inference, since these distributions provide the key evidence for head-wise KV cache budget allocation.

\textbf{We summarize two typical attention-score distribution patterns:}

\begin{figure}[t]
    \centering

    \begin{subfigure}[t]{0.98\linewidth}
        \centering
        \includegraphics[width=\linewidth]{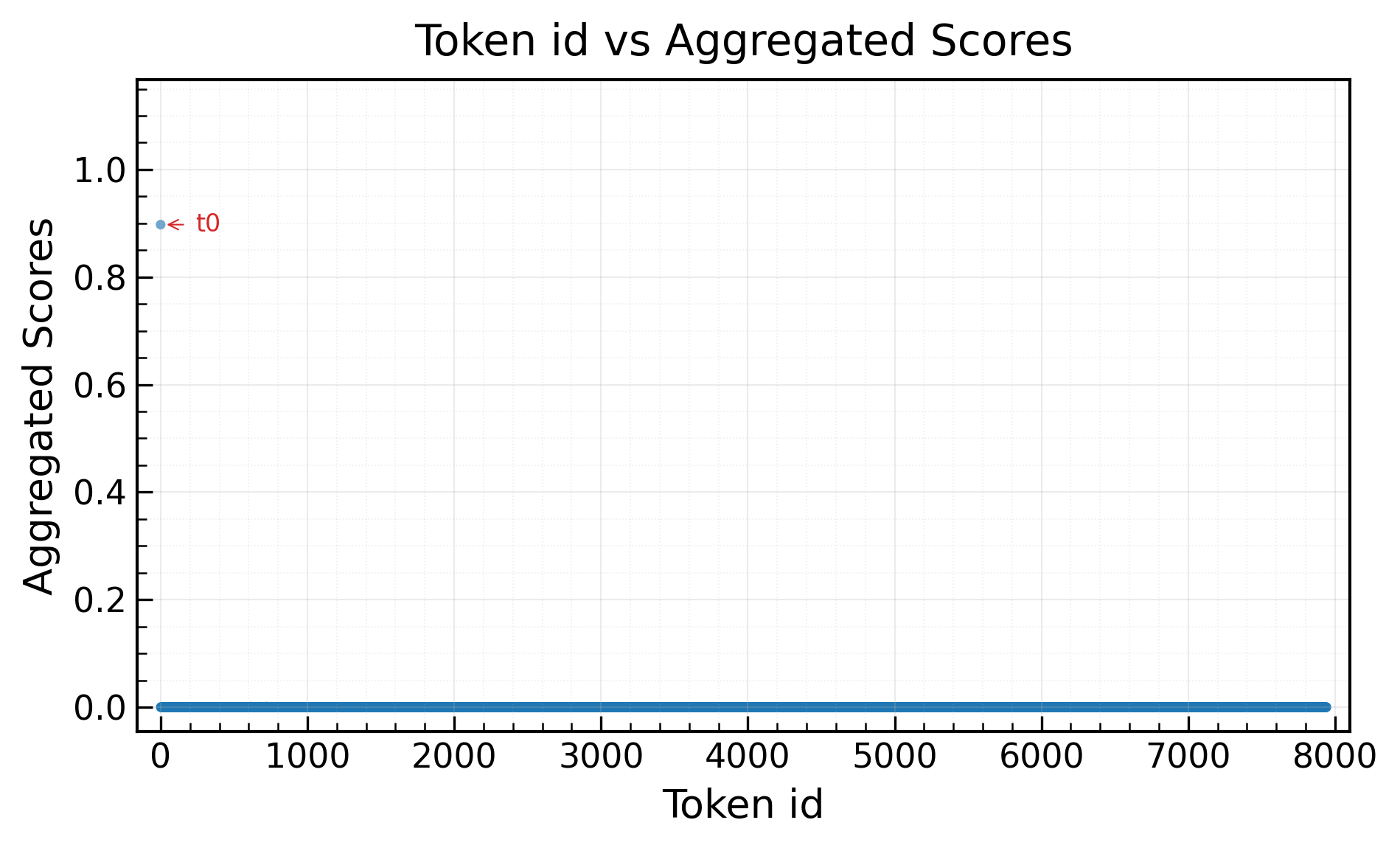}
        \caption{Linear Y-axis.}
        \label{fig:layer0-head31-attention}
    \end{subfigure}
    \vspace{0.6em}
    \begin{subfigure}[t]{1\linewidth}
        \centering
        \includegraphics[width=\linewidth]{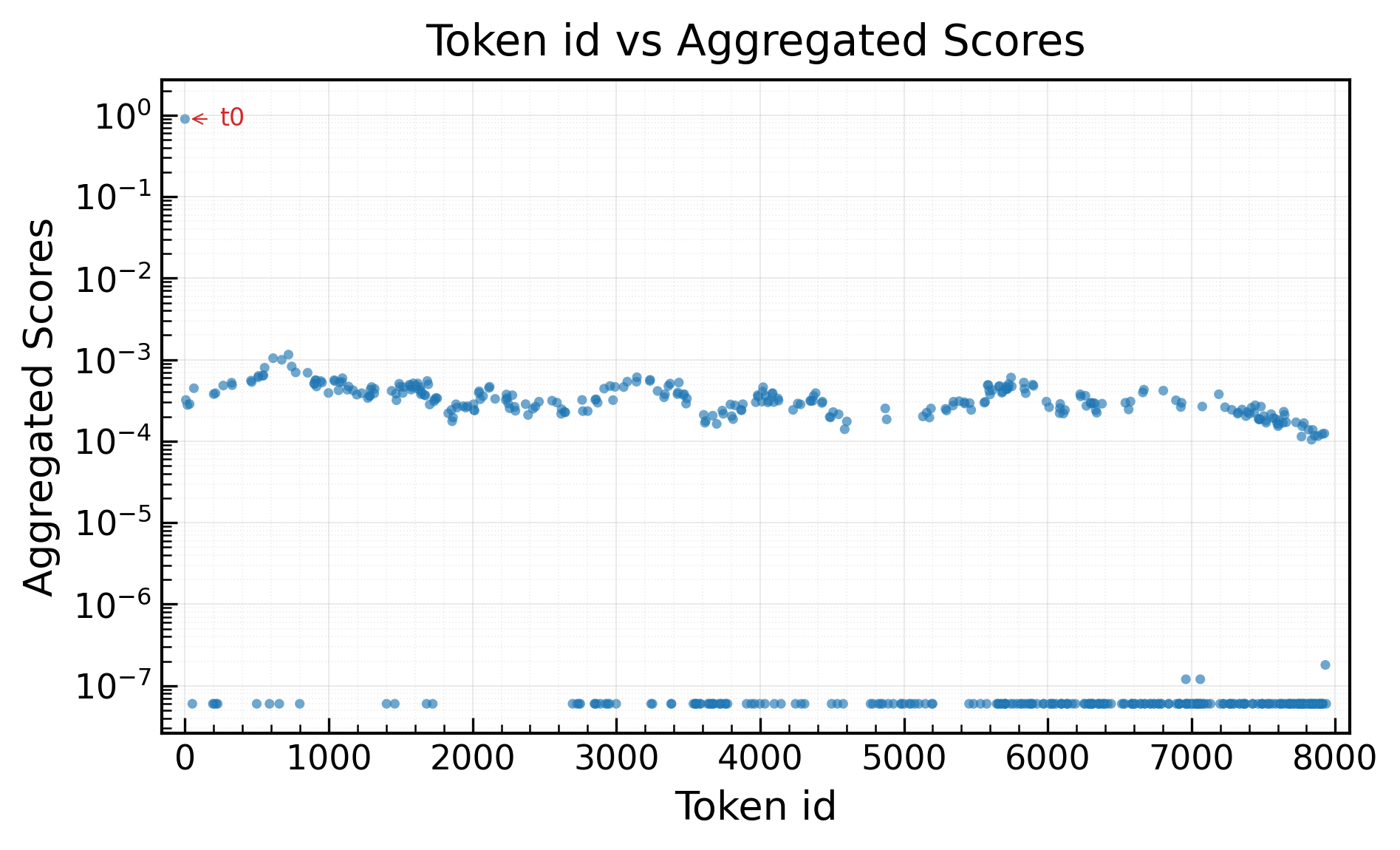}
        \caption{Log scale Y-axis.}
        \label{fig:layer0-head31-attention_log}
    \end{subfigure}

    \caption{Concentrated aggregated observation-window attention distribution of Layer 0, Head 31 in Llama3.1-8B-Instruct. The maximum score occurs at token 0 and most remaining historical tokens receive much lower scores.}

\end{figure}

\paragraph{``Concentrated Attention Distribution''}
\label{para:Concentrated_Attn}
This pattern means that the current attention head focuses on only a small number of historical tokens, with most observation-window attention mass concentrated on a few tokens, typically \texttt{<bos>}~\citep{xiao2024efficient} or task-relevant evidence tokens.  For example, Layer0 Head31 exhibits this pattern in Figure~\ref{fig:layer0-head31-attention}; the log-scale view in Figure~\ref{fig:layer0-head31-attention_log} makes the concentrated behavior clearer. This concentrated pattern suggests that the main attention mass of this head can be covered with relatively few KV entries.

\begin{figure}[t]
    \centering

    \begin{subfigure}[t]{1\linewidth}
        \centering
        \includegraphics[width=\linewidth]{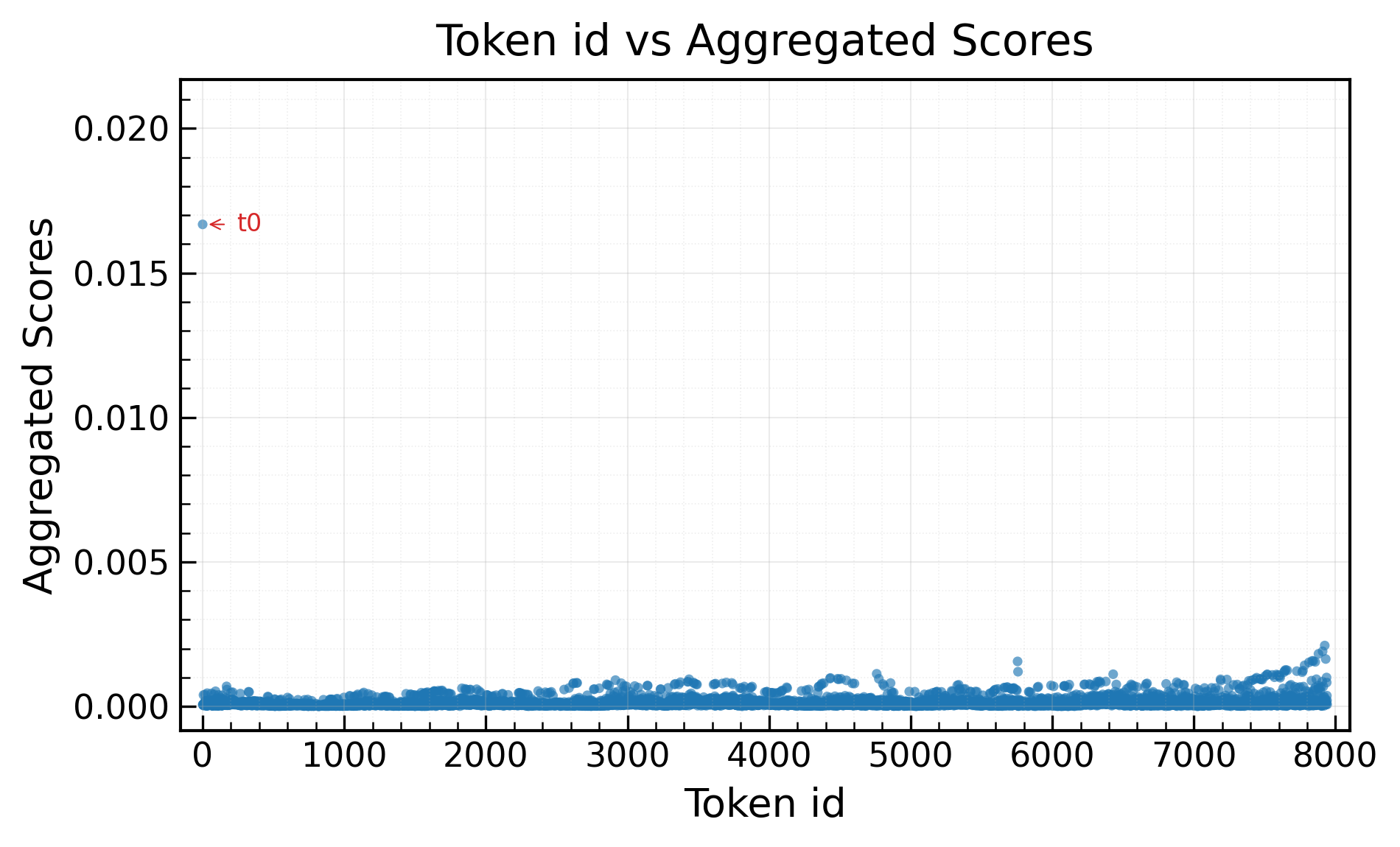}
        \caption{Linear Y-axis.}
        \label{fig:layer0-head22-attention}
    \end{subfigure}
    \vspace{0.6em}
    \begin{subfigure}[t]{1\linewidth}
        \centering
        \includegraphics[width=\linewidth]{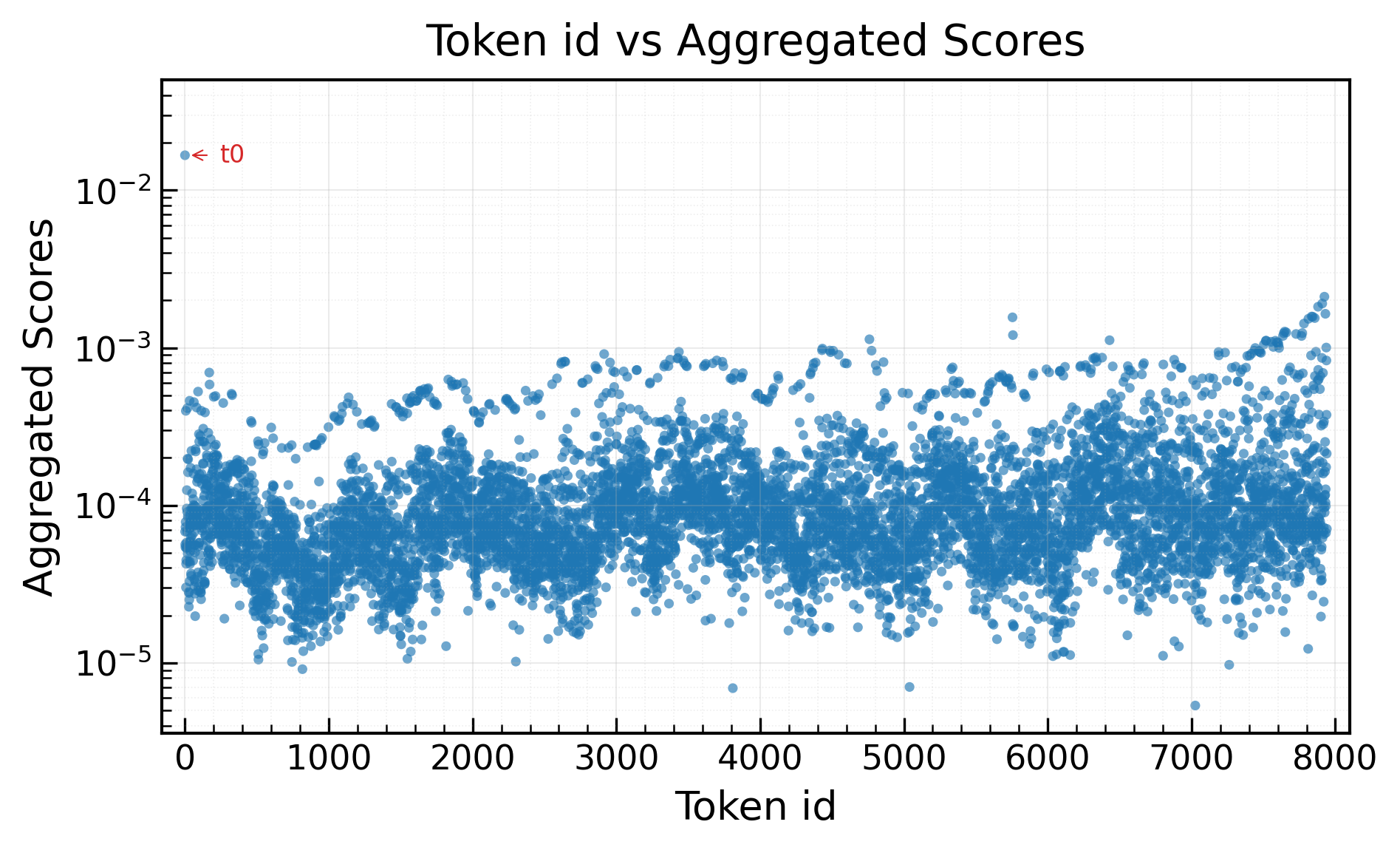}
        \caption{Log scale Y-axis.}
        \label{fig:layer0-head22-attention_log}
    \end{subfigure}

    \caption{Dispersed aggregated observation-window attention distribution of Layer 0, Head 22 in Llama3.1-8B-Instruct. Although the maximum score occurs at token 0, many other historical tokens receive relatively higher scores across the sequence.}

\end{figure}

\paragraph{``Dispersed Attention Distribution''}
\label{para:Dispersed_Attn}
This pattern means that the current attention head assigns relatively similar attention scores to historical tokens, with attention scores dispersed across a wide range of historical tokens. For Layer0 Head22 in Figure~\ref{fig:layer0-head22-attention}, except for \texttt{<bos>}, tokens that originally receive low scores under the concentrated pattern obtain relatively higher attention throughout the sequence. The log-scale view in Figure~\ref{fig:layer0-head22-attention_log} further highlights this change. This dispersed pattern indicates higher attention mass coverage difficulty and therefore motivates a larger head-wise KV cache budget.

These two types of heads correspond to the ``difficult-to-accumulate attention'' and ``quick-to-accumulate attention'' heads described in Ada-KV~\cite{feng2026ada}. 

\subsection{Validation of the Max-Attention Proxy}
\label{appsubsec:Validation_of_the_Max-Attention_Proxy}

We take the observation-window attention of the $h$-th head in the $l$-th layer and compute it as follows:

From the attention-score matrix computed by the standard attention mechanism:

\[A=\mathrm{softmax}\left(\frac{Q\cdot K^\top}{\sqrt{d_{key}}}\right)\]

\[A\in\mathbb{R}^{L\times L}\]

where $L$ denotes the input sequence length.

The observation-window attention score matrix corresponds to the lower-left part of $A$ (where the observation window has length $W$):

\[A'\in\mathbb{R}^{W\times (L-W)}\]

Averaging $A'$ along its columns yields the aggregated observation-window attention score:

\[a_i=\frac{1}{W}\sum_{k=0}^{W-1}A'_{ki},\qquad i\in[0,L-W-1]\]

The mean of all elements in vector $a$ is:
\[\bar{a}=\frac{\sum_{i=0}^{L-W-1}a_i}{L-W}\]

The variance of all elements in vector $a$ is:

\[
\begin{aligned}
\sigma^2
&=\frac{1}{L-W}\sum_{i=0}^{L-W-1}(a_i-\bar{a})^2 \\
&=\frac{1}{L-W}\sum_{i=0}^{L-W-1}a_i^2-\bar{a}^2
\end{aligned}
\]

$\bar{a}^2$ is on the order of $\frac{1}{(L-W)^2}$, which is a higher-order small term compared with: 
\[\frac{1}{L-W}\sum_{i=0}^{L-W-1}a_i^2\]

Therefore:

\[\sigma^2\approx\frac{1}{L-W}\sum_{i=0}^{L-W-1}a_i^2\]

The corresponding standard deviation is:

\[\sigma=\sqrt{\frac{1}{L-W}\sum_{i=0}^{L-W-1}a_i^2}\]

The aggregated observation-window attention score vector $a_i$ is non-negative, has a bounded sum, and is typically sparse because it results from softmax normalization. Its second-order statistics are therefore strongly controlled by the maximum score. We can thus hypothesize a linear relationship between $\sigma$ and the maximum attention score $a_{\max}$:

\[\sigma\approx\frac{1}{\sqrt{L-W}}a_{\max}\]

\begin{figure*}[!t]
    \centering

    \begin{subfigure}[t]{0.24\linewidth}
        \centering
        \includegraphics[width=\linewidth,height=0.27\textheight,keepaspectratio]{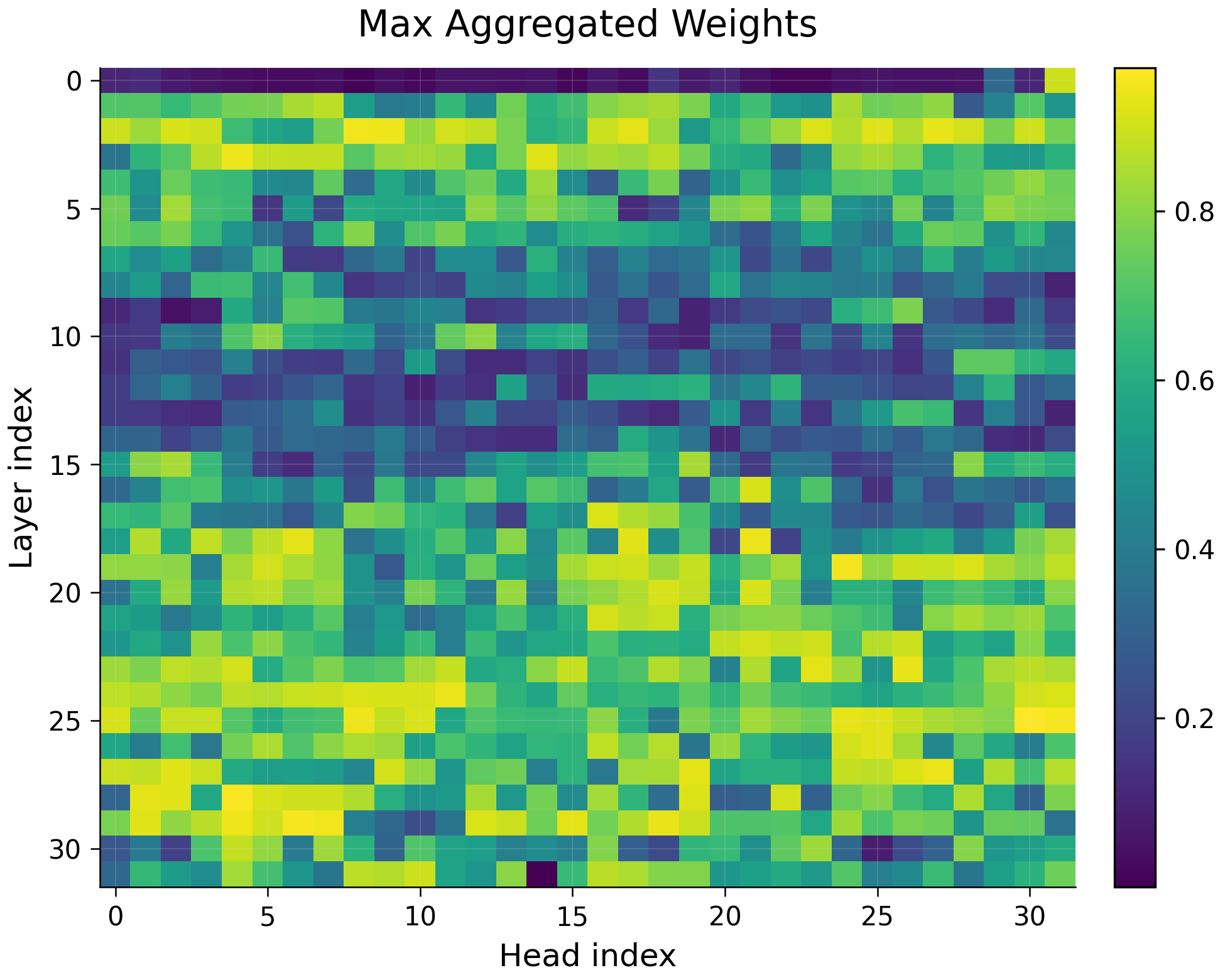}
        \caption{Max.}
        \label{fig:Llama8b_w8_max}
    \end{subfigure}
    \hfill
    \begin{subfigure}[t]{0.24\linewidth}
        \centering
        \includegraphics[width=\linewidth,height=0.27\textheight,keepaspectratio]{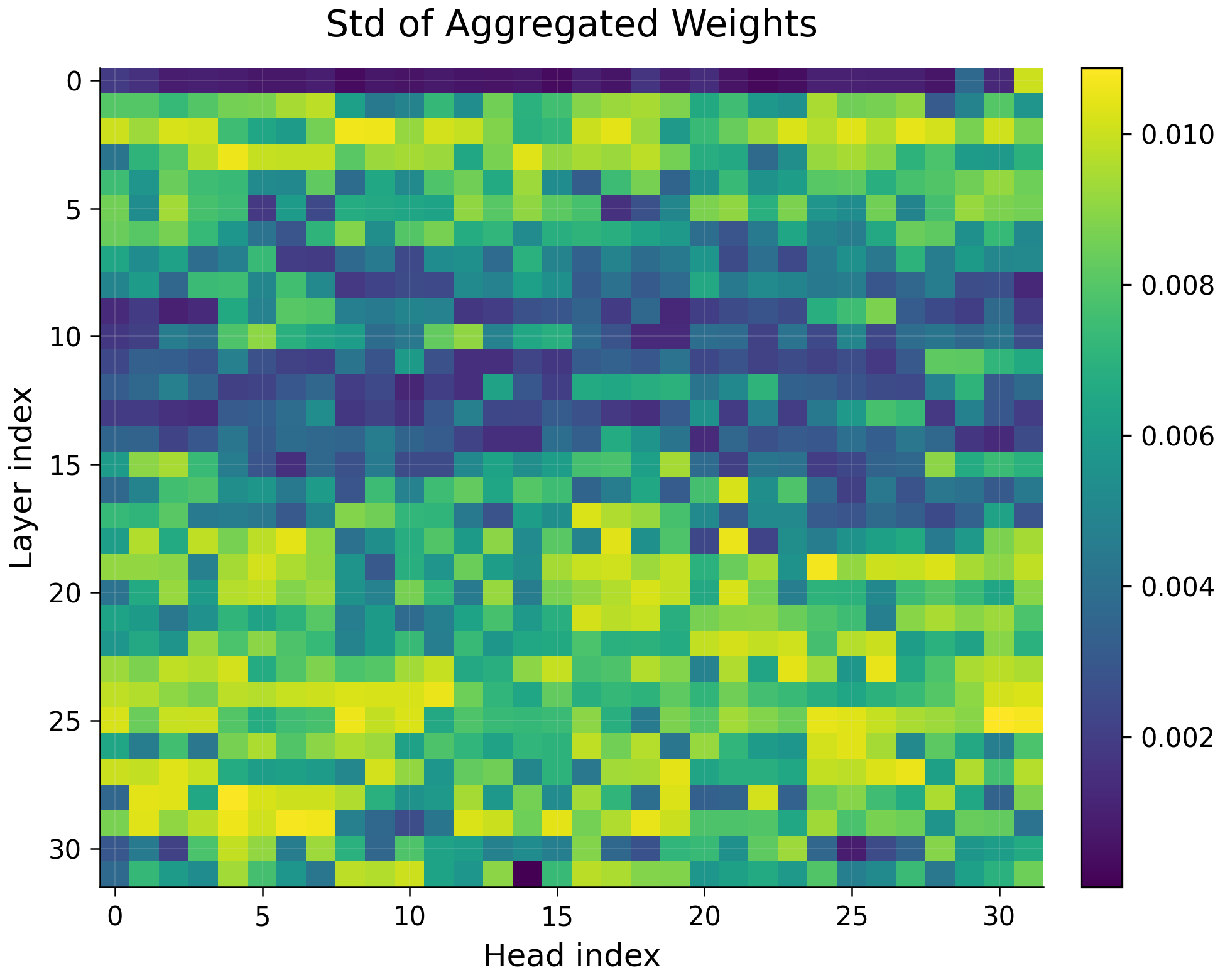}
        \caption{Std.}
        \label{fig:Llama8b_w8_std}
    \end{subfigure}
    \hfill
    \begin{subfigure}[t]{0.24\linewidth}
        \centering
        \includegraphics[width=\linewidth,height=0.27\textheight,keepaspectratio]{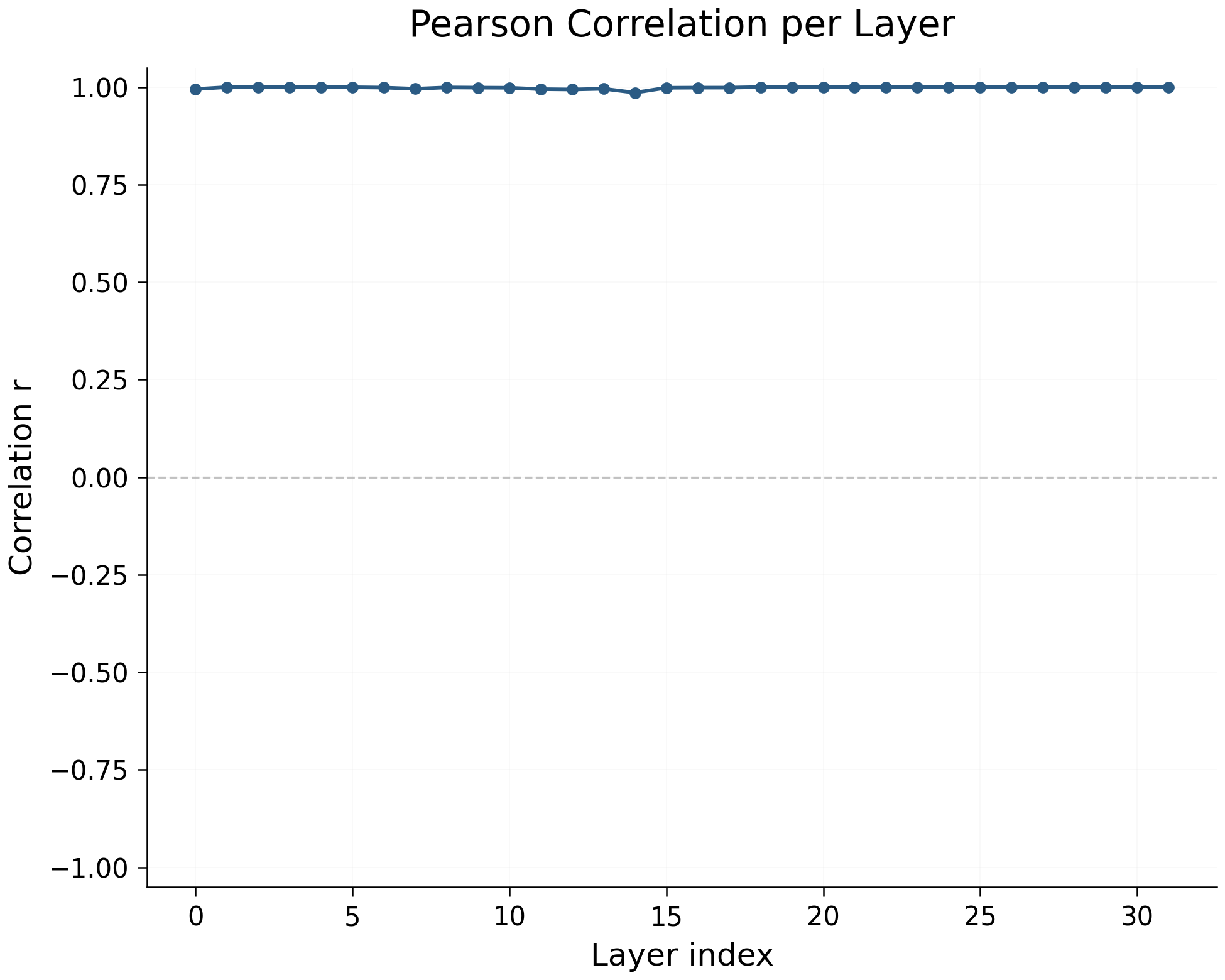}
        \caption{Pearson \(r\).}
        \label{fig:Llama8b_w8_r}
    \end{subfigure}
    \hfill
    \begin{subfigure}[t]{0.24\linewidth}
        \centering
        \includegraphics[width=\linewidth,height=0.27\textheight,keepaspectratio]{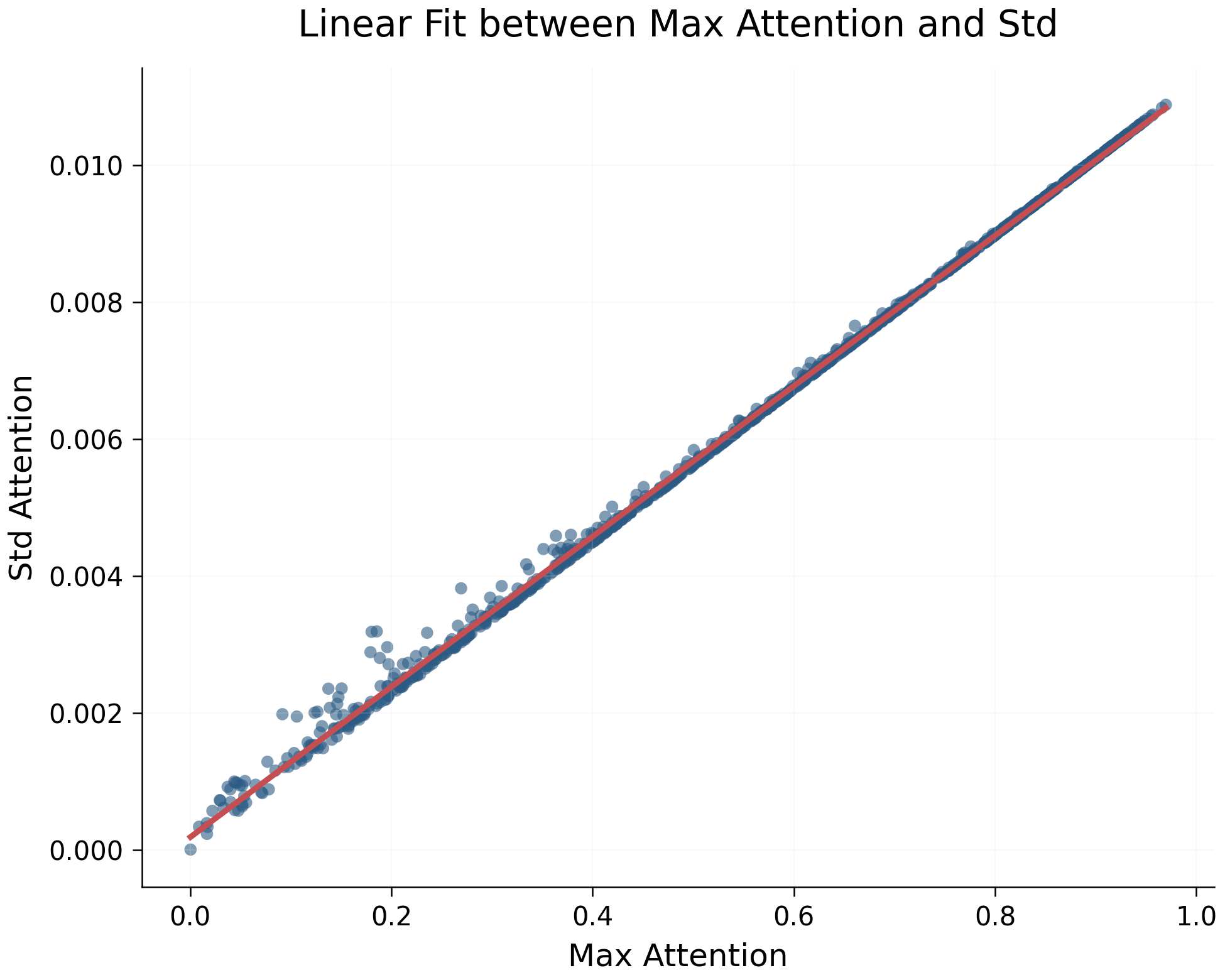}
        \caption{Linear fit.}
        \label{fig:Llama8b_w8_lin}
    \end{subfigure}

    \caption{Correlation analysis on NarrativeQA with Llama3.1-8B-Instruct with \(W=8\).}
    \label{fig:Llama8b_w8_all}
\end{figure*}

\begin{figure*}[!t]
    \centering

    \begin{subfigure}[t]{0.24\linewidth}
        \centering
        \includegraphics[width=\linewidth,height=0.27\textheight,keepaspectratio]{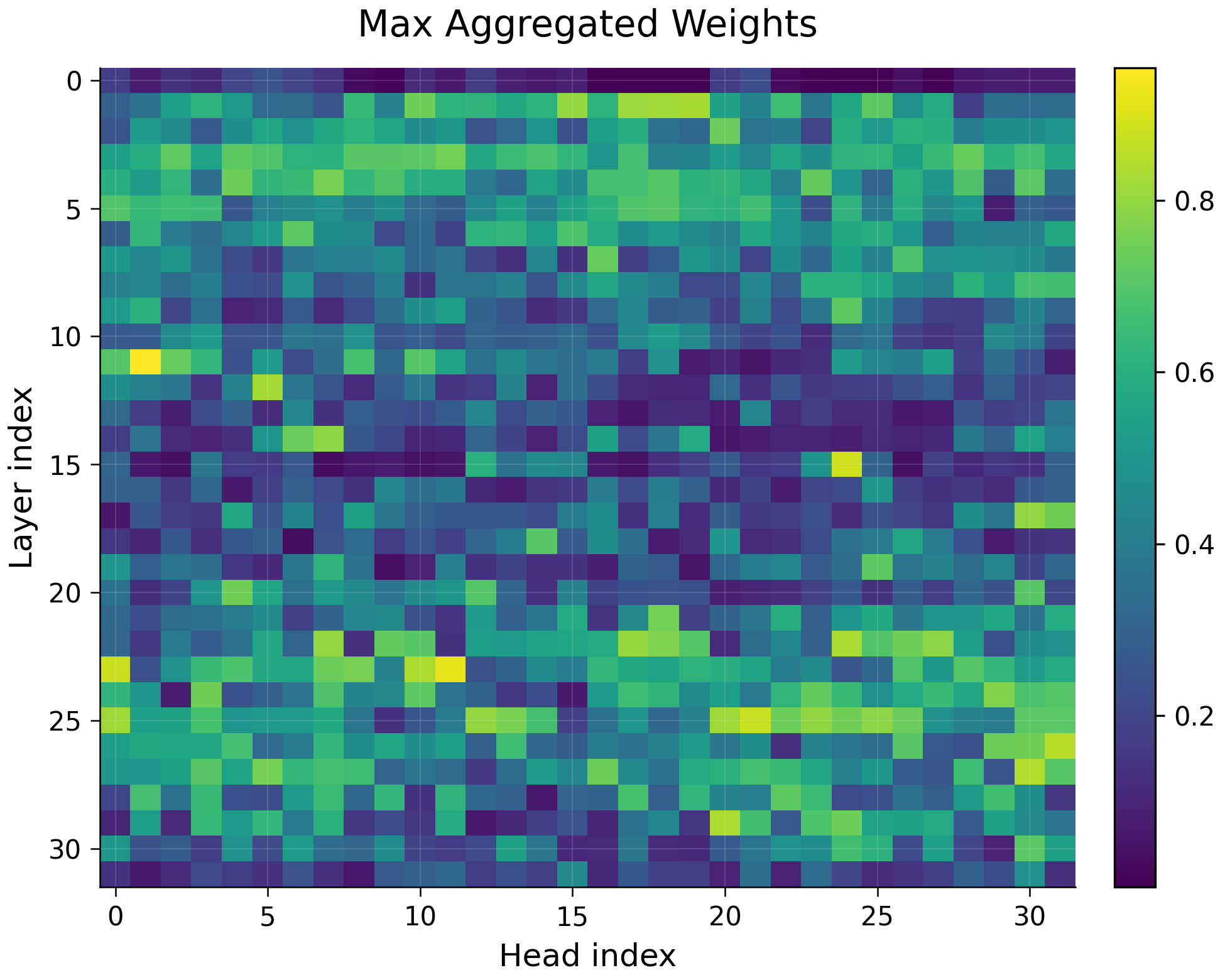}
        \caption{Max.}
        \label{fig:mistral_w8_max}
    \end{subfigure}
    \hfill
    \begin{subfigure}[t]{0.24\linewidth}
        \centering
        \includegraphics[width=\linewidth,height=0.27\textheight,keepaspectratio]{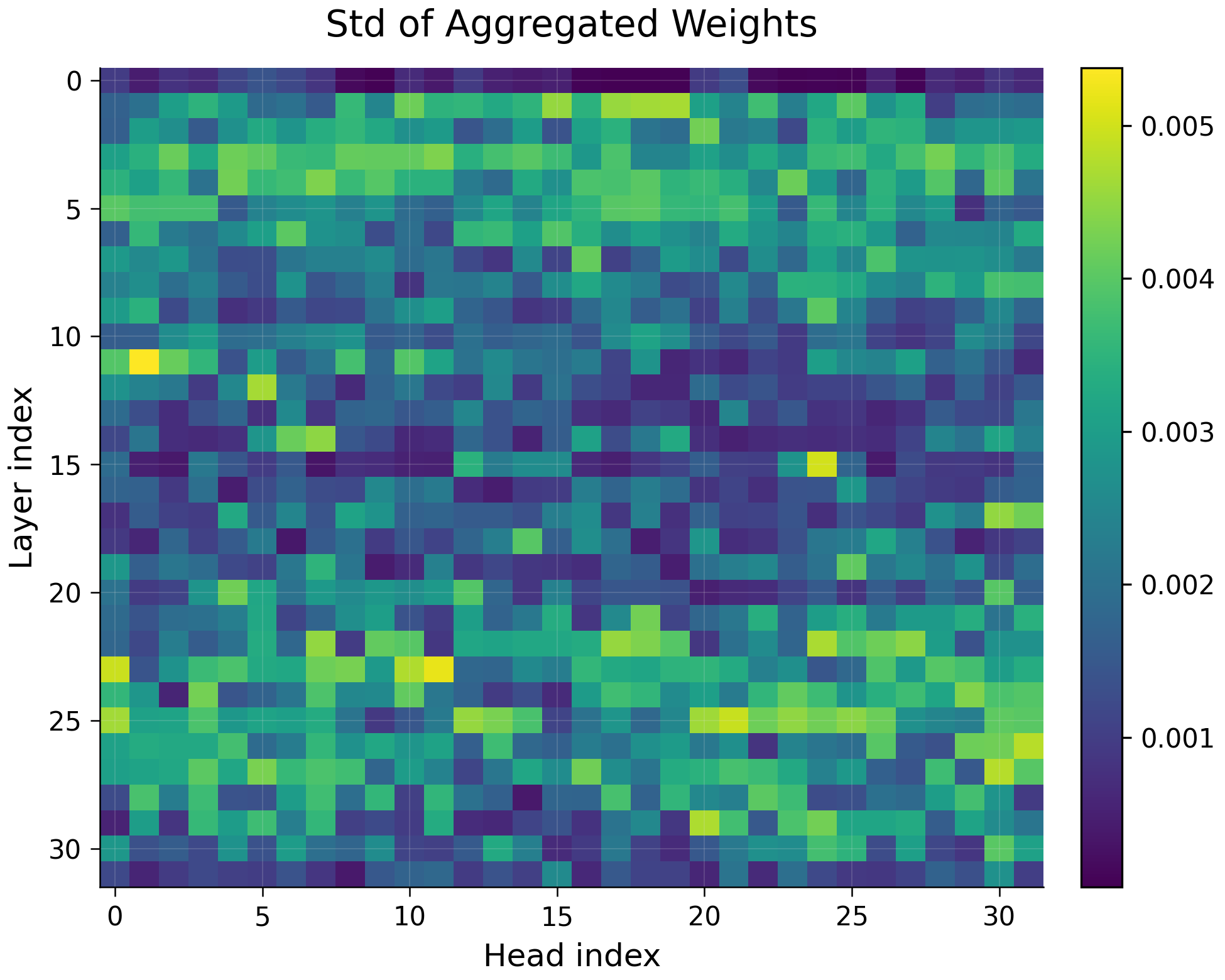}
        \caption{Std.}
        \label{fig:mistral_w8_std}
    \end{subfigure}
    \hfill
    \begin{subfigure}[t]{0.24\linewidth}
        \centering
        \includegraphics[width=\linewidth,height=0.27\textheight,keepaspectratio]{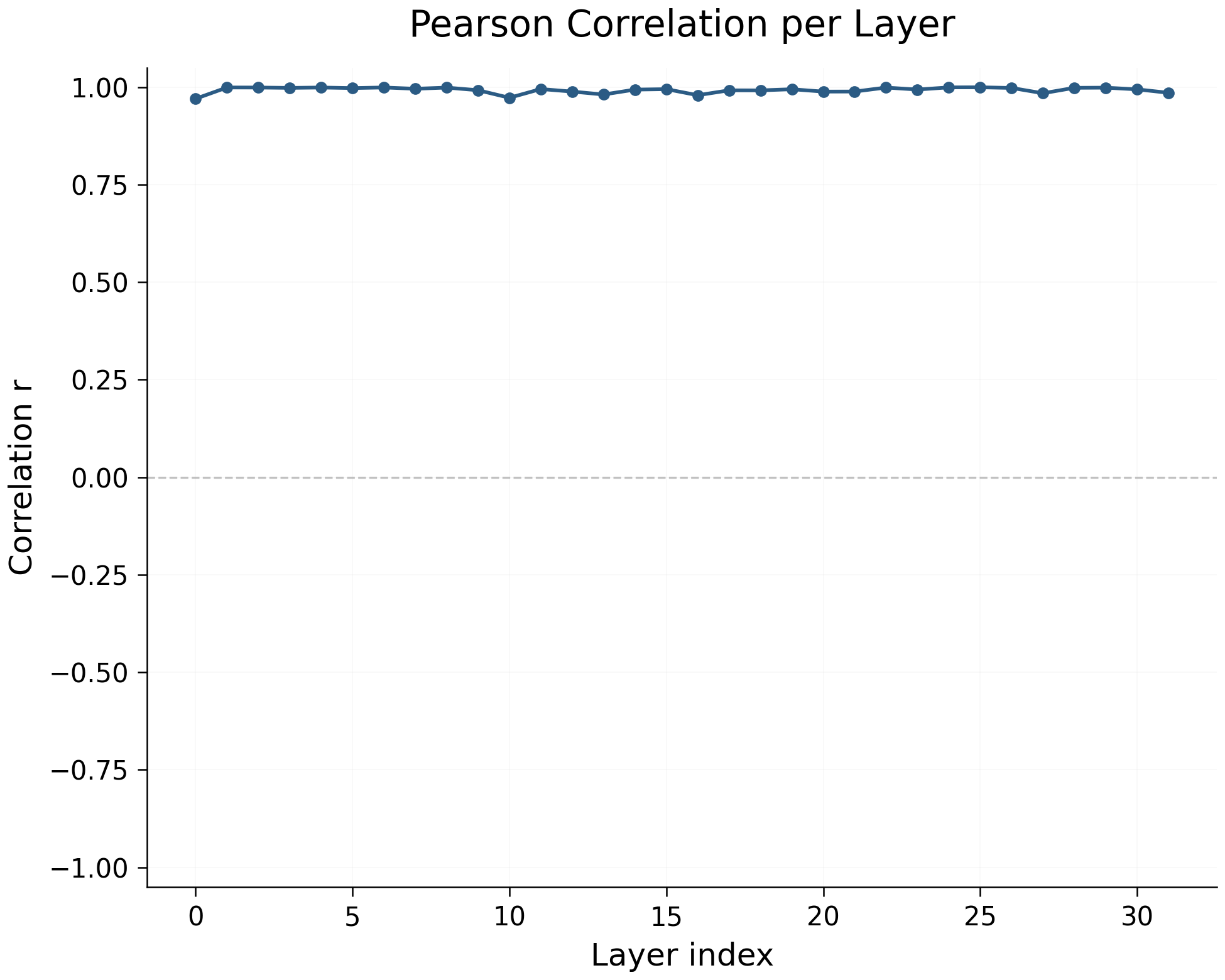}
        \caption{Pearson \(r\).}
        \label{fig:mistral_w8_r}
    \end{subfigure}
    \hfill
    \begin{subfigure}[t]{0.24\linewidth}
        \centering
        \includegraphics[width=\linewidth,height=0.27\textheight,keepaspectratio]{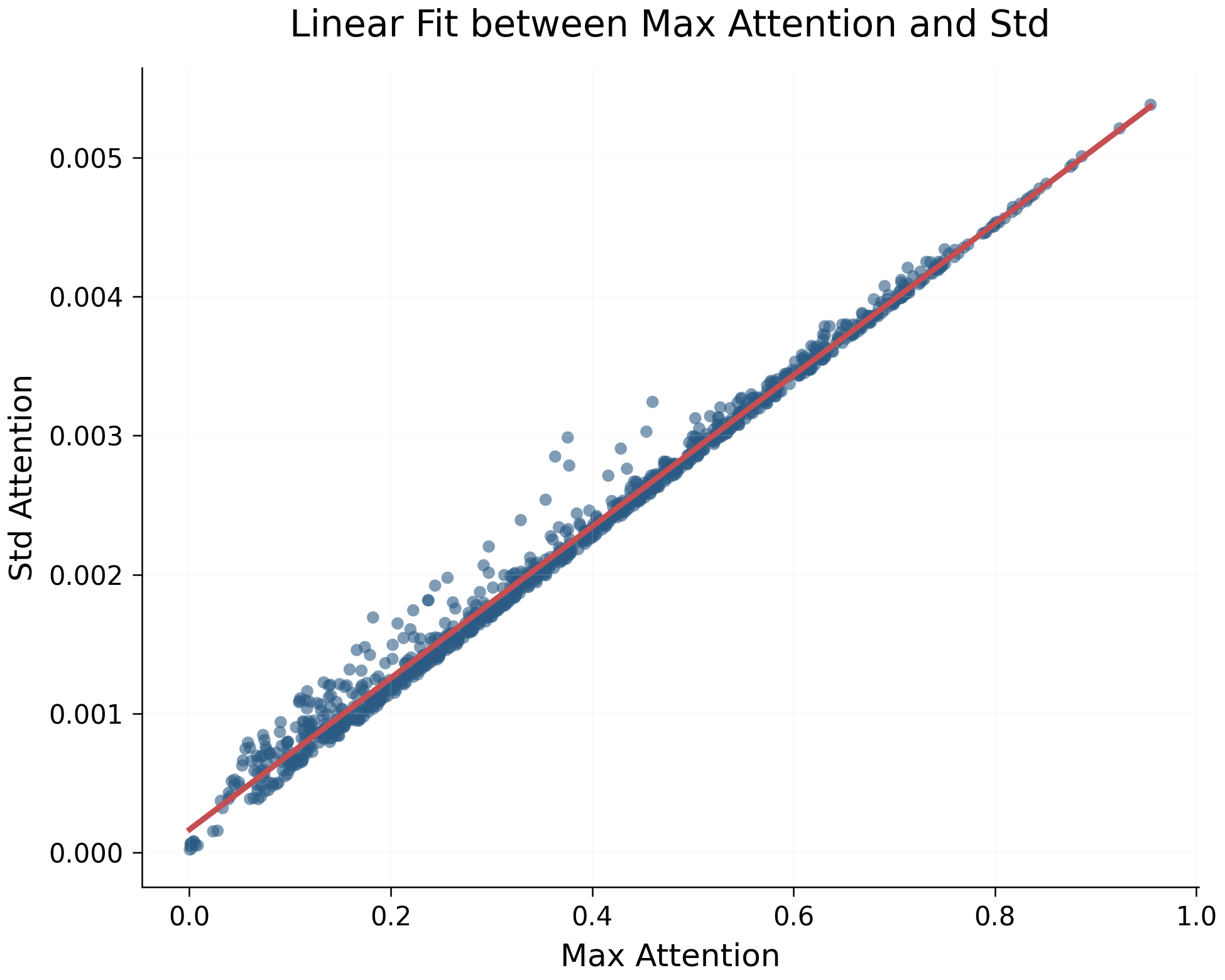}
        \caption{Linear fit.}
        \label{fig:mistral_w8_lin}
    \end{subfigure}

    \caption{Correlation analysis on Mistral-7B-Instruct-v0.3 with \(W=8\).}
    \label{fig:mistral_w8_all}
\end{figure*}

\begin{figure*}[!t]
    \centering

    \begin{subfigure}[t]{0.24\linewidth}
        \centering
        \includegraphics[width=\linewidth,height=0.27\textheight,keepaspectratio]{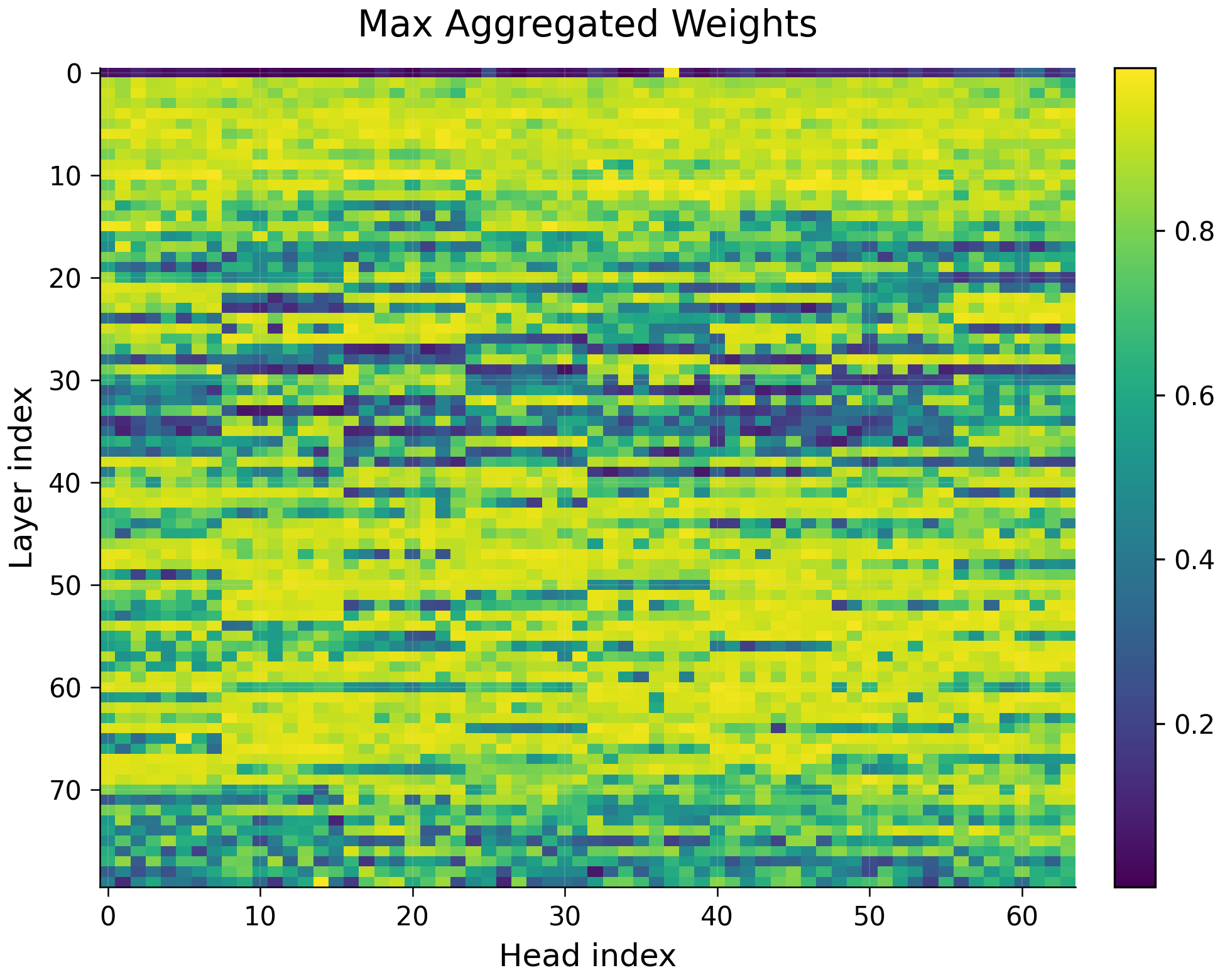}
        \caption{Max.}
        \label{fig:Llama70b_w8_max}
    \end{subfigure}
    \hfill
    \begin{subfigure}[t]{0.24\linewidth}
        \centering
        \includegraphics[width=\linewidth,height=0.27\textheight,keepaspectratio]{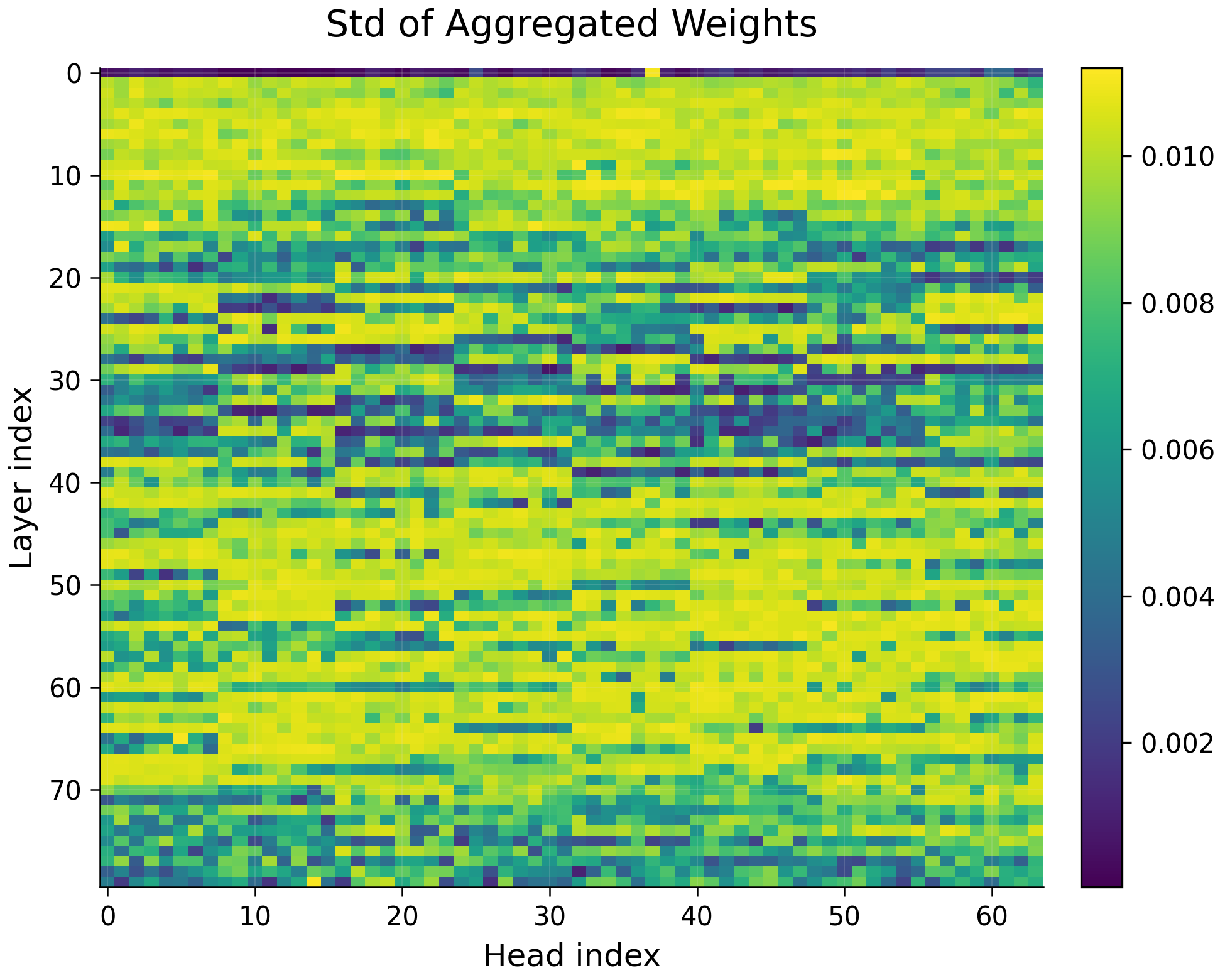}
        \caption{Std.}
        \label{fig:Llama70b_w8_std}
    \end{subfigure}
    \hfill
    \begin{subfigure}[t]{0.24\linewidth}
        \centering
        \includegraphics[width=\linewidth,height=0.27\textheight,keepaspectratio]{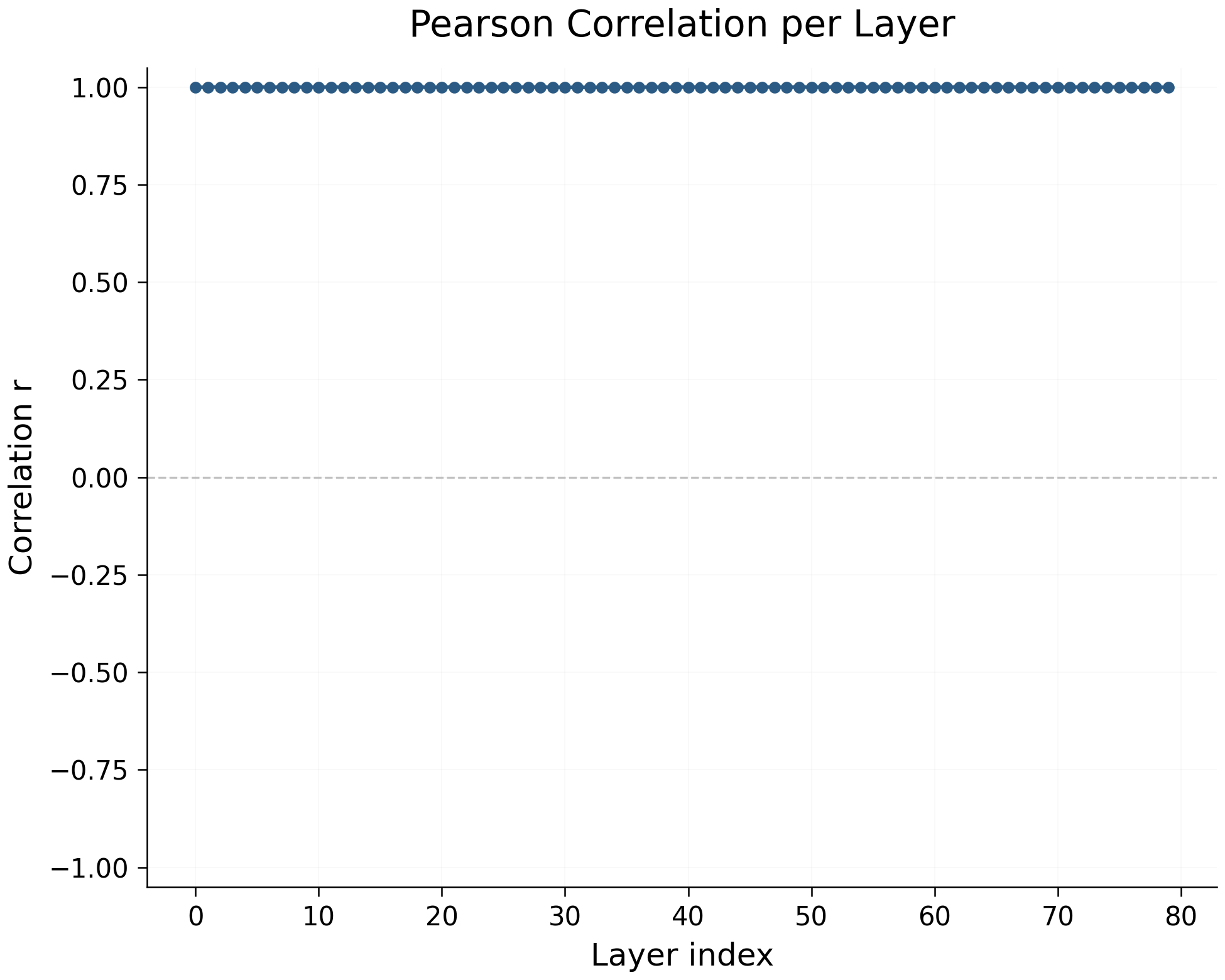}
        \caption{Pearson \(r\).}
        \label{fig:Llama70b_w8_r}
    \end{subfigure}
    \hfill
    \begin{subfigure}[t]{0.24\linewidth}
        \centering
        \includegraphics[width=\linewidth,height=0.27\textheight,keepaspectratio]{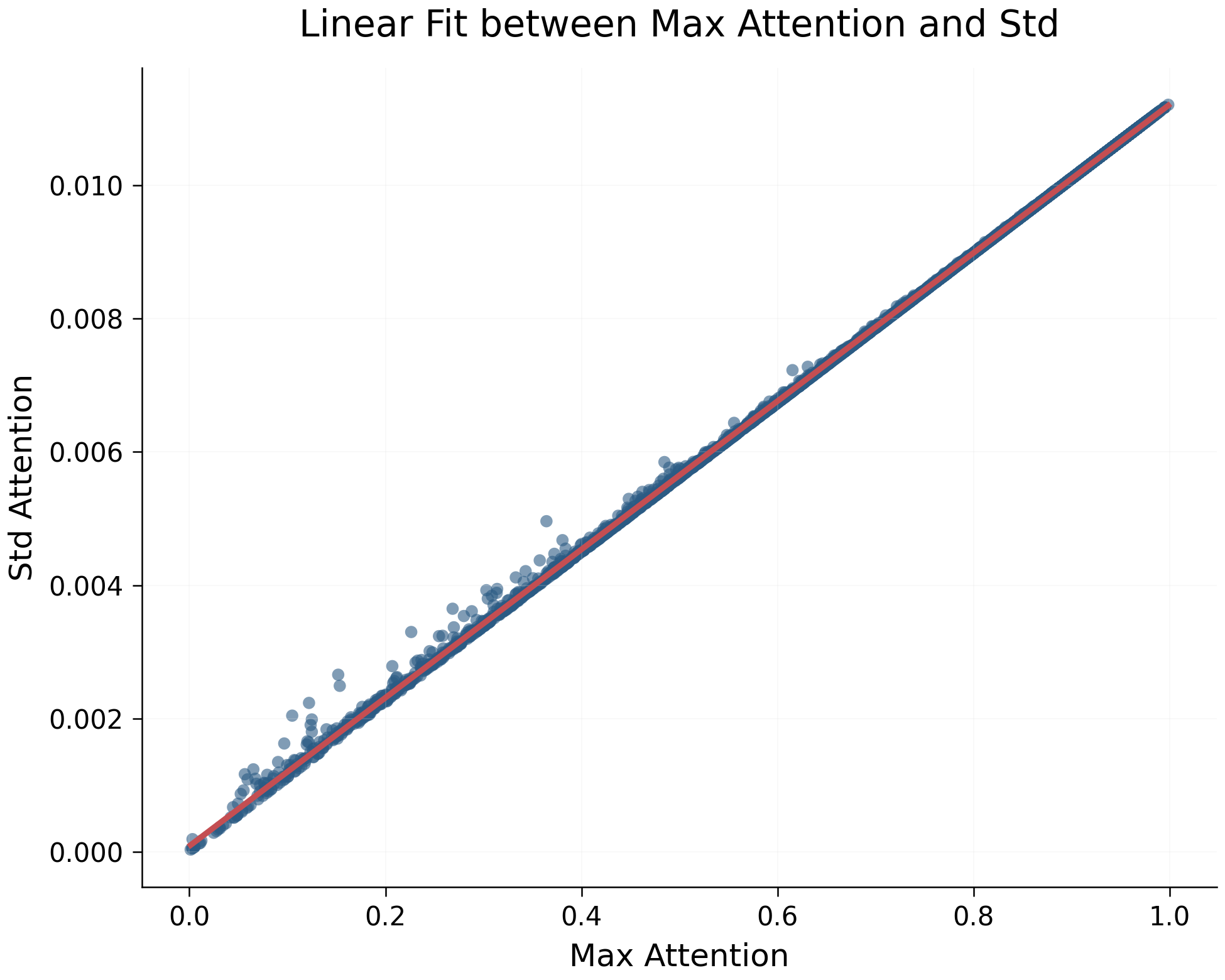}
        \caption{Linear fit.}
        \label{fig:Llama70b_w8_lin}
    \end{subfigure}

  \caption{Correlation analysis on Llama3.1-70B-Instruct with \(W=8\).}
    \label{fig:Llama70b_w8_all}
\end{figure*}

\begin{figure*}[!t]
    \centering

    \begin{subfigure}[t]{0.24\linewidth}
        \centering
        \includegraphics[width=\linewidth,height=0.27\textheight,keepaspectratio]{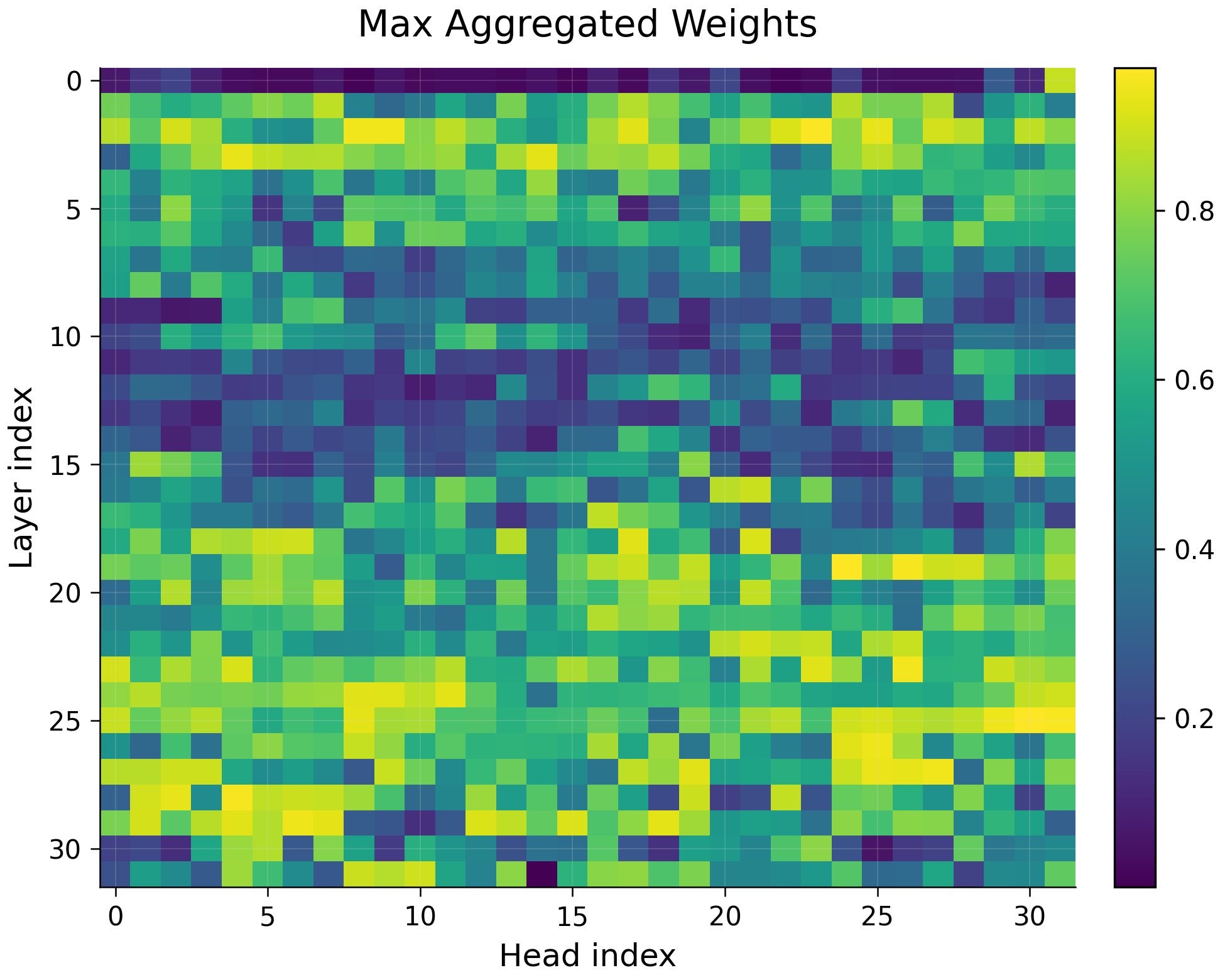}
        \caption{Max.}
        \label{fig:Llama8b_w4_max}
    \end{subfigure}
    \hfill
    \begin{subfigure}[t]{0.24\linewidth}
        \centering
        \includegraphics[width=\linewidth,height=0.27\textheight,keepaspectratio]{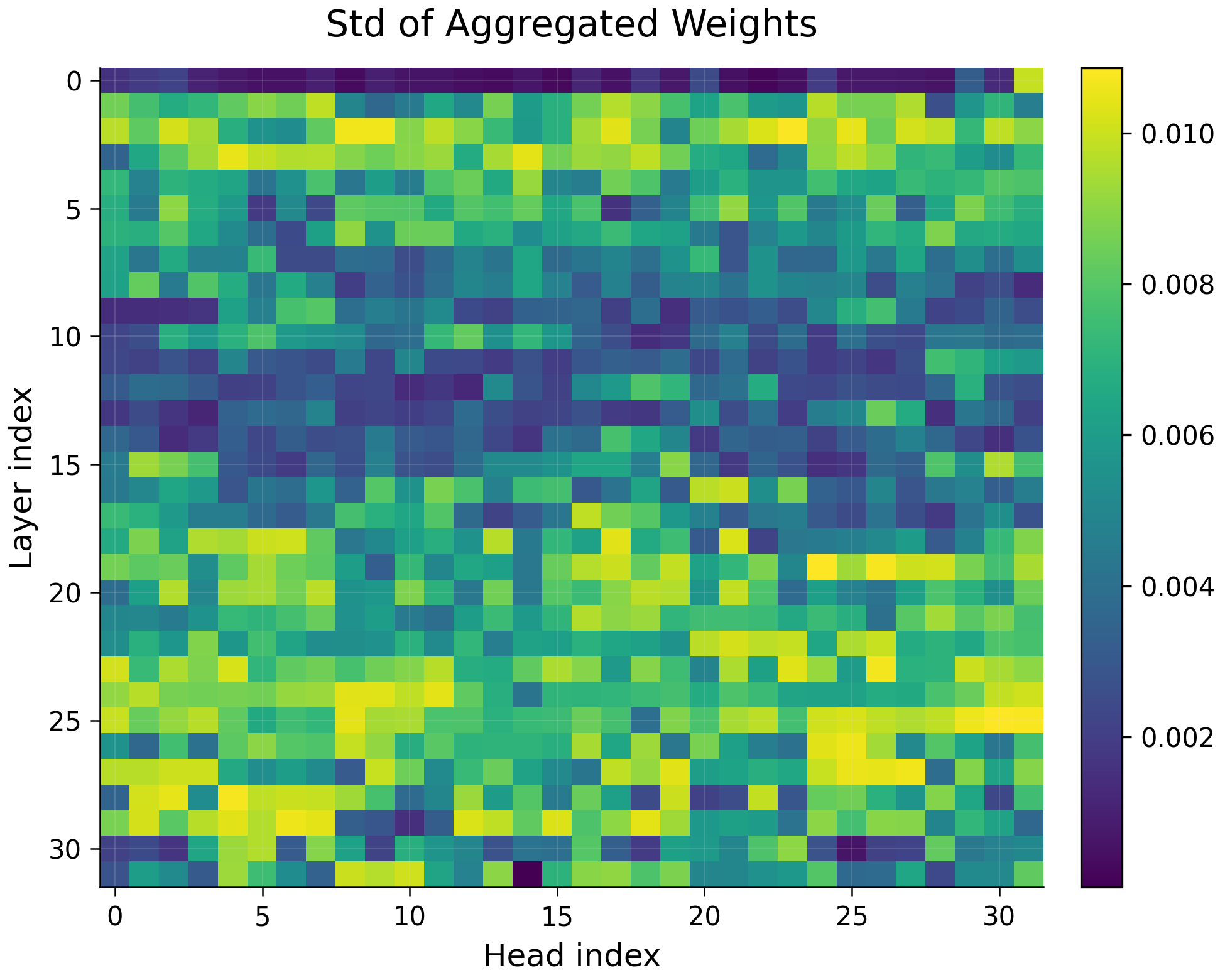}
        \caption{Std.}
        \label{fig:Llama8b_w4_std}
    \end{subfigure}
    \hfill
    \begin{subfigure}[t]{0.24\linewidth}
        \centering
        \includegraphics[width=\linewidth,height=0.27\textheight,keepaspectratio]{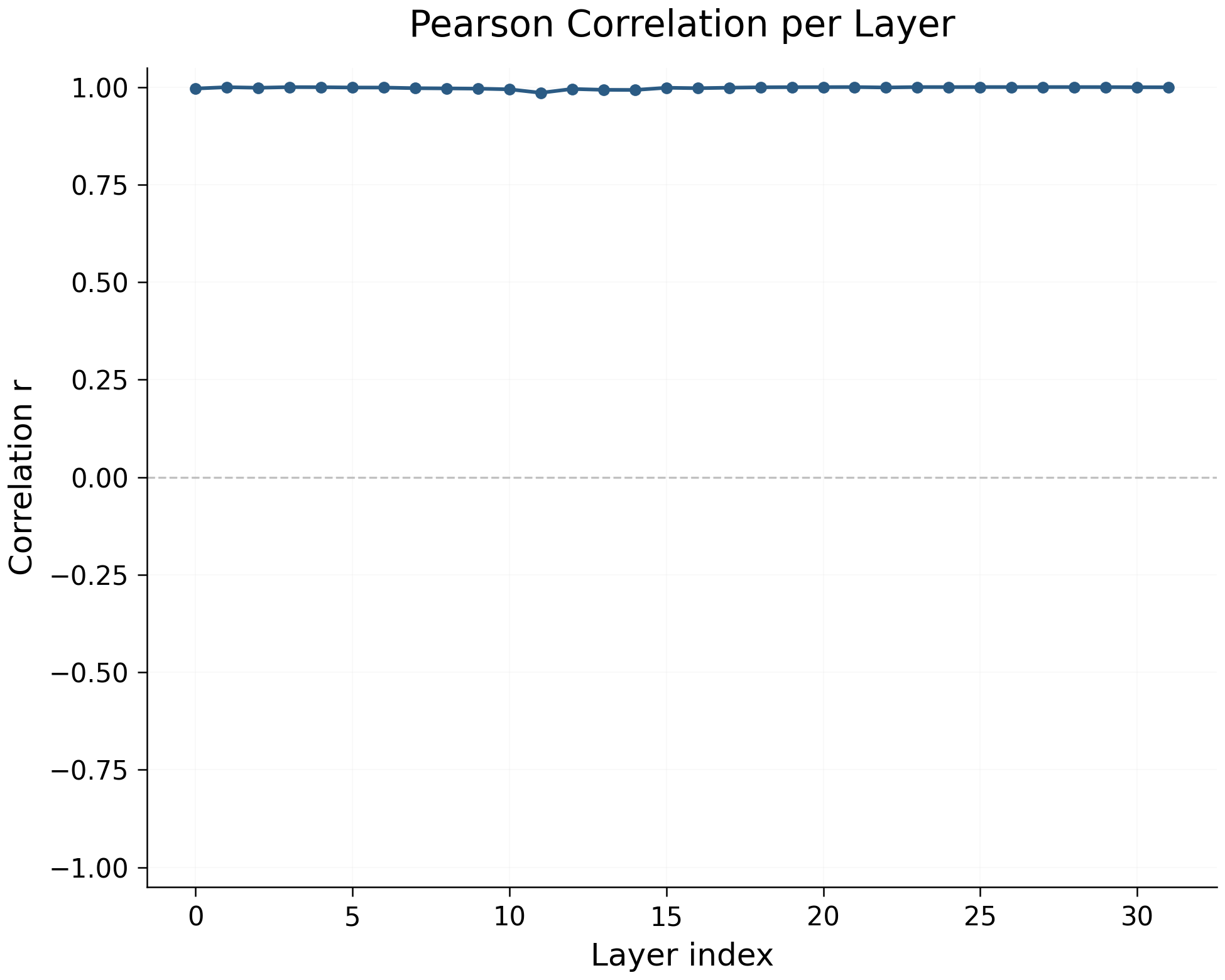}
        \caption{Pearson \(r\).}
        \label{fig:Llama8b_w4_r}
    \end{subfigure}
    \hfill
    \begin{subfigure}[t]{0.24\linewidth}
        \centering
        \includegraphics[width=\linewidth,height=0.27\textheight,keepaspectratio]{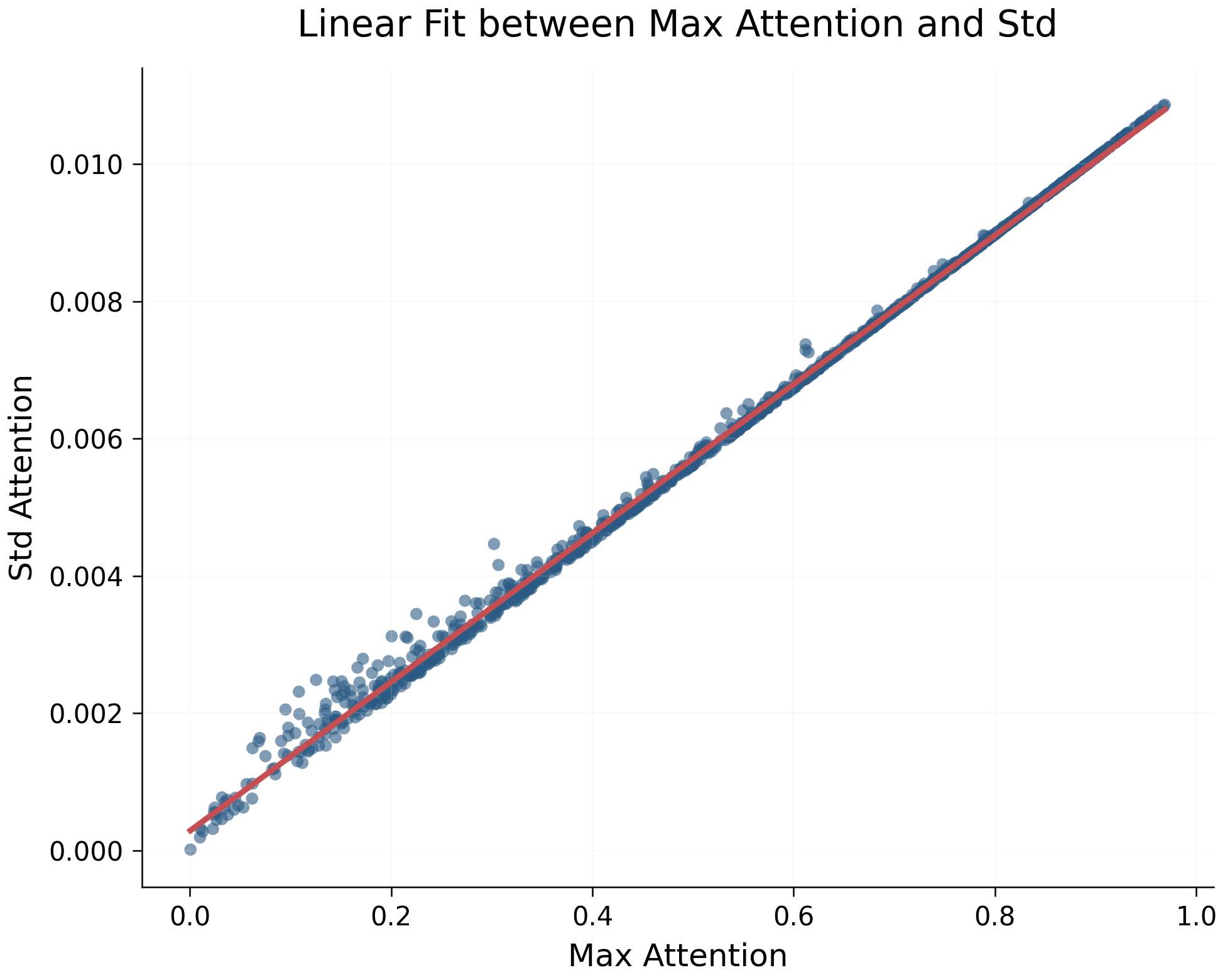}
        \caption{Linear fit.}
        \label{fig:Llama8b_w4_lin}
    \end{subfigure}

\caption{Correlation analysis on Llama3.1-8B-Instruct with \(W=4\). }
    \label{fig:Llama8b_w4_all}
\end{figure*}

\begin{figure*}[!t]
    \centering

    \begin{subfigure}[t]{0.24\linewidth}
        \centering
        \includegraphics[width=\linewidth,height=0.27\textheight,keepaspectratio]{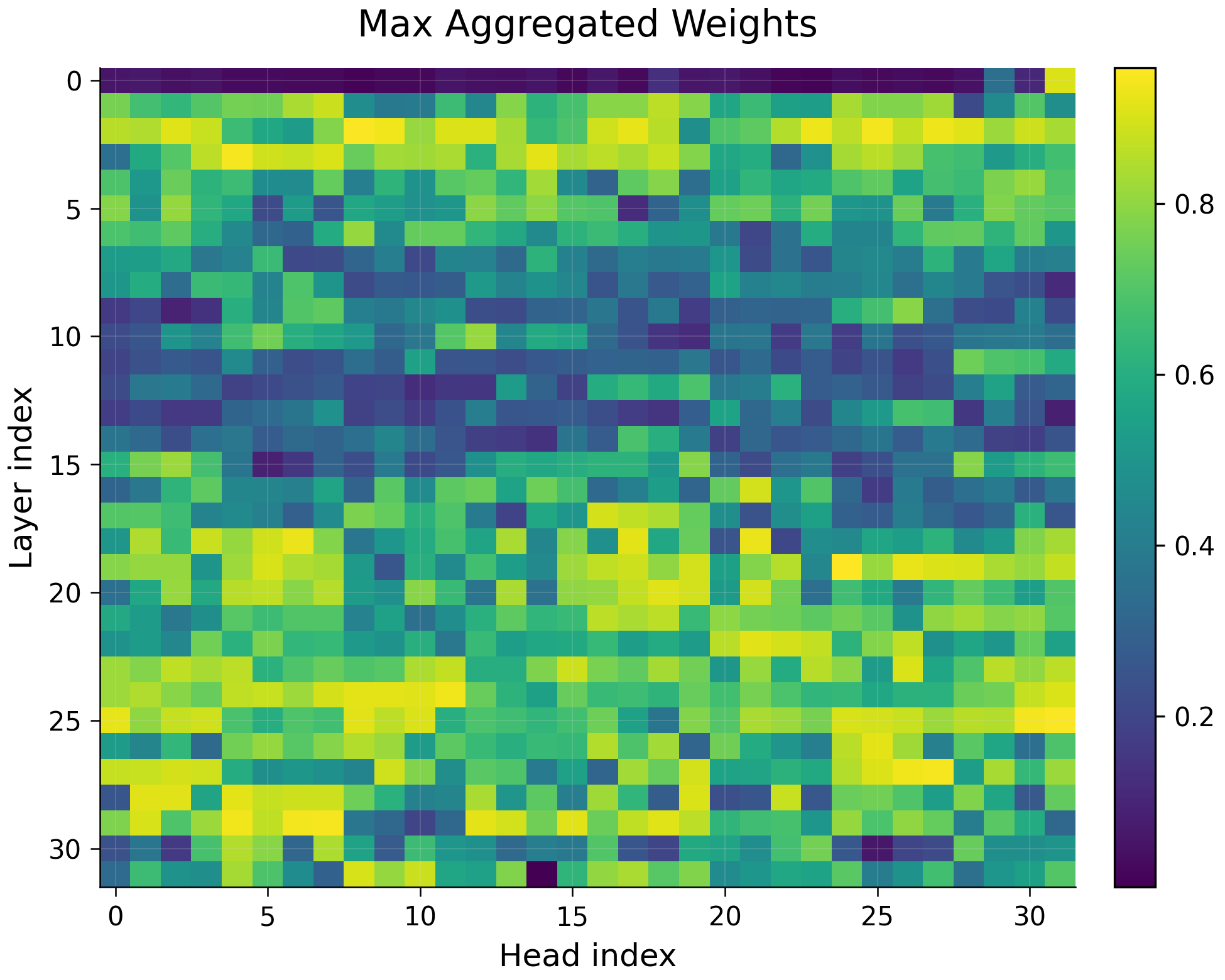}
        \caption{Max.}
        \label{fig:Llama8b_w16_max}
    \end{subfigure}
    \hfill
    \begin{subfigure}[t]{0.24\linewidth}
        \centering
        \includegraphics[width=\linewidth,height=0.27\textheight,keepaspectratio]{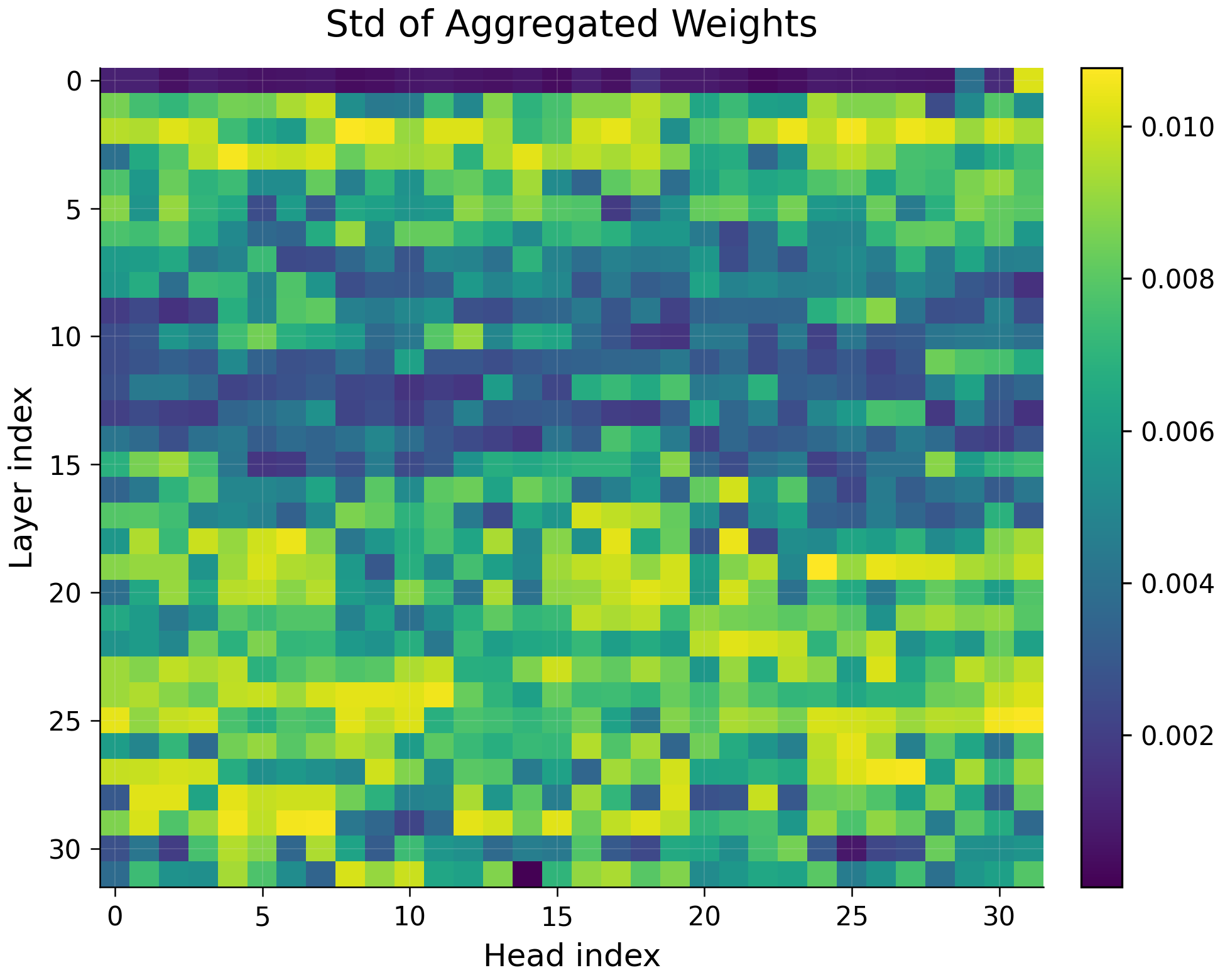}
        \caption{Std.}
        \label{fig:Llama8b_w16_std}
    \end{subfigure}
    \hfill
    \begin{subfigure}[t]{0.24\linewidth}
        \centering
        \includegraphics[width=\linewidth,height=0.27\textheight,keepaspectratio]{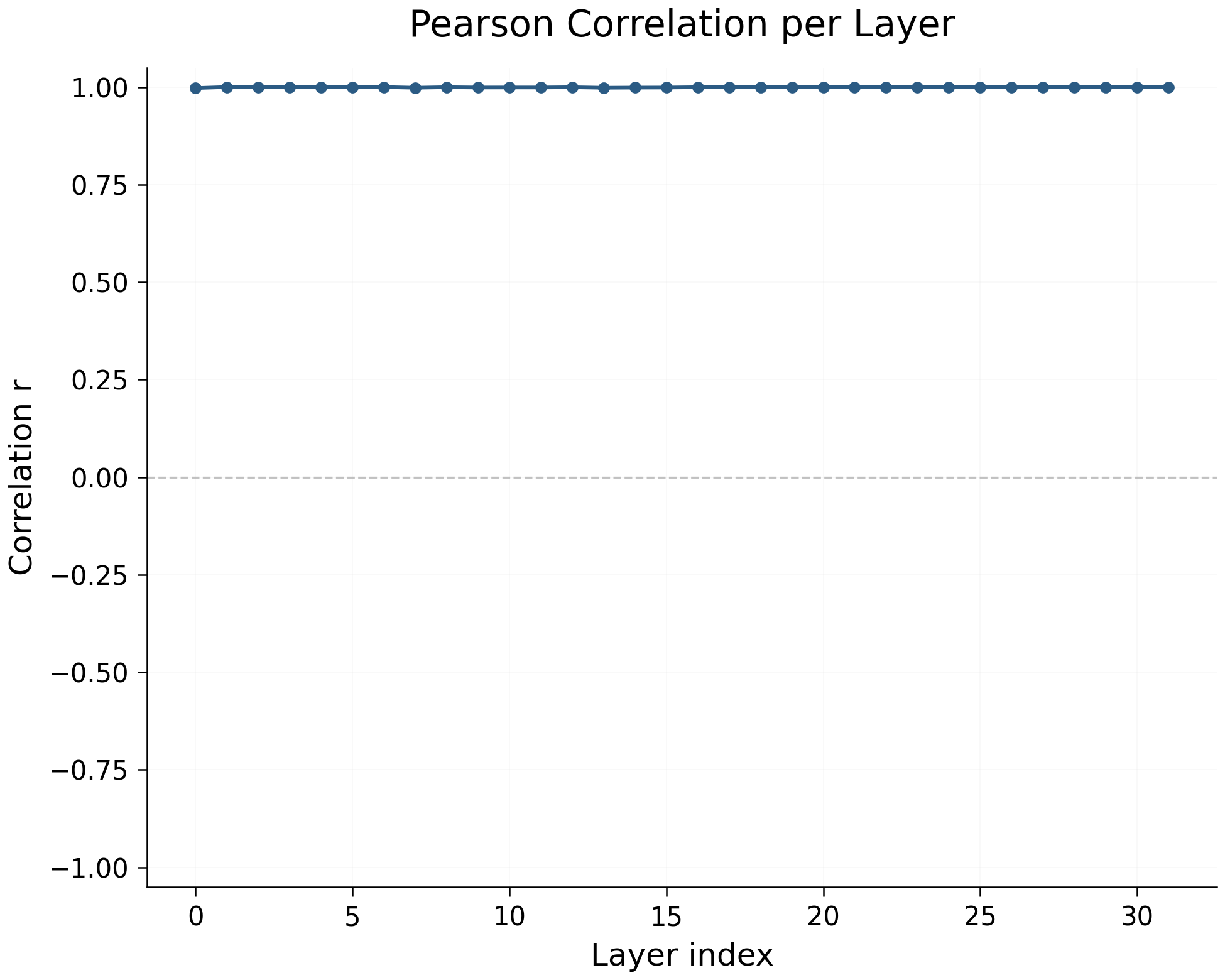}
        \caption{Pearson \(r\).}
        \label{fig:Llama8b_w16_r}
    \end{subfigure}
    \hfill
    \begin{subfigure}[t]{0.24\linewidth}
        \centering
        \includegraphics[width=\linewidth,height=0.27\textheight,keepaspectratio]{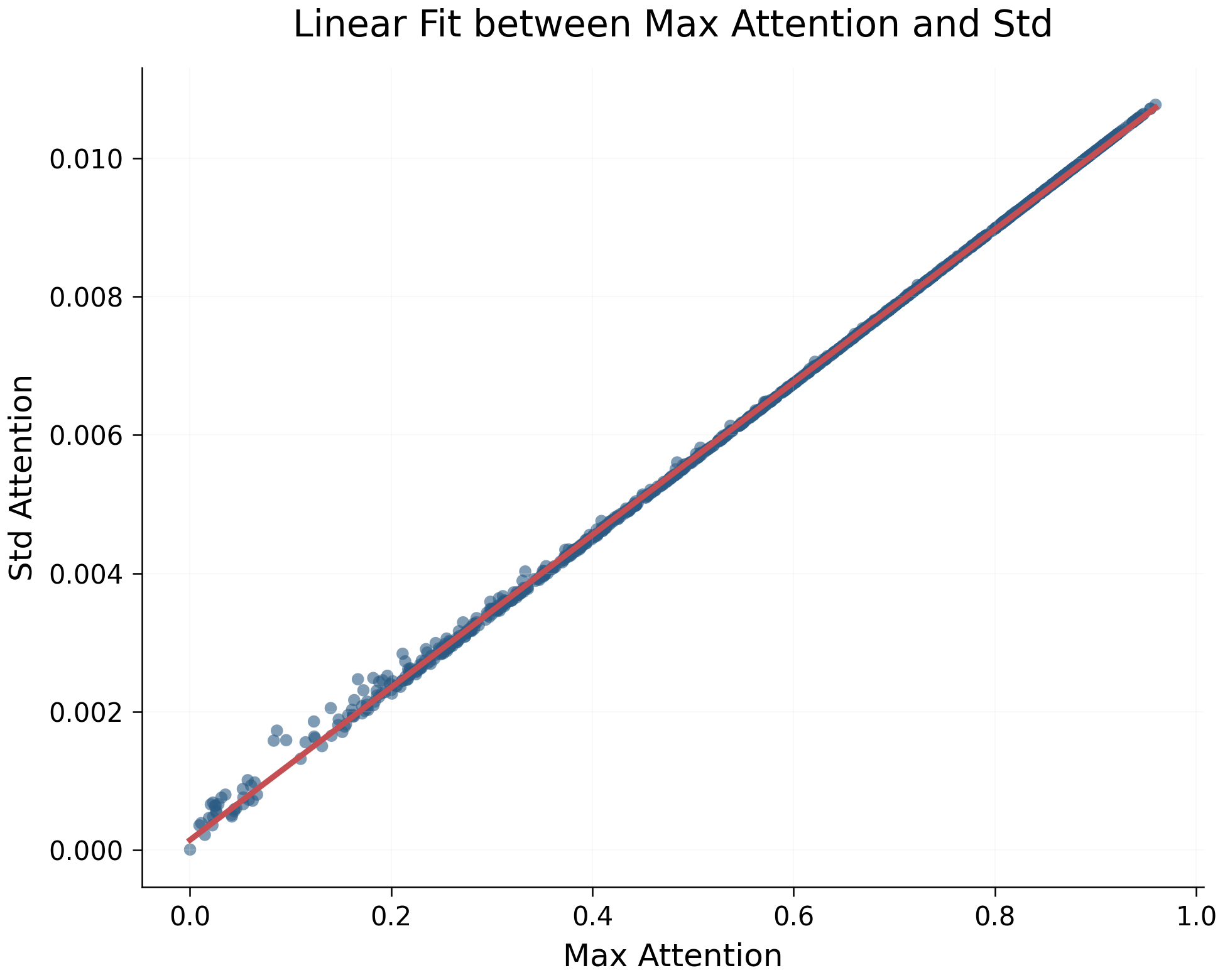}
        \caption{Linear fit.}
        \label{fig:Llama8b_w16_lin}
    \end{subfigure}

\caption{Correlation analysis on Llama3.1-8B-Instruct with \(W=16\).}
    \label{fig:Llama8b_w16_all}
\end{figure*}

\begin{figure*}[!t]
    \centering

    \begin{subfigure}[t]{0.24\linewidth}
        \centering
        \includegraphics[width=\linewidth,height=0.27\textheight,keepaspectratio]{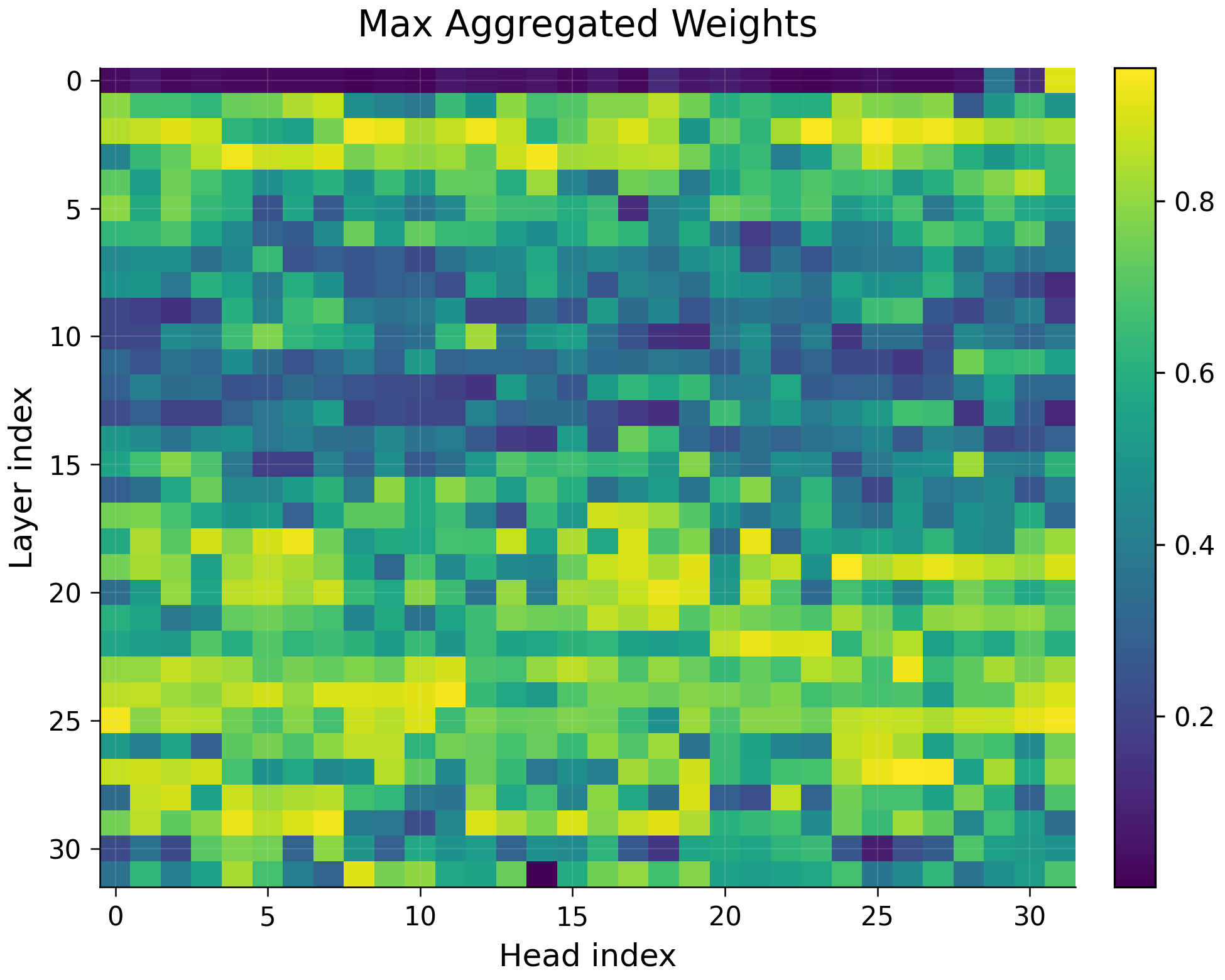}
        \caption{Max.}
        \label{fig:Llama8b_w32_max}
    \end{subfigure}
    \hfill
    \begin{subfigure}[t]{0.24\linewidth}
        \centering
        \includegraphics[width=\linewidth,height=0.27\textheight,keepaspectratio]{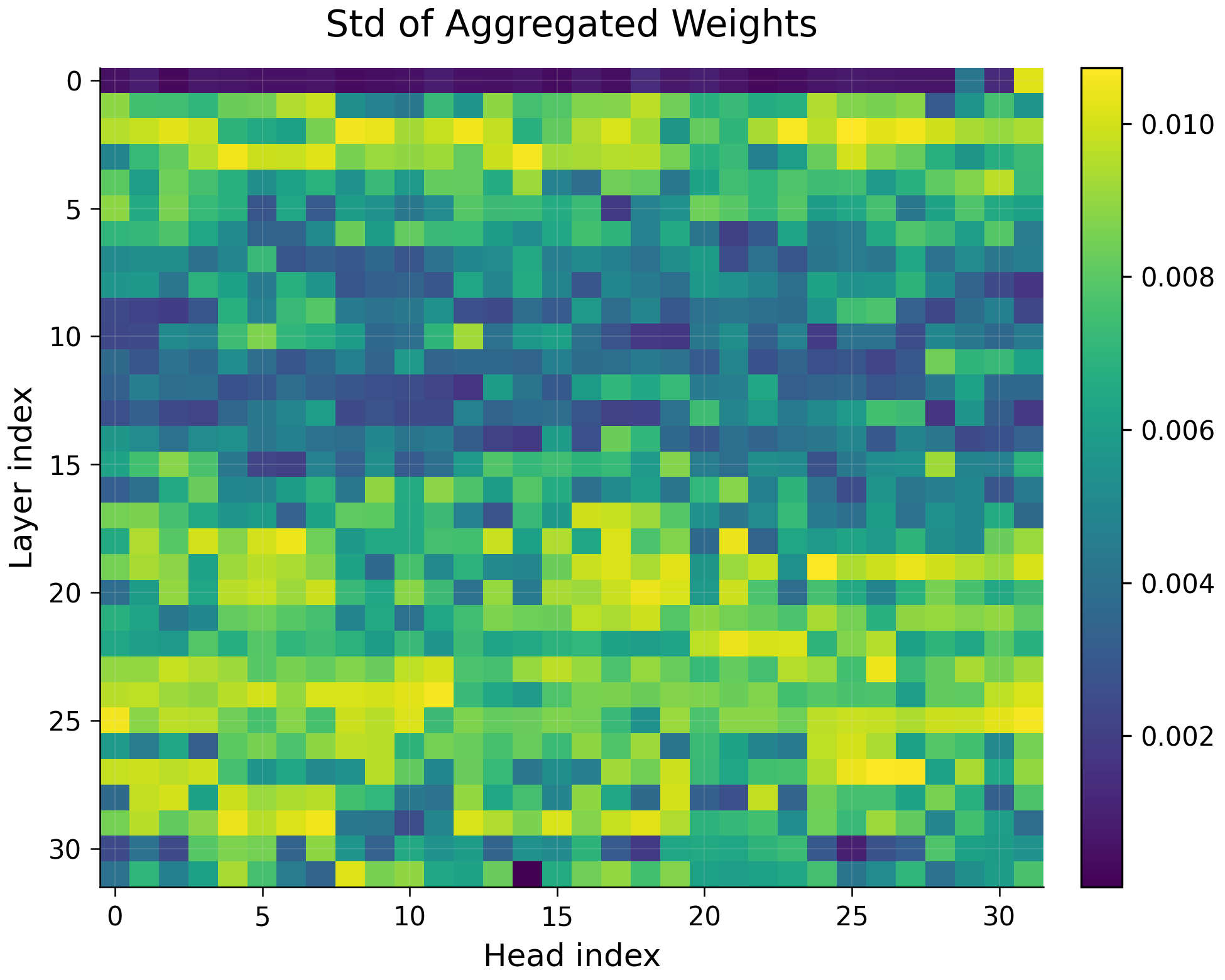}
        \caption{Std.}
        \label{fig:Llama8b_w32_std}
    \end{subfigure}
    \hfill
    \begin{subfigure}[t]{0.24\linewidth}
        \centering
        \includegraphics[width=\linewidth,height=0.27\textheight,keepaspectratio]{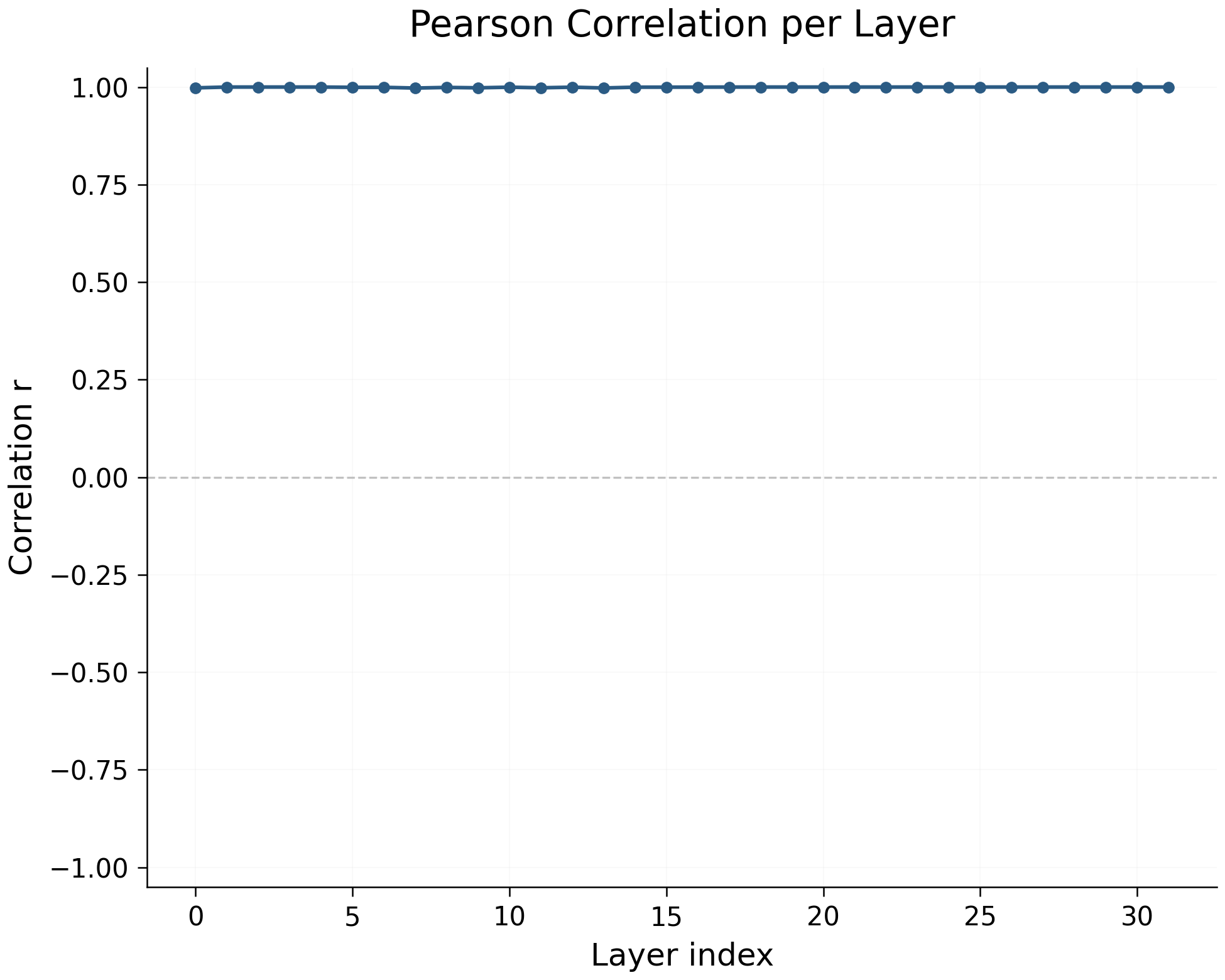}
        \caption{Pearson \(r\).}
        \label{fig:Llama8b_w32_r}
    \end{subfigure}
    \hfill
    \begin{subfigure}[t]{0.24\linewidth}
        \centering
        \includegraphics[width=\linewidth,height=0.27\textheight,keepaspectratio]{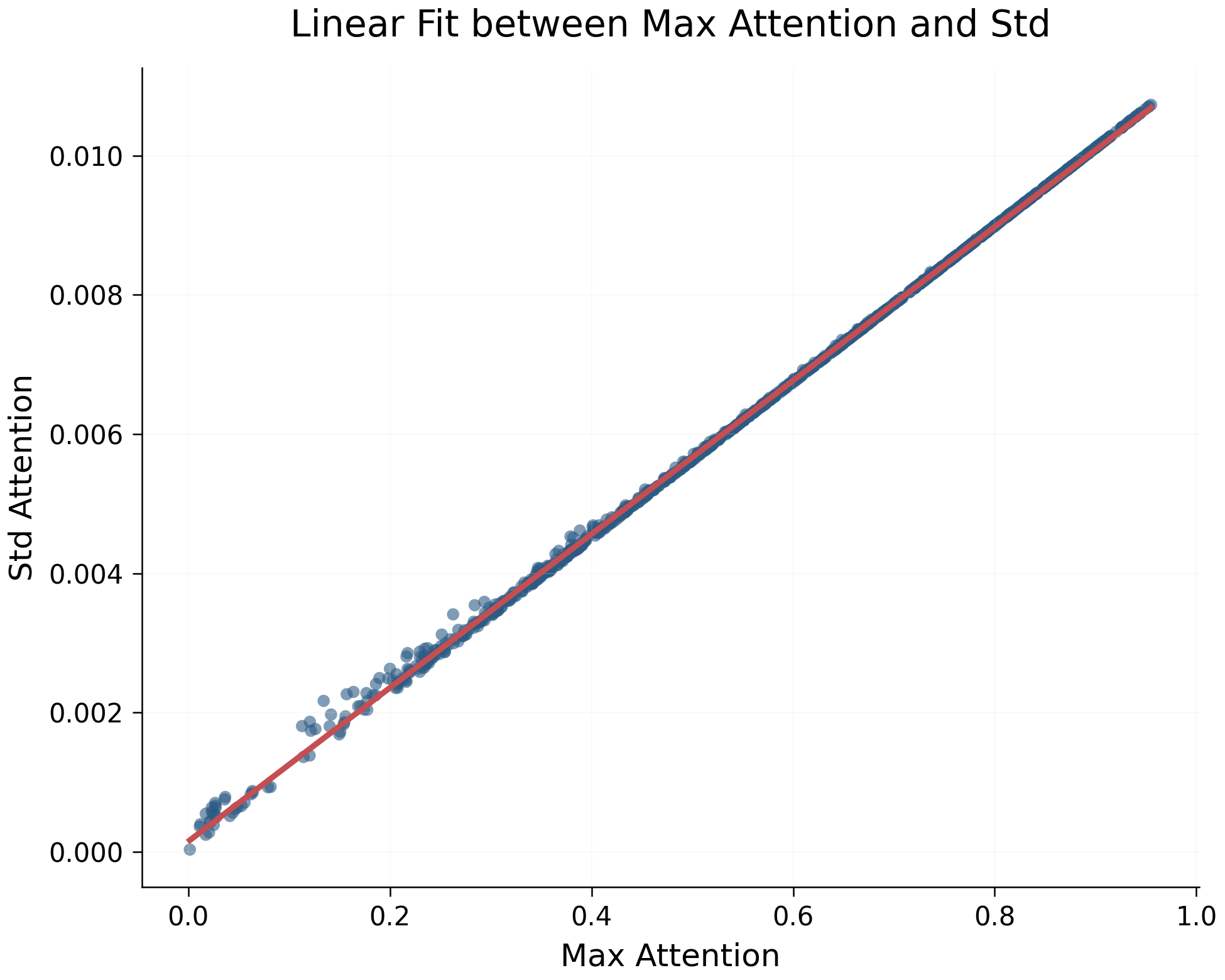}
        \caption{Linear fit.}
        \label{fig:Llama8b_w32_lin}
    \end{subfigure}

\caption{Correlation analysis on Llama3.1-8B-Instruct with \(W=32\).}
    \label{fig:Llama8b_w32_all}
\end{figure*}

\begin{figure*}[!t]
    \centering

    \begin{subfigure}[t]{0.24\linewidth}
        \centering
        \includegraphics[width=\linewidth,height=0.27\textheight,keepaspectratio]{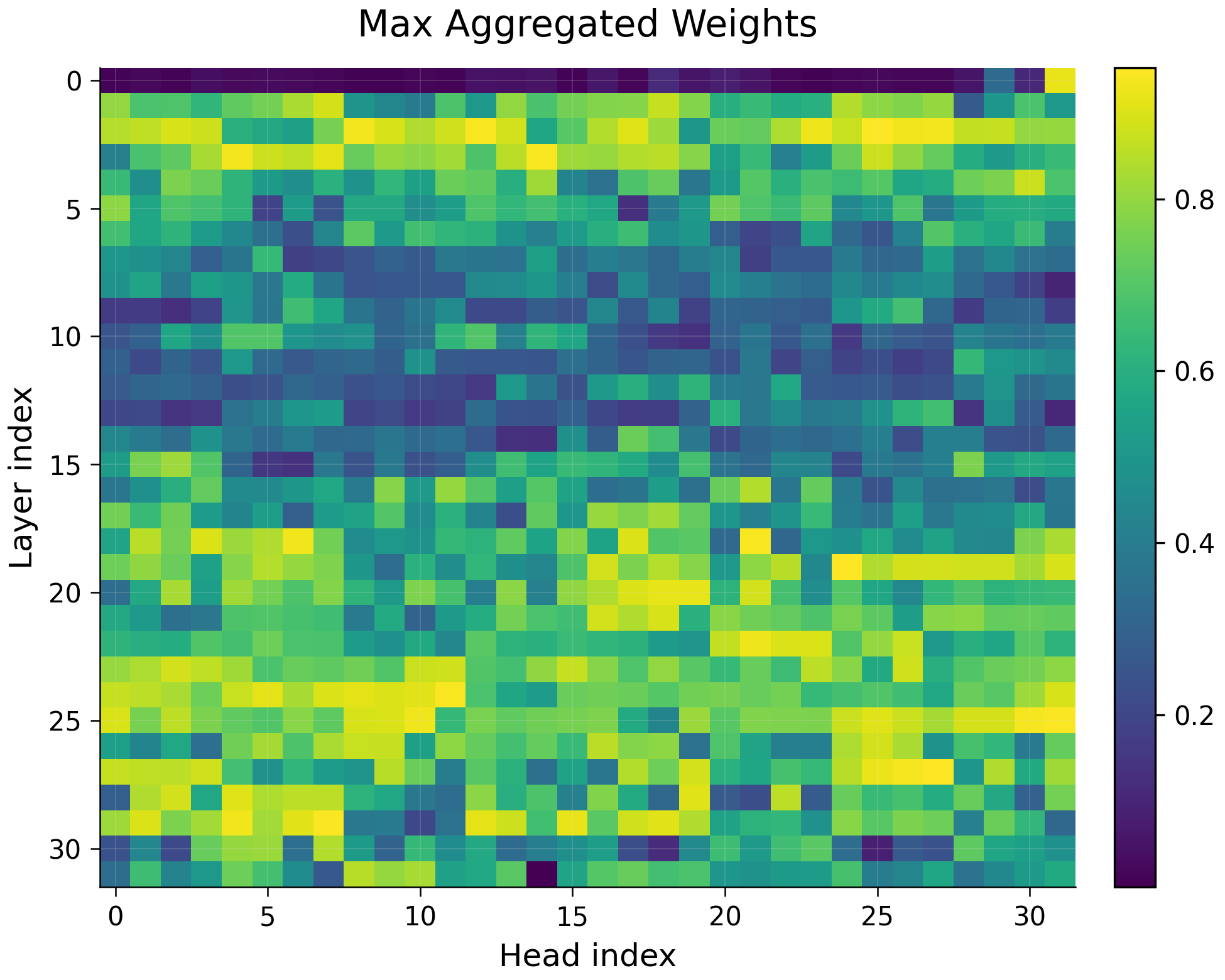}
        \caption{Max.}
        \label{fig:Llama8b_w64_max}
    \end{subfigure}
    \hfill
    \begin{subfigure}[t]{0.24\linewidth}
        \centering
        \includegraphics[width=\linewidth,height=0.27\textheight,keepaspectratio]{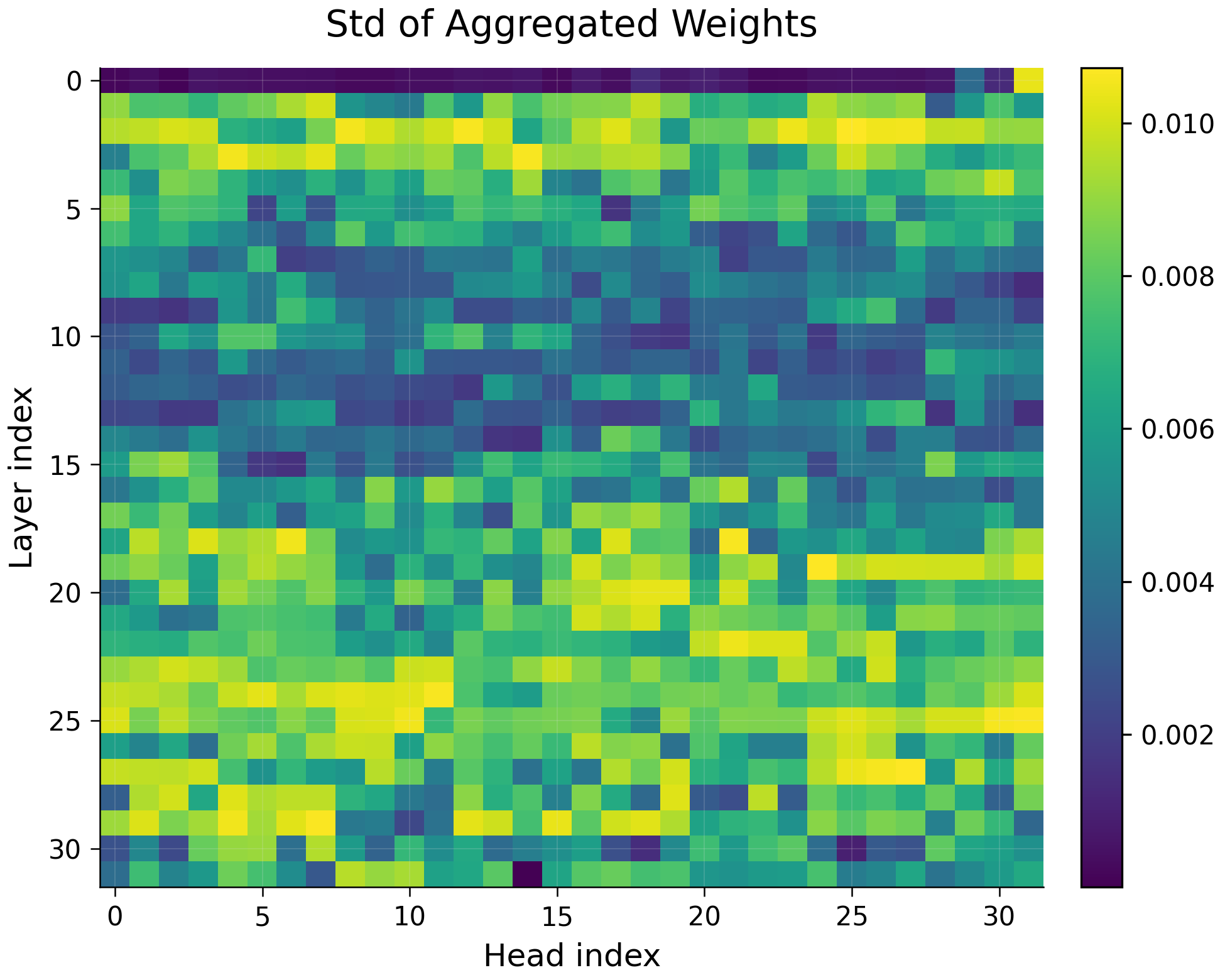}
        \caption{Std.}
        \label{fig:Llama8b_w64_std}
    \end{subfigure}
    \hfill
    \begin{subfigure}[t]{0.24\linewidth}
        \centering
        \includegraphics[width=\linewidth,height=0.27\textheight,keepaspectratio]{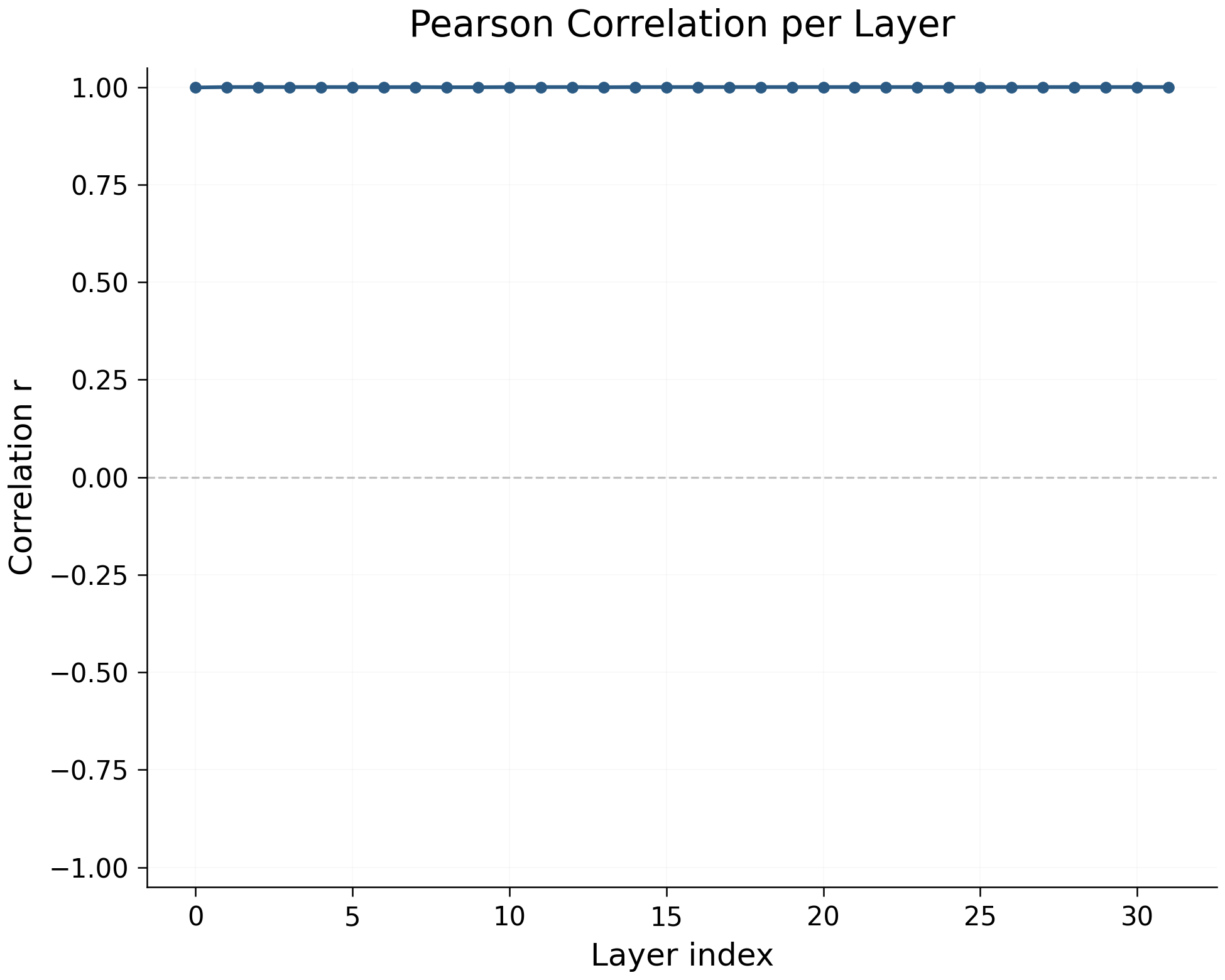}
        \caption{Pearson \(r\).}
        \label{fig:Llama8b_w64_r}
    \end{subfigure}
    \hfill
    \begin{subfigure}[t]{0.24\linewidth}
        \centering
        \includegraphics[width=\linewidth,height=0.27\textheight,keepaspectratio]{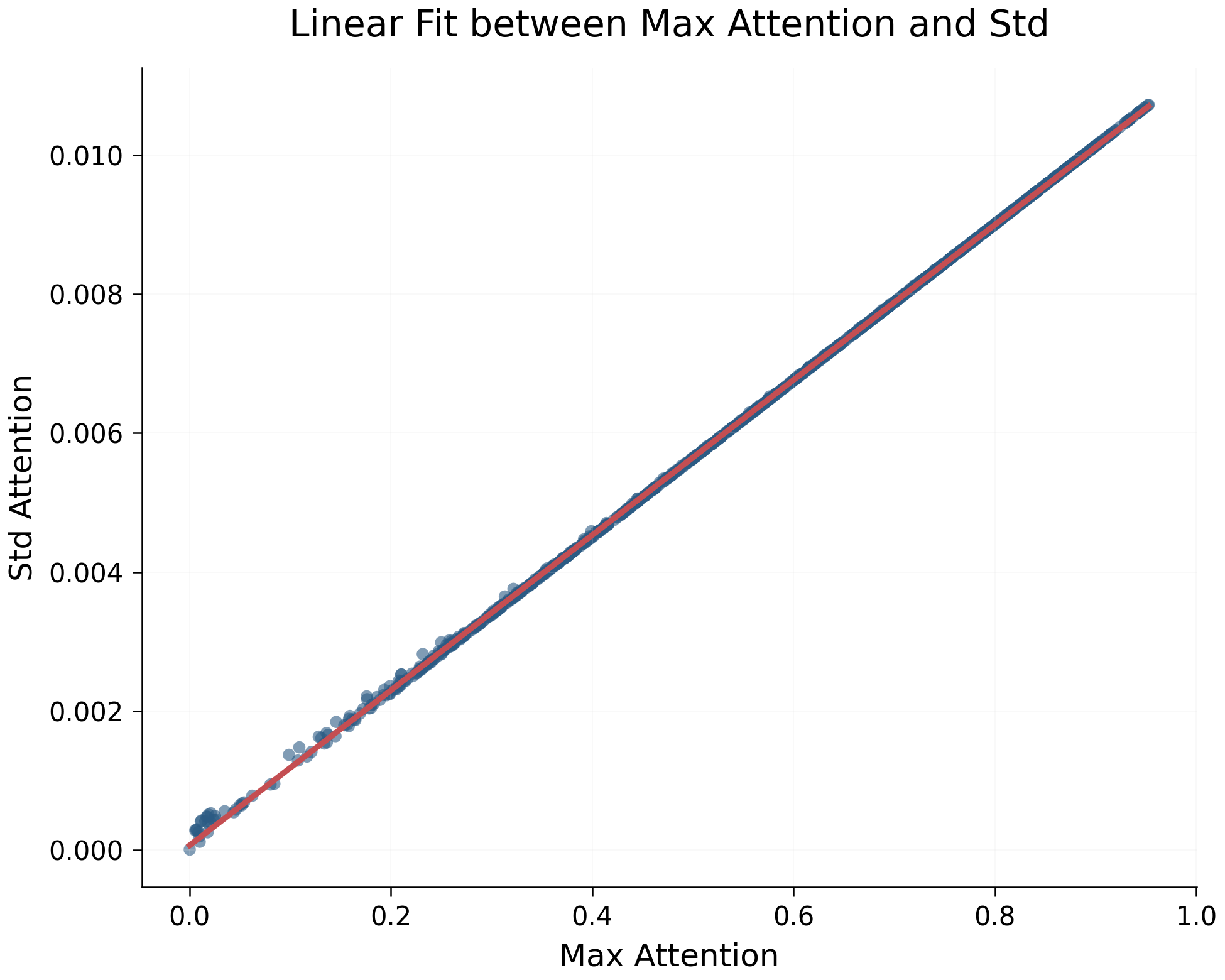}
        \caption{Linear fit.}
        \label{fig:Llama8b_w64_lin}
    \end{subfigure}

\caption{Correlation analysis on Llama3.1-8B-Instruct with \(W=64\).}
    \label{fig:Llama8b_w64_all}
\end{figure*}

It should be noted that this approximation does not strictly hold for all attention heads. It relies on the premise that the second-order statistics of the aggregated observation-window attention distribution are mainly dominated by the maximum attention score. In other words, this relationship is more suitable for sparse or approximately unimodal attention distributions, where a small number of historical tokens account for most of the attention mass. As indicated by the oscillating outlier points in the lower-left regions of the linear fitting plots in Figures~\ref{fig:Llama8b_w8_lin},~\ref{fig:mistral_w8_lin},~\ref{fig:Llama70b_w8_lin},~\ref{fig:Llama8b_w4_lin},~\ref{fig:Llama8b_w16_lin},~\ref{fig:Llama8b_w32_lin},~\ref{fig:Llama8b_w64_lin}, when the maximum aggregated observation-window attention score is very low, using \(a_{\max}\) alone as a lightweight proxy for the standard deviation \(\sigma\) is not reliable. We refer to such heads as abnormal heads. These low-response outliers motivate the abnormal-head mask used in Section~\ref{sec:methodology}. Instead of allowing inverse-max allocation to amplify their weights, \textsc{ChampionKV} conservatively assigns them only the floor budget. Heads identified by this mask do not participate in the adaptive budget allocation process and are only assigned the minimum floor budget. This mechanism avoids extremely unfair budget allocation caused by abnormal attention distributions and improves robustness across models and tasks.

In terms of asymptotic complexity, both finding $a_{\max}$ and computing $\sigma$ require a linear scan; however, finding $a_{\max}$ avoids second-moment computation and can be more easily integrated with existing attention-score accumulation or Top-$K$ selection procedures. It is important to note that Ada-KV~\cite{feng2026ada} directly optimizes attention mass retention through global cross-head Top-$K$ selection and therefore has an optimality property. \textsc{ChampionKV} does not attempt to theoretically replace Ada-KV~\cite{feng2026ada}'s global Top-$K$ mechanism; rather, it provides a lighter head-wise budget allocation proxy that avoids large-scale cross-head concatenation and selection, improving engineering friendliness for ultra-long prompt scenarios.

Based on the above derivation, this section empirically evaluates whether \(a_{\max}\) can serve as a stable ranking-level proxy for \(\sigma\) across model architectures, model scales, observation-window sizes, and datasets.

Unless otherwise specified, each correlation-analysis figure in this subsection contains four subplots: 
(a) the heatmap of \(a_{\max}^{l,h}\), 
(b) the heatmap of \(\sigma^{l,h}\), 
(c) the layer-wise Pearson correlation between them, and 
(d) the corresponding linear fit. 

\paragraph{Correlation between maximum attention and standard deviation.}

First, we record the aggregated observation-window attention scores of every layer and every head during the prefilling stage of model inference.
For each head, we compute its aggregated observation-window attention vector \({a}^{l,h}\),
and then calculate the maximum attention score \(a_{\max}^{l,h}\) and the standard deviation \(\sigma^{l,h}\).
Taking the NarrativeQA single-document question answering task as an example, when the observation window size is set to \(W=8\),
the experimental results on Llama3.1-8B-Instruct show that the heatmaps of \(a_{\max}^{l,h}\) and \(\sigma^{l,h}\) over the layer-head dimensions exhibit highly consistent distribution patterns, as shown in Figure~\ref{fig:Llama8b_w8_max} and Figure~\ref{fig:Llama8b_w8_std};
the scatter fitting results further show that the two quantities approximately lie on a straight line, as shown in Figure~\ref{fig:Llama8b_w8_lin}, with a Pearson correlation coefficient of \(r=0.9993\), as shown in Figure~\ref{fig:Llama8b_w8_r}.
This indicates that, under the same model and task setting, the maximum aggregated attention score can stably reflect the degree of attention concentration captured by the standard deviation.

\paragraph{Robustness across model architectures and scales.}

To verify that this proxy is not specific to Llama3.1-8B-Instruct or a single model architecture, we further conduct the same experiment on Mistral-7B-Instruct-v0.3 and Llama3.1-70B-Instruct. Mistral-7B-Instruct-v0.3 has a parameter scale similar to Llama3.1-8B-Instruct but a different architecture, while Llama3.1-70B-Instruct is used to test the stability of the relationship at a larger model scale.

The results on Mistral-7B-Instruct-v0.3 and Llama3.1-70B-Instruct show that the heatmaps of maximum attention scores and standard deviations remain highly consistent across models (Figure~\ref{fig:mistral_w8_max} and Figure~\ref{fig:mistral_w8_std}; Figure~\ref{fig:Llama70b_w8_max} and Figure~\ref{fig:Llama70b_w8_std}). The scatter plots also exhibit near-linear relationships (Figure~\ref{fig:mistral_w8_lin}; Figure~\ref{fig:Llama70b_w8_lin}) with Pearson correlation coefficients close to one for both models (Figure~\ref{fig:mistral_w8_r}; Figure~\ref{fig:Llama70b_w8_r}). This suggests that the max-attention proxy is not tied to a specific model architecture or parameter scale.

\paragraph{Robustness across observation window sizes.}

In addition to model architecture and scale, we examine the effect of the observation window size on this linear proxy. Specifically, on Llama3.1-8B-Instruct, we set \(W\in\{4,16,32,64\}\), fix the prompt length to \(L=8,000\), and repeat the correlation analysis between \(a_{\max}^{l,h}\) and \(\sigma^{l,h}\) on a NarrativeQA example.
The experimental results are shown in Figures~\ref{fig:Llama8b_w4_all},
\ref{fig:Llama8b_w16_all}, \ref{fig:Llama8b_w32_all}, and~\ref{fig:Llama8b_w64_all}. We therefore use concise subfigure captions in the following figures for readability.
Under different window sizes, the maximum attention score and the standard deviation remain highly linearly correlated, indicating that the proxy does not depend on a fixed observation window size.

\begin{table}[ht!]
\centering
\scriptsize
\setlength{\tabcolsep}{2.2pt}
\renewcommand{\arraystretch}{1.03}

\resizebox{\columnwidth}{!}{
\begin{tabular}{lcccc}
\toprule
Len. & Slope & $L-W$ & $1/\sqrt{L-W}$ & Err. (\%) \\
\midrule
2k  & 0.022322 & 1934  & 0.022739 & 1.8682 \\
4k  & 0.015649 & 3934  & 0.015943 & 1.8817 \\
6k  & 0.012753 & 5934  & 0.012982 & 1.7921 \\
8k  & 0.011034 & 7934  & 0.011227 & 1.7468 \\
10k & 0.009863 & 9934  & 0.010033 & 1.7253 \\
12k & 0.009004 & 11934 & 0.009154 & 1.6650 \\
14k & 0.008335 & 13934 & 0.008472 & 1.6381 \\
16k & 0.007794 & 15934 & 0.007922 & 1.6429 \\
32k & 0.005556 & 31358 & 0.005647 & 1.6397 \\
\bottomrule
\end{tabular}
}
\caption{Scaling relation between the fitted slope and \(1/\sqrt{L-W}\) on NarrativeQA with Llama3.1-8B-Instruct.}
\label{tab:llama8b_nrtv}
\end{table}

\begin{table}[ht!]
\centering
\scriptsize
\setlength{\tabcolsep}{2.2pt}
\renewcommand{\arraystretch}{1.03}

\resizebox{\columnwidth}{!}{
\begin{tabular}{lcccc}
\toprule
Dataset & $L-W$ & Slope & $1/\sqrt{L-W}$ & Err. (\%) \\
\midrule
2WikiMQA & 7063 & 0.011524 & 0.011899 & 3.2529 \\
GovReport & 7942 & 0.010821 & 0.011221 & 3.6974 \\
HotpotQA & 7941 & 0.010768 & 0.011222 & 4.2143 \\
LCC & 7918 & 0.011030 & 0.011238 & 1.8865 \\
MultiNews & 2656 & 0.018709 & 0.019404 & 3.7135 \\
MF-en & 1522 & 0.024590 & 0.025633 & 4.2399 \\
Musique & 7942 & 0.010838 & 0.011221 & 3.5347 \\
NrtvQA & 7942 & 0.010979 & 0.011221 & 2.2050 \\
PCount & 7942 & 0.010958 & 0.011221 & 2.4009 \\
PRe & 7942 & 0.010772 & 0.011221 & 4.1691 \\
Qasper & 3992 & 0.015424 & 0.015827 & 2.6143 \\
QMSum & 7933 & 0.010898 & 0.011227 & 3.0231 \\
RB-P & 7938 & 0.011061 & 0.011224 & 1.4729 \\
SAMSum & 7940 & 0.010950 & 0.011223 & 2.4886 \\
TREC & 4949 & 0.013884 & 0.014215 & 2.3827 \\
TriviaQA & 5499 & 0.012989 & 0.013485 & 3.8203 \\
\bottomrule
\end{tabular}
}
\caption{Slope validation across LongBench datasets on Llama3.1-8B-Instruct with \(W=8\).}
\label{tab:llama8b_datasets}
\end{table}

\begin{table}[ht!]
\centering
\scriptsize
\setlength{\tabcolsep}{2.2pt}
\renewcommand{\arraystretch}{1.03}

\resizebox{\columnwidth}{!}{
\begin{tabular}{lcccc}
\toprule
Dataset & $L-W$ & Slope & $1/\sqrt{L-W}$ & Err. (\%) \\
\midrule
GovReport & 7942 & 0.011058 & 0.011221 & 1.4749 \\
HotpotQA & 7941 & 0.010997 & 0.011222 & 2.0442 \\
LCC & 7918 & 0.011153 & 0.011238 & 0.7629 \\
NrtvQA & 7942 & 0.011121 & 0.011221 & 0.9000 \\
PCount & 7942 & 0.011094 & 0.011221 & 1.1456 \\
RB-P & 7938 & 0.011169 & 0.011224 & 0.4917 \\
TREC & 4949 & 0.014124 & 0.014215 & 0.6430 \\
\bottomrule
\end{tabular}
}
\caption{Slope validation across LongBench datasets on Llama3.1-70B-Instruct with \(W=8\).}
\label{tab:llama70b_datasets}
\end{table}

\begin{table}[ht!]
\centering
\scriptsize
\setlength{\tabcolsep}{2.2pt}
\renewcommand{\arraystretch}{1.03}

\resizebox{\columnwidth}{!}{
\begin{tabular}{lcccc}
\toprule
Dataset & $L-W$ & Slope & $1/\sqrt{L-W}$ & Err. (\%) \\
\midrule
2WikiMQA & 8093  & 0.010713 & 0.011116 & 3.7610 \\
GovReport & 11927 & 0.008692 & 0.009157 & 5.3452 \\
HotpotQA & 13132 & 0.008392 & 0.008726 & 3.9846 \\
LCC & 17798 & 0.007337 & 0.007496 & 2.1635 \\
MF-en & 1877 & 0.022153 & 0.023082 & 4.1922 \\
MultiNews & 3104 & 0.016902 & 0.017949 & 6.1943 \\
Musique & 18387 & 0.007159 & 0.007375 & 3.0130 \\
NrtvQA & 31492 & 0.005448 & 0.005635 & 3.4339 \\
PCount & 18048 & 0.007138 & 0.007444 & 4.2819 \\
PRe & 12604 & 0.008469 & 0.008907 & 5.1753 \\
Qasper & 4486 & 0.014236 & 0.014930 & 4.8775 \\
QMSum & 11811 & 0.008925 & 0.009201 & 3.0976 \\
RB-P & 16187 & 0.007111 & 0.007860 & 1.9310 \\
SAMSum & 11176 & 0.009132 & 0.009459 & 3.5836 \\
TREC & 5520 & 0.013019 & 0.013460 & 3.3839 \\
TriviaQA & 6551 & 0.011831 & 0.012355 & 4.4299 \\
\bottomrule
\end{tabular}
}
\caption{Slope validation across LongBench datasets on Mistral-7B-Instruct-v0.3 with \(W=8\).}
\label{tab:mistral_datasets}
\end{table}

\paragraph{Scaling relation with sequence length.}
Furthermore, we validate the relationship between the fitted slope and the number of historical tokens \(L-W\). According to the derivation in Section~\ref{appsubsec:Validation_of_the_Max-Attention_Proxy}, if the second-order statistics are mainly dominated by the maximum attention score, then
\[
\sigma^{l,h}\approx\frac{1}{\sqrt{L-W}}a_{\max}^{l,h},
\]
and the fitted slope should approximately satisfy
\[
\mathrm{slope}=\frac{1}{\sqrt{L-W}}.
\]
To verify this relation, we use NarrativeQA prompts on Llama3.1-8B-Instruct, fix the observation window size to \(W=16\), and vary the input length. Table~\ref{tab:llama8b_nrtv} shows that the fitted slope remains close to the theoretical value across different sequence lengths.

\paragraph{Robustness across datasets.}
To further rule out the possibility that this observation is specific to a single task example, we repeat the slope validation on multiple LongBench datasets. Tables~\ref{tab:llama8b_datasets},~\ref{tab:llama70b_datasets}, and~\ref{tab:mistral_datasets} report the results on Llama3.1-8B-Instruct, Llama3.1-70B-Instruct, and Mistral-7B-Instruct-v0.3, respectively. Across datasets and model scales, the fitted slope remains close to \(1/\sqrt{L-W}\).

\section{Appendix: Additional Experimental Results}
\label{app:experiments}

\paragraph{Hyperparameter analysis on Llama3.1-8B-Instruct.}
The Llama3.1-8B-Instruct sweep underlying the main-text conclusions is reported below. A larger floor ratio improves stability at larger budgets but reduces adaptive redistribution, while the best abnormal-head threshold varies with budget; therefore \((\rho,\tau)=(0.2,0.01)\) is used as a practical default rather than a universal optimum.

\vspace{0.6em}
\noindent
\begin{minipage}{\columnwidth}
    \centering
    \small
    \setlength{\tabcolsep}{6pt}
    \renewcommand{\arraystretch}{1.1}
    \begin{tabular}{lccc}
        \toprule
        \(\rho\) & Budget 128 & Budget 512 & Budget 2048 \\
        \midrule
        0.1 & \textbf{35.57} & 39.86 & 40.28 \\
        0.2 & 35.25 & 39.90 & 40.38 \\
        0.5 & 35.53 & \textbf{40.34} & \textbf{40.75} \\
        \bottomrule
    \end{tabular}
    \captionof{table}{Robustness analysis of the floor-budget ratio \(\rho\) on LongBench with Llama3.1-8B-Instruct. Average scores over six representative LongBench tasks.}
    \label{tab:robustness_floor_rate_Llama31_8b}
\end{minipage}

\vspace{1.0em}
\noindent
\begin{minipage}{\columnwidth}
    \centering
    \small
    \setlength{\tabcolsep}{6pt}
    \renewcommand{\arraystretch}{1.1}
    \begin{tabular}{lccc}
        \toprule
        \(\tau\) & Budget 128 & Budget 512 & Budget 2048 \\
        \midrule
        0.01 & \textbf{30.49} & \textbf{33.67} & 34.00 \\
        0.001 & 30.39 & 33.51 & \textbf{34.40} \\
        0.0001 & 29.92 & 33.40 & 34.32 \\
        \bottomrule
    \end{tabular}
    \captionof{table}{Robustness analysis of the abnormal-head threshold \(\tau\) on LongBench with Llama3.1-8B-Instruct. Average scores over six representative LongBench tasks.}
    \label{tab:robustness_abnormal_threshold_Llama31_8b}
\end{minipage}

\paragraph{Maximum decoding length for each dataset.} We evaluate on 16 LongBench datasets, and the maximum decoding length for each dataset is shown in Table~\ref{tab:LongBench_dataset_prompts}.

\paragraph{Hardware profiling across input lengths.}
Table~\ref{tab:hardware_efficiency_detail} reports the available profiling results for Llama3.1-8B-Instruct at input lengths of 8,100, 16,100, and 32,100 tokens. All settings use a KV cache budget of 1024 and generate 15 new tokens on \textbf{NVIDIA H200}. Prefill time covers processing the input sequence, decoding latency is the average time per generated token, and total generation time includes both prefilling and decoding. Across the three input lengths, the Champion variants consistently use slightly less peak GPU memory than Ada-KV. The timing differences vary by method and input length, so these measurements should be interpreted as configuration-specific profiling rather than a general speedup claim.

\begin{table}[ht!]
    \centering
    \small
    \setlength{\tabcolsep}{3pt}
    \renewcommand{\arraystretch}{1.1}
    \begin{tabular}{lcccc}
        \toprule
        Method & Mem. & Prefill & Dec. & Total \\
               & (GB) & (ms) & (ms/tok) & (ms) \\
        \midrule
        SnapKV            & 38.605 & \textbf{3795.82} & \textbf{29.46} & \textbf{4237.64} \\
        Ada-KV            & 38.605 & 3887.61 & 44.99 & 4562.39 \\
        Champion-Snap     & \textbf{38.230} & 3885.97 & 44.78 & 4557.73 \\
        Champion-Pyramid  & 38.240 & 3875.88 & 42.21 & 4509.08 \\
        \bottomrule
    \end{tabular}
    \caption{Hardware profiling on Llama3.1-8B-Instruct with a 32,100-token input, 15 generated tokens, and KV cache budget 1024. Bold marks the lowest value in each column.}
    \label{tab:hardware_efficiency_main_app}
\end{table}

\begin{table}[ht!]
    \centering
    \footnotesize
    \setlength{\tabcolsep}{8pt}
    \renewcommand{\arraystretch}{1}

    \begin{tabular}{lclc}
        \toprule
        \textbf{Dataset} & \textbf{Max len} &
        \textbf{Dataset} & \textbf{Max len} \\
        \midrule
        NrtvQA     & 128 & MultiNews & 512 \\
        Qasper     & 128 & PCount    & 32  \\
        MF-en      & 64  & PRe       & 32  \\
        HotpotQA   & 32  & Lcc       & 64  \\
        2WikiMQA   & 32  & TriviaQA  & 32  \\
        Musique    & 32  & SAMSum    & 128 \\
        GovReport  & 512 & TREC      & 64  \\
        QMSum      & 512 & RB-P      & 64  \\
        \bottomrule
    \end{tabular}
    \caption{Maximum decoding lengths for the 16 LongBench datasets used in evaluation.}
    \label{tab:LongBench_dataset_prompts}
\end{table}

Table~\ref{tab:hardware_efficiency_main_app} reports the profiling at the primary 1024-budget configuration described in the main text.

\paragraph{Hyperparameter analysis on Mistral-7B-Instruct-v0.3.}
Table~\ref{tab:mistral_hyperparameter_analysis} complements the Llama3.1-8B-Instruct analysis by jointly varying the floor-budget ratio \(\rho\) and abnormal-head threshold \(\tau\) on the same six representative LongBench task categories. The default pair \((0.2,0.01)\) yields the highest average among the tested settings, but other configurations achieve the best result on individual tasks. This supports using \(\rho=0.2\) and \(\tau=0.01\) as robustness-oriented defaults while allowing model-specific tuning.

\paragraph{Abnormal-head prevalence.}
We count a layer--head instance as abnormal when \(a_{\max}^{l,h}<0.01\), following the default threshold used in the main experiments. Table~\ref{tab:abnormal_head_prevalence} reports both the aggregate rate over all evaluated prompts and the range of prompt-level rates. The observed prevalence is low but varies across models and datasets, reinforcing that \(\tau=0.01\) is an empirical safeguard rather than a universal boundary.

\paragraph{Long-context slice.}
We further evaluate Champion-Snap at input lengths of 8k, 16k, and 32k on the six representative LongBench tasks used in the robustness analysis. As shown in Table~\ref{tab:long_context_slice}, the average scores vary by at most 0.55 points on Llama3.1-8B-Instruct and 0.32 points on Mistral-7B-Instruct-v0.3. These results support stability with respect to input length under this fixed evaluation protocol. They should not be interpreted as evidence for generation-heavy, long-chain-of-thought, or multi-turn settings, which are not evaluated here.

% Allow the two original RULER figures to share a text page as in the first version.
\renewcommand{\dbltopfraction}{0.95}
\renewcommand{\textfraction}{0.03}
\renewcommand{\dblfloatpagefraction}{0.95}
\setcounter{dbltopnumber}{2}

Detailed RULER scores on the three base models are shown in Tables~\ref{tab:ruler_llama31_8b_detail},~\ref{tab:ruler_mistral_detail} and~\ref{tab:ruler_llama31_70b_detail}.

\begin{table*}[t!]
    \centering
    \small
    \setlength{\tabcolsep}{2.5pt}
    \renewcommand{\arraystretch}{1.05}
    \begin{tabular}{lccccc}
        \toprule
        Method
        & Input len.
        & Peak Mem.
        & Prefill time
        & Decode Lat.
        & Total Gen. time \\
        & & (GB) & (ms) & (ms/token) & (ms) \\
        \midrule
        Ada-KV
        & 8,100
        & 21.221
        & 1097.12
        & 45.64
        & 1781.68 \\
        Champion-Snap
        & 8,100
        & 20.845
        & 800.56
        & 49.43
        & 1542.04 \\
        Champion-Pyramid
        & 8,100
        & 20.855
        & 922.82
        & 46.50
        & 1620.34 \\
        \midrule
        Ada-KV
        & 16,100
        & 27.014
        & 1677.00
        & 45.79
        & 2363.83 \\
        Champion-Snap
        & 16,100
        & 26.640
        & 1652.47
        & 46.33
        & 2347.44 \\
        Champion-Pyramid
        & 16,100
        & 26.650
        & 1636.34
        & 42.25
        & 2270.13 \\
        \midrule
        Ada-KV
        & 32,100
        & 38.605
        & 3887.61
        & 44.99
        & 4562.39 \\
        Champion-Snap
        & 32,100
        & 38.230
        & 3885.97
        & 44.78
        & 4557.73 \\
        Champion-Pyramid
        & 32,100
        & 38.240
        & 3875.88
        & 42.21
        & 4509.08 \\
        \bottomrule
    \end{tabular}
    \caption{Hardware profiling of Llama3.1-8B-Instruct across input lengths
    with 15 generated tokens and KV cache budget 1024.}
    \label{tab:hardware_efficiency_detail}
\end{table*}

\begin{table*}[t!]
    \centering
    \small
    \setlength{\tabcolsep}{2.5pt}
    \renewcommand{\arraystretch}{1.05}
    \begin{tabular}{ccccccccc}
        \toprule
        \(\rho\)
        & \(\tau\)
        & NrtvQA
        & HotpotQA
        & GovReport
        & TREC
        & PCount
        & Lcc
        & Avg. \\
        \midrule
        0.1
        & 0.01
        & 26.26
        & 45.37
        & 24.97
        & \textbf{71.00}
        & \textbf{7.00}
        & 57.88
        & 38.75 \\
        0.2
        & 0.0001
        & 26.09
        & 46.39
        & 24.78
        & 70.00
        & \textbf{7.00}
        & 57.70
        & 38.66 \\
        0.2
        & 0.001
        & 26.09
        & 46.39
        & \textbf{25.21}
        & 70.00
        & \textbf{7.00}
        & 57.74
        & 38.74 \\
        0.2
        & 0.01
        & \textbf{27.81}
        & 45.87
        & 25.15
        & 70.00
        & \textbf{7.00}
        & 57.73
        & \textbf{38.93} \\
        0.5
        & 0.01
        & 25.96
        & \textbf{46.95}
        & 24.87
        & \textbf{71.00}
        & 5.00
        & \textbf{58.51}
        & 38.72 \\
        \bottomrule
    \end{tabular}
    \caption{Hyperparameter analysis of Champion-Snap on
    Mistral-7B-Instruct-v0.3 over six representative LongBench tasks.
    Bold marks the best result in each column among the tested configurations.}
    \label{tab:mistral_hyperparameter_analysis}
\end{table*}

\paragraph{LongBench results on 7B/8B-scale models.} Table~\ref{tab:LongBench_Llama31_8b} reports the Llama3.1-8B-Instruct results. Ada-KV achieves the highest 16-task average at budgets 64, 128, and 2048 (36.36, 38.60, and 43.42), whereas SnapKV is strongest at budget 512 (41.57). At budget 2048, Champion-Pyramid reaches 43.08, but remains below Ada-KV's 43.42. Thus, the Champion variants win selected tasks but do not provide the strongest average on this model.

Table~\ref{tab:LongBench_mistral_7b} shows a similar adaptive-baseline advantage on Mistral-7B-Instruct-v0.3. Ada-KV obtains the highest average at all four budgets (41.06, 43.69, 46.41, and 47.27). Champion-Snap reaches 47.04 at budget 2048 and remains competitive with SnapKV and Champion-Pyramid, but does not exceed Ada-KV. These results reinforce that the smaller-model behavior is workload-dependent and that \textsc{ChampionKV} should not be interpreted as uniformly outperforming global cross-head allocation.

\paragraph{RULER results on 7B/8B-scale models.} The Llama3.1-8B-Instruct results are shown in Figure~\ref{fig:ruler_Llama31_8b} and Table~\ref{tab:ruler_llama31_8b_detail}. Champion-Snap improves over SnapKV by 0.56, 0.96, 0.14, and 0.52 points at 20\%, 40\%, 60\%, and 80\% retention. It also exceeds Ada-KV at 20\%--60\% retention (88.15 vs.\ 82.88, 93.57 vs.\ 90.63, and 97.16 vs.\ 95.99), while the two are nearly tied at 80\% (99.28 vs.\ 99.29). Champion-Pyramid improves over PyramidKV at 20\%, 60\%, and 80\%, but is slightly lower at 40\%.

The Mistral-7B-Instruct-v0.3 results are shown in Figure~\ref{fig:ruler_mistral} and Table~\ref{tab:ruler_mistral_detail}. Ada-KV obtains 78.83, 86.64, 90.56, and 94.10 across the four retention ratios, below the strongest result at each ratio. Champion-Snap exceeds SnapKV only at 20\% and 60\%, while Champion-Pyramid exceeds PyramidKV at 20\% and 80\%. Thus, neither adaptive method provides a uniform advantage on this model.

\begin{table*}[t]
    \centering
    \normalsize
    \setlength{\tabcolsep}{5.0pt}
    \renewcommand{\arraystretch}{1.05}
    \resizebox{\textwidth}{!}{
    \begin{tabular}{llccc}
        \toprule
        Model & Dataset & Abnormal instances & Total instances & Aggregate rate / Prompt-level range \\
        \midrule
        Llama3.1-8B-Instruct & GovReport   & 1,343 & 102,400 & 1.31\% / 0.59--1.56\% \\
        Llama3.1-8B-Instruct & HotpotQA    & 1,643 & 102,400 & 1.60\% / 0.39--2.64\% \\
        Llama3.1-8B-Instruct & Lcc         & 1,757 & 307,200 & 0.57\% / 0.10--1.27\% \\
        Llama3.1-8B-Instruct & NarrativeQA & 1,700 & 102,400 & 1.66\% / 1.07--3.42\% \\
        Llama3.1-8B-Instruct & PassageCount & 1,394 & 102,400 & 1.36\% / 1.17--1.46\% \\
        Llama3.1-8B-Instruct & TREC         & 1,069 & 102,400 & 1.04\% / 0.29--1.76\% \\
        \midrule
        Mistral-7B-Instruct-v0.3 & GovReport   & 1,169 & 102,400 & 1.14\% / 0.59--2.25\% \\
        Mistral-7B-Instruct-v0.3 & HotpotQA    & 1,336 & 102,400 & 1.30\% / 0.78--1.95\% \\
        Mistral-7B-Instruct-v0.3 & Lcc         & 2,477 & 307,200 & 0.81\% / 0.49--3.32\% \\
        Mistral-7B-Instruct-v0.3 & NarrativeQA & 1,531 & 102,400 & 1.50\% / 0.98--2.73\% \\
        Mistral-7B-Instruct-v0.3 & PassageCount & 1,175 & 102,400 & 1.15\% / 0.88--1.46\% \\
        Mistral-7B-Instruct-v0.3 & TREC         & 1,436 & 102,400 & 1.40\% / 0.68--2.54\% \\
        \bottomrule
    \end{tabular}
    }
    \caption{Prevalence of abnormal heads under \(\tau=0.01\). Total instances count layer--head pairs over all evaluated prompts; the final column reports the aggregate rate followed by the minimum--maximum prompt-level rate.}
    \label{tab:abnormal_head_prevalence}
\end{table*}
\vspace{-1.2em} 
\begin{table*}[t]
    \centering
    \normalsize
    \setlength{\tabcolsep}{4.8pt}
    \renewcommand{\arraystretch}{1.05}
    \resizebox{\textwidth}{!}{
    \begin{tabular}{lcrrrrrrr}
        \toprule
        Model & Input length & NrtvQA & HotpotQA & GovReport & TREC & PCount & Lcc & Avg. \\
        \midrule
        Llama3.1-8B-Instruct & 8k  & 20.32 & 15.95 & 25.69 & 66.00 & 8.42 & 60.11 & 32.75 \\
        Llama3.1-8B-Instruct & 16k & 24.28 & 14.88 & 25.51 & 65.00 & 8.06 & 59.42 & 32.86 \\
        Llama3.1-8B-Instruct & 32k & 28.01 & 14.36 & 25.24 & 66.00 & 6.64 & 59.53 & 33.30 \\
        \midrule
        Mistral-7B-Instruct-v0.3 & 8k  & 27.81 & 45.87 & 25.15 & 70.00 & 7.00 & 57.73 & 38.93 \\
        Mistral-7B-Instruct-v0.3 & 16k & 23.80 & 47.59 & 25.54 & 70.00 & 7.00 & 57.71 & 38.61 \\
        Mistral-7B-Instruct-v0.3 & 32k & 26.36 & 45.87 & 24.93 & 70.00 & 7.00 & 57.50 & 38.61 \\
        \bottomrule
    \end{tabular}
    }
    \caption{Champion-Snap results across input lengths on six representative LongBench tasks. Avg. is the mean over the six tasks.}
    \label{tab:long_context_slice}
\end{table*}

\begin{figure*}[!t]
    \centering

    \begin{tikzpicture}
        \begin{groupplot}[
            group style={
                group size=2 by 1,
                horizontal sep=1.0cm
            },
            width=0.40\textwidth,
            height=0.30\textwidth,
            xlabel={KV Cache Ratio},
            ylabel={Avg. Score},
            xmin=0.18,
            xmax=1.02,
            ymin=82,
            ymax=100,
            xtick={0.2,0.4,0.6,0.8,1.0},
            xticklabels={20\%,40\%,60\%,80\%,100\%},
            grid=both,
            grid style={
                line width=.1pt,
                draw=gray!20
            },
            major grid style={
                line width=.2pt,
                draw=gray!35
            },
            tick label style={font=\small},
            label style={font=\small},
            title style={font=\small\bfseries},
            legend style={
                font=\tiny,
                at={(0.5,-0.30)},
                anchor=north,
                legend columns=2
            },
            every axis plot/.append style={
                line width=1.1pt,
                mark size=2.2pt
            }
        ]

            \nextgroupplot[
                title={(a) SnapKV-based Allocation}
            ]

            \addplot+[
                mark=*,
                teal!70!black
            ] coordinates {
                (0.2,87.59)
                (0.4,92.61)
                (0.6,97.02)
                (0.8,98.76)
                (1.0,99.25)
            };
            \addlegendentry{SnapKV}

            \addplot+[
                mark=square*,
                orange!85!black
            ] coordinates {
                (0.2,88.15)
                (0.4,93.57)
                (0.6,97.16)
                (0.8,99.28)
                (1.0,99.25)
            };
            \addlegendentry{Champion-Snap}

            \nextgroupplot[
                title={(b) PyramidKV-based Allocation},
                ylabel={}
            ]

            \addplot+[
                mark=*,
                teal!70!black
            ] coordinates {
                (0.2,83.80)
                (0.4,90.27)
                (0.6,96.77)
                (0.8,98.79)
                (1.0,99.25)
            };
            \addlegendentry{PyramidKV}

            \addplot+[
                mark=square*,
                orange!85!black
            ] coordinates {
                (0.2,84.39)
                (0.4,89.90)
                (0.6,97.02)
                (0.8,99.13)
                (1.0,99.25)
            };
            \addlegendentry{Champion-Pyramid}

        \end{groupplot}
    \end{tikzpicture}

    \caption{RULER performance on Llama3.1-8B-Instruct under different
    KV cache ratios.}
    \label{fig:ruler_Llama31_8b}

    \vspace{0.6em}

    \begin{tikzpicture}
        \begin{groupplot}[
            group style={
                group size=2 by 1,
                horizontal sep=1.0cm
            },
            width=0.40\textwidth,
            height=0.30\textwidth,
            xlabel={KV Cache Ratio},
            ylabel={Avg. Score},
            xmin=0.18,
            xmax=1.02,
            ymin=82,
            ymax=100,
            xtick={0.2,0.4,0.6,0.8,1.0},
            xticklabels={20\%,40\%,60\%,80\%,100\%},
            grid=both,
            grid style={
                line width=.1pt,
                draw=gray!20
            },
            major grid style={
                line width=.2pt,
                draw=gray!35
            },
            tick label style={font=\small},
            label style={font=\small},
            title style={font=\small\bfseries},
            legend style={
                font=\tiny,
                at={(0.5,-0.30)},
                anchor=north,
                legend columns=2
            },
            every axis plot/.append style={
                line width=1.1pt,
                mark size=2.2pt
            }
        ]

            \nextgroupplot[
                title={(a) SnapKV-based Allocation}
            ]

            \addplot+[
                mark=*,
                teal!70!black
            ] coordinates {
                (0.2,85.26)
                (0.4,91.58)
                (0.6,94.93)
                (0.8,96.39)
                (1.0,96.85)
            };
            \addlegendentry{SnapKV}

            \addplot+[
                mark=square*,
                orange!85!black
            ] coordinates {
                (0.2,85.87)
                (0.4,91.01)
                (0.6,95.36)
                (0.8,96.03)
                (1.0,96.85)
            };
            \addlegendentry{Champion-Snap}

            \nextgroupplot[
                title={(b) PyramidKV-based Allocation},
                ylabel={}
            ]

            \addplot+[
                mark=*,
                teal!70!black
            ] coordinates {
                (0.2,84.08)
                (0.4,90.38)
                (0.6,94.98)
                (0.8,96.45)
                (1.0,96.85)
            };
            \addlegendentry{PyramidKV}

            \addplot+[
                mark=square*,
                orange!85!black
            ] coordinates {
                (0.2,84.46)
                (0.4,89.91)
                (0.6,94.69)
                (0.8,96.64)
                (1.0,96.85)
            };
            \addlegendentry{Champion-Pyramid}

        \end{groupplot}
    \end{tikzpicture}

    \caption{RULER performance on Mistral-7B-Instruct-v0.3 under different
    KV cache ratios.}
    \label{fig:ruler_mistral}

\end{figure*}

\clearpage

\begin{table*}[!t]
    \centering
    \normalsize
    \setlength{\tabcolsep}{2.4pt}
    \renewcommand{\arraystretch}{1.0}
    \resizebox{\textwidth}{!}{
    \begin{tabular}{l ccc ccc ccc ccc cc cc c}
    \toprule
    & \multicolumn{3}{c}{Single-Doc QA}
    & \multicolumn{3}{c}{Multi-Doc QA}
    & \multicolumn{3}{c}{Summarization}
    & \multicolumn{3}{c}{Few-shot Learning}
    & \multicolumn{2}{c}{Synthetic}
    & \multicolumn{2}{c}{Code} & \\
    \cmidrule(lr){2-4}\cmidrule(lr){5-7}\cmidrule(lr){8-10}
    \cmidrule(lr){11-13}\cmidrule(lr){14-15}\cmidrule(lr){16-17}
    \raisebox{1em}{Method}
    & \rotatebox[origin=c]{35}{NrtvQA}
    & \rotatebox[origin=c]{35}{Qasper}
    & \rotatebox[origin=c]{35}{MF-en}
    & \rotatebox[origin=c]{35}{HotpotQA}
    & \rotatebox[origin=c]{35}{2WikiMQA}
    & \rotatebox[origin=c]{35}{Musique}
    & \rotatebox[origin=c]{35}{GovReport}
    & \rotatebox[origin=c]{35}{QMSum}
    & \rotatebox[origin=c]{35}{MultiNews}
    & \rotatebox[origin=c]{35}{TREC}
    & \rotatebox[origin=c]{35}{TriviaQA}
    & \rotatebox[origin=c]{35}{SAMSum}
    & \rotatebox[origin=c]{35}{PCount}
    & \rotatebox[origin=c]{35}{PRe}
    & \rotatebox[origin=c]{35}{LCC}
    & \rotatebox[origin=c]{35}{RB-P}
    & \rotatebox[origin=c]{35}{Avg.} \\
    \midrule

    \multicolumn{18}{c}{KV Budget = Full} \\
    \midrule
    FullKV & 21.77 & 43.07 & 27.82 & 52.29 & 43.86 & 27.05
    & 34.20 & 22.31 & 26.86 & 67.00 & 88.26 & 44.07
    & 7.50 & 65.67 & 64.66 & 55.32 & 43.23 \\

    \midrule
    \multicolumn{18}{c}{KV Budget = 64} \\
    \midrule
    SnapKV & 15.04 & 23.29 & 21.19 & 40.06 & 33.41 & 20.02
    & 20.64 & 22.01 & 19.85 & \textbf{56.00} & \textbf{83.27}
    & \textbf{40.67} & 8.17 & 61.70 & 57.85 & 44.69 & 35.49 \\

    Ada-KV & \textbf{17.78} & \textbf{27.16} & \textbf{22.56}
    & \textbf{41.80} & 34.03 & 20.34 & \textbf{21.75}
    & \textbf{22.23} & \textbf{21.01} & 54.00 & 80.58 & 39.92
    & \textbf{9.03} & \textbf{67.86} & 56.15 & \textbf{45.63}
    & \textbf{36.36} \\

    Champion-Snap & 14.17 & 26.61 & 21.12 & 40.48
    & \textbf{35.14} & 19.89 & 20.35 & 21.35 & 18.49
    & 51.00 & 82.13 & 40.27 & 7.75 & 61.36
    & \textbf{58.86} & 44.66 & 35.23 \\

    PyramidKV & 12.01 & 25.12 & 20.10 & 41.05 & 33.28
    & \textbf{21.92} & 20.83 & 21.49 & 19.78
    & \textbf{56.00} & 81.37 & 38.93 & 7.42 & 61.53
    & 56.76 & 44.11 & 35.11 \\

    Champion-Pyramid & 11.02 & 25.71 & 19.34 & 40.77
    & 34.02 & 18.61 & 19.87 & 21.95 & 18.11 & 55.00
    & 79.76 & 39.98 & 8.30 & 62.50 & 55.79 & 44.17 & 34.68 \\

    \midrule
    \multicolumn{18}{c}{KV Budget = 128} \\
    \midrule
    SnapKV & 17.44 & 25.68 & 22.83 & 44.37 & 34.07 & 19.99
    & 22.39 & 21.73 & 21.46 & 59.00 & 86.94
    & \textbf{41.55} & 7.15 & 63.50 & 61.99 & 46.54 & 37.29 \\

    Ada-KV & \textbf{19.69} & \textbf{30.88} & \textbf{23.41}
    & 43.38 & \textbf{35.79} & \textbf{22.97} & \textbf{23.00}
    & 22.44 & \textbf{21.78} & 60.00 & 87.77 & 41.48
    & 7.36 & \textbf{70.87} & 58.49 & \textbf{48.29}
    & \textbf{38.60} \\

    Champion-Snap & 16.02 & 27.84 & 21.90 & 42.57 & 33.20
    & 20.96 & 22.21 & 22.36 & 21.50 & 58.00
    & \textbf{87.78} & 41.16 & 7.68 & 61.02
    & \textbf{63.99} & 45.52 & 37.11 \\

    PyramidKV & 12.13 & 27.58 & 22.31 & 40.35 & 33.84
    & 19.77 & 22.42 & 21.72 & 21.47 & \textbf{61.00}
    & 83.25 & 41.05 & 7.50 & 64.33 & 60.44 & 45.14 & 36.52 \\

    Champion-Pyramid & 14.34 & 25.83 & 21.98 & \textbf{46.37}
    & 33.38 & 19.39 & 22.42 & \textbf{22.86} & 20.19
    & 60.00 & 84.97 & 40.57 & \textbf{9.00}
    & 62.83 & 60.19 & 45.26 & 36.85 \\

    \midrule
    \multicolumn{18}{c}{KV Budget = 512} \\
    \midrule
    SnapKV & \textbf{21.38} & 34.07 & 27.30 & \textbf{55.87}
    & 38.14 & \textbf{28.64} & \textbf{25.75} & \textbf{23.86}
    & 24.59 & 66.00 & 88.28 & \textbf{42.89} & 9.16
    & 64.23 & 63.39 & 51.61 & \textbf{41.57} \\

    Ada-KV & 20.31 & \textbf{35.92} & 24.11 & 50.08
    & \textbf{38.40} & 28.11 & 24.98 & 22.54 & 24.45
    & \textbf{67.00} & \textbf{91.50} & 42.44 & 6.54
    & \textbf{72.30} & 61.42 & 50.69 & 41.30 \\

    Champion-Snap & 21.30 & 34.74 & \textbf{27.52} & 52.39
    & 37.77 & 28.33 & 25.60 & 22.75 & \textbf{24.70}
    & 65.00 & 87.67 & 42.79 & \textbf{9.17}
    & 64.50 & 64.18 & \textbf{52.58} & 41.31 \\

    PyramidKV & 20.67 & 34.86 & 25.32 & 48.85 & 38.09
    & 25.39 & 25.58 & 22.62 & 24.67 & 65.00
    & 87.66 & 42.36 & 6.89 & 64.92 & \textbf{64.68}
    & 50.54 & 40.51 \\

    Champion-Pyramid & 20.67 & 33.40 & 26.70 & 47.24
    & 37.85 & 27.08 & 25.37 & 22.92 & 23.83 & 65.00
    & 88.28 & 42.45 & 7.19 & 65.33 & 63.80 & 49.10 & 40.39 \\

    \midrule
    \multicolumn{18}{c}{KV Budget = 2048} \\
    \midrule
    SnapKV & 20.59 & 41.35 & \textbf{27.38} & 52.65
    & 40.54 & 26.67 & 30.81 & 22.56 & 25.55 & 67.00
    & 87.55 & 43.57 & 6.95 & 65.33 & 58.97 & 54.19 & 41.98 \\

    Ada-KV & \textbf{20.96} & 39.47 & 26.46 & 53.17
    & 42.75 & \textbf{31.77} & 29.91 & 22.91 & 26.67
    & \textbf{69.50} & \textbf{91.37} & 43.80 & 7.08
    & \textbf{71.86} & 62.99 & 54.06 & \textbf{43.42} \\

    Champion-Snap & 19.83 & 42.54 & 26.97 & 52.13 & 40.01
    & 29.26 & 30.34 & 22.88 & 26.77 & 66.00 & 87.67
    & 44.21 & \textbf{7.15} & 65.67 & 64.24
    & \textbf{55.31} & 42.56 \\

    PyramidKV & 20.70 & 40.72 & 26.32 & \textbf{54.03}
    & \textbf{42.77} & 28.67 & \textbf{30.94}
    & \textbf{23.60} & 26.69 & 67.00 & 88.26 & 43.95
    & 6.59 & 65.67 & \textbf{64.36} & 55.00 & 42.83 \\

    Champion-Pyramid & 19.76 & \textbf{44.41} & 27.01 & 53.68
    & 42.67 & 31.07 & 29.87 & 23.18 & \textbf{26.82}
    & 67.00 & 88.27 & \textbf{44.31} & 6.69
    & 65.67 & 64.06 & 54.85 & 43.08 \\

    \bottomrule
    \end{tabular}
    }
    \caption{LongBench results on Llama3.1-8B-Instruct. Avg. is the mean over all 16 tasks. FullKV is the uncompressed reference; bold marks the best compressed result within each budget.}
    \label{tab:LongBench_Llama31_8b}
\end{table*}

\begin{table*}[!t]
    \centering
    \normalsize
    \setlength{\tabcolsep}{2.4pt}
    \renewcommand{\arraystretch}{1.0}
    \resizebox{\textwidth}{!}{
    \begin{tabular}{l ccc ccc ccc ccc cc cc c}
    \toprule
    & \multicolumn{3}{c}{Single-Doc QA}
    & \multicolumn{3}{c}{Multi-Doc QA}
    & \multicolumn{3}{c}{Summarization}
    & \multicolumn{3}{c}{Few-shot Learning}
    & \multicolumn{2}{c}{Synthetic}
    & \multicolumn{2}{c}{Code} & \\
    \cmidrule(lr){2-4}\cmidrule(lr){5-7}\cmidrule(lr){8-10}
    \cmidrule(lr){11-13}\cmidrule(lr){14-15}\cmidrule(lr){16-17}
    \raisebox{1em}{Method}
    & \rotatebox[origin=c]{35}{NrtvQA}
    & \rotatebox[origin=c]{35}{Qasper}
    & \rotatebox[origin=c]{35}{MF-en}
    & \rotatebox[origin=c]{35}{HotpotQA}
    & \rotatebox[origin=c]{35}{2WikiMQA}
    & \rotatebox[origin=c]{35}{Musique}
    & \rotatebox[origin=c]{35}{GovReport}
    & \rotatebox[origin=c]{35}{QMSum}
    & \rotatebox[origin=c]{35}{MultiNews}
    & \rotatebox[origin=c]{35}{TREC}
    & \rotatebox[origin=c]{35}{TriviaQA}
    & \rotatebox[origin=c]{35}{SAMSum}
    & \rotatebox[origin=c]{35}{PCount}
    & \rotatebox[origin=c]{35}{PRe}
    & \rotatebox[origin=c]{35}{LCC}
    & \rotatebox[origin=c]{35}{RB-P}
    & \rotatebox[origin=c]{35}{Avg.} \\
    \midrule

    \multicolumn{18}{c}{KV Budget = Full} \\
    \midrule
    FullKV & 29.21 & 41.24 & 54.06 & 46.08 & 37.94 & 24.66
    & 34.71 & 26.22 & 28.22 & 73.00 & 85.26 & 47.16
    & 7.00 & 99.00 & 62.61 & 61.65 & 47.38 \\

    \midrule
    \multicolumn{18}{c}{KV Budget = 64} \\
    \midrule
    SnapKV & 24.89 & 27.46 & 45.20 & 40.94 & 34.85 & 22.31
    & 19.67 & 21.37 & 18.64 & 53.00 & 83.79 & 40.15
    & 8.00 & 92.00 & \textbf{56.15} & 50.51 & 39.93 \\

    Ada-KV & \textbf{25.71} & 29.03 & \textbf{46.99}
    & \textbf{45.33} & 33.06 & \textbf{23.90}
    & \textbf{21.05} & 20.87 & \textbf{20.18}
    & \textbf{67.50} & \textbf{88.95} & \textbf{42.47}
    & 4.50 & 85.00 & 51.79 & \textbf{50.58} & \textbf{41.06} \\

    Champion-Snap & 24.67 & \textbf{29.88} & 40.68 & 39.97
    & \textbf{36.82} & 20.85 & 19.84 & 20.70 & 18.72
    & 53.00 & 82.63 & 39.10 & \textbf{9.00}
    & 89.00 & 53.24 & 43.90 & 38.88 \\

    PyramidKV & 22.32 & 28.14 & 46.55 & 42.13 & 34.21
    & 21.32 & 19.81 & \textbf{22.27} & 19.11 & 53.00
    & 83.69 & 40.00 & \textbf{9.00} & \textbf{95.00}
    & 53.12 & 47.00 & 39.79 \\

    Champion-Pyramid & 22.67 & 29.20 & 45.61 & 41.47
    & 34.73 & 19.98 & 19.79 & 21.80 & 19.23 & 49.00
    & 82.62 & 39.27 & \textbf{9.00} & 94.00
    & 53.51 & 47.12 & 39.31 \\

    \midrule
    \multicolumn{18}{c}{KV Budget = 128} \\
    \midrule
    SnapKV & 27.02 & 30.61 & 48.45 & 43.53 & 36.57 & 22.90
    & 21.47 & 23.24 & 21.55 & 67.00 & 83.76 & 42.78
    & 8.00 & 94.00 & \textbf{59.10} & 54.52 & 42.78 \\

    Ada-KV & 26.88 & 32.00 & 49.45 & \textbf{47.44}
    & 37.60 & \textbf{26.64} & \textbf{22.82} & 23.01
    & \textbf{22.71} & \textbf{70.50} & \textbf{89.07}
    & 44.17 & 5.00 & 92.50 & 54.92 & 54.37 & \textbf{43.69} \\

    Champion-Snap & \textbf{27.81} & \textbf{33.39} & 49.78
    & 42.57 & 37.61 & 23.08 & 21.01 & 23.13 & 21.70
    & 66.00 & 84.19 & 42.19 & 8.00 & 94.00
    & 52.95 & \textbf{54.75} & 42.64 \\

    PyramidKV & 25.56 & 32.35 & \textbf{49.93} & 43.44
    & 36.50 & 21.94 & 21.97 & 22.95 & 21.31 & 65.00
    & 84.46 & \textbf{44.58} & 8.00 & 94.00
    & 56.96 & 51.45 & 42.53 \\

    Champion-Pyramid & 25.08 & 31.94 & 49.07 & 42.19
    & \textbf{37.68} & 21.82 & 21.16 & \textbf{23.37}
    & 21.33 & 64.00 & 85.03 & 44.07 & \textbf{9.00}
    & \textbf{96.00} & 55.46 & 50.81 & 42.38 \\

    \midrule
    \multicolumn{18}{c}{KV Budget = 512} \\
    \midrule
    SnapKV & 27.76 & 34.58 & 54.47 & 47.61 & 38.74 & 22.98
    & \textbf{25.44} & 24.21 & 25.39 & 71.00 & 86.46 & 46.07
    & 5.00 & \textbf{99.00} & \textbf{62.65}
    & \textbf{61.04} & 45.78 \\

    Ada-KV & \textbf{29.37} & \textbf{37.00} & 53.61
    & \textbf{49.65} & 38.70 & \textbf{28.38} & 23.67
    & \textbf{25.46} & 24.79 & \textbf{75.50}
    & \textbf{89.14} & 46.42 & 5.50 & 97.00
    & 58.72 & 59.69 & \textbf{46.41} \\

    Champion-Snap & 28.27 & 34.93 & 54.39 & 44.92 & 37.60
    & 23.51 & 24.34 & 25.41 & 25.09 & 71.00 & 85.80
    & 46.08 & \textbf{9.00} & 97.00 & 61.31 & 59.60 & 45.52 \\

    PyramidKV & 27.70 & 34.59 & 54.31 & 47.54
    & \textbf{38.99} & 23.62 & 25.21 & 24.44
    & \textbf{25.55} & 72.00 & 86.46 & \textbf{46.45}
    & 6.00 & 98.00 & 62.45 & 59.07 & 45.77 \\

    Champion-Pyramid & 27.47 & 34.62 & \textbf{54.55} & 45.89
    & 38.41 & 23.92 & 24.24 & \textbf{25.46} & 24.86
    & 70.00 & 86.12 & 46.23 & 7.00 & 97.00
    & 61.76 & 58.80 & 45.40 \\

    \midrule
    \multicolumn{18}{c}{KV Budget = 2048} \\
    \midrule
    SnapKV & 30.35 & 39.82 & 53.79 & 46.49 & 38.64 & 24.91
    & \textbf{30.23} & 25.84 & 27.98 & 72.00 & 85.69
    & 46.30 & 6.00 & \textbf{99.00}
    & \textbf{63.82} & \textbf{60.70} & 46.97 \\

    Ada-KV & 29.37 & 40.56 & 52.49 & \textbf{50.00}
    & 38.64 & \textbf{28.50} & 28.97 & 25.10
    & \textbf{28.00} & \textbf{75.50} & \textbf{88.80}
    & \textbf{47.15} & 5.50 & 98.00 & 59.51 & 60.30
    & \textbf{47.27} \\

    Champion-Snap & 28.13 & \textbf{41.25} & 55.12 & 46.84
    & 38.24 & 24.60 & 29.44 & 25.81 & 27.78 & 73.00
    & 85.76 & 46.30 & 7.00 & \textbf{99.00}
    & 63.72 & 60.60 & 47.04 \\

    PyramidKV & \textbf{30.48} & 40.22 & 51.82 & 46.45
    & \textbf{38.93} & 24.23 & 29.96 & \textbf{26.06}
    & 27.18 & 72.00 & 86.19 & 46.91 & 6.00
    & \textbf{99.00} & 62.49 & 58.46 & 46.65 \\

    Champion-Pyramid & 29.99 & 40.72 & \textbf{55.25} & 46.49
    & 38.13 & 24.18 & 29.35 & 24.63 & 27.85 & 73.00
    & 85.86 & 46.07 & \textbf{8.00} & \textbf{99.00}
    & 63.48 & 59.17 & 46.95 \\

    \bottomrule
    \end{tabular}
    }
    \caption{LongBench results on Mistral-7B-Instruct-v0.3. Avg. is the mean over all 16 tasks. FullKV is the uncompressed reference; bold marks the best compressed result within each budget.}
    \label{tab:LongBench_mistral_7b}
\end{table*}

\begin{table*}[ht]
    \centering
    \scriptsize
    \setlength{\tabcolsep}{3.2pt}
    \resizebox{\textwidth}{!}{
        \begin{tabular}{llcccccccccccc}
            \toprule
            Method & Ratio & S1 & S2 & S3 & MK1 & MK2 & MK3 & MQ & MV & CWE & FWE & VT & Avg. \\
            \midrule
            FullKV & 100\% & 100.00 & 100.00 & 100.00 & 100.00 & 100.00 & 99.22 & 100.00 & 95.82 & 99.52 & 97.40 & 99.81 & 99.25 \\
            \midrule
            SnapKV & 20\% & 100.00 & 100.00 & 60.21 & 100.00 & 100.00 & 37.88 & 99.02 & 96.32 & 73.99 & 96.07 & 100.00 & 87.59 \\
            Ada-KV & 20\% & 100.00 & 100.00 & 44.16 & 100.00 & 99.42 & 32.20 & 99.51 & 97.21 & 43.92 & 95.39 & 99.85 & 82.88 \\
            Champion-Snap & 20\% & 100.00 & 100.00 & 62.36 & 100.00 & 99.04 & 38.56 & 99.76 & 96.80 & 77.26 & 96.09 & 99.81 & 88.15 \\
            
            SnapKV & 40\% & 100.00 & 100.00 & 87.38 & 100.00 & 100.00 & 41.41 & 100.00 & 95.05 & 97.69 & 97.39 & 99.81 & 92.61 \\
            Ada-KV & 40\% & 100.00 & 100.00 & 80.84 & 100.00 & 99.81 & 37.43 & 99.90 & 96.81 & 85.85 & 96.36 & 99.96 & 90.63 \\
            Champion-Snap & 40\% & 100.00 & 100.00 & 88.61 & 100.00 & 100.00 & 47.90 & 100.00 & 97.83 & 97.72 & 97.40 & 99.81 & 93.57 \\
            
            SnapKV & 60\% & 100.00 & 100.00 & 96.64 & 100.00 & 100.00 & 78.85 & 100.00 & 96.07 & 98.42 & 97.39 & 99.81 & 97.02 \\
            Ada-KV & 60\% & 100.00 & 100.00 & 98.04 & 100.00 & 100.00 & 66.15 & 99.90 & 96.56 & 98.17 & 97.14 & 99.96 & 95.99 \\
            Champion-Snap & 60\% & 100.00 & 100.00 & 96.98 & 100.00 & 100.00 & 80.73 & 100.00 & 95.83 & 98.66 & 96.75 & 99.81 & 97.16 \\
            
            SnapKV & 80\% & 100.00 & 100.00 & 99.78 & 100.00 & 100.00 & 95.60 & 100.00 & 95.31 & 99.13 & 96.74 & 99.81 & 98.76 \\
            Ada-KV & 80\% & 100.00 & 100.00 & 99.27 & 100.00 & 100.00 & 99.39 & 99.90 & 96.27 & 99.73 & 97.66 & 99.96 & 99.29 \\
            Champion-Snap & 80\% & 100.00 & 100.00 & 99.78 & 100.00 & 100.00 & 99.22 & 100.00 & 96.82 & 99.42 & 97.07 & 99.81 & 99.28 \\
            
            \midrule
            PyramidKV & 20\% & 100.00 & 100.00 & 42.41 & 100.00 & 100.00 & 35.49 & 98.56 & 95.33 & 55.72 & 94.46 & 99.81 & 83.80 \\
            Champion-Pyramid & 20\% & 100.00 & 100.00 & 44.38 & 100.00 & 100.00 & 36.44 & 99.28 & 95.85 & 57.69 & 94.80 & 99.81 & 84.39 \\
            
            PyramidKV & 40\% & 100.00 & 100.00 & 73.72 & 100.00 & 100.00 & 44.48 & 100.00 & 96.07 & 82.15 & 96.74 & 99.81 & 90.27 \\
            Champion-Pyramid & 40\% & 100.00 & 100.00 & 72.47 & 100.00 & 100.00 & 45.12 & 100.00 & 95.83 & 80.57 & 95.11 & 99.81 & 89.90 \\
            
            PyramidKV & 60\% & 100.00 & 100.00 & 93.98 & 100.00 & 100.00 & 78.85 & 100.00 & 96.07 & 99.42 & 96.39 & 99.81 & 96.77 \\
            Champion-Pyramid & 60\% & 100.00 & 100.00 & 96.25 & 100.00 & 100.00 & 81.90 & 100.00 & 95.58 & 96.22 & 97.41 & 99.81 & 97.02 \\
            
            PyramidKV & 80\% & 100.00 & 100.00 & 99.78 & 100.00 & 100.00 & 95.60 & 100.00 & 95.31 & 99.42 & 96.74 & 99.81 & 98.79 \\
            Champion-Pyramid & 80\% & 100.00 & 100.00 & 99.13 & 100.00 & 100.00 & 98.47 & 100.00 & 96.58 & 99.33 & 97.07 & 99.81 & 99.13 \\
            \bottomrule
        \end{tabular}
    }
    \caption{Detailed RULER subtask results on Llama3.1-8B-Instruct under different KV cache retention ratios, including Ada-KV.}
    \label{tab:ruler_llama31_8b_detail}
\end{table*}

\begin{table*}[t]
    \centering
    \scriptsize
    \setlength{\tabcolsep}{3.2pt}
    \resizebox{\textwidth}{!}{
        \begin{tabular}{llcccccccccccc}
            \toprule
            Method & Ratio & S1 & S2 & S3 & MK1 & MK2 & MK3 & MQ & MV & CWE & FWE & VT & Avg. \\
            \midrule
            FullKV & 100\% & 100.00 & 100.00 & 100.00 & 100.00 & 100.00 & 97.55 & 81.08 & 94.23 & 98.81 & 94.28 & 99.43 & 96.85 \\
            \midrule
            SnapKV & 20\% & 100.00 & 100.00 & 80.63 & 100.00 & 98.25 & 31.49 & 74.24 & 78.74 & 86.59 & 90.21 & 97.70 & 85.26 \\
            Ada-KV & 20\% & 100.00 & 99.76 & 84.51 & 99.65 & 97.20 & 26.10 & 68.89 & 74.90 & 32.07 & 86.82 & 97.23 & 78.83 \\
            Champion-Snap & 20\% & 100.00 & 100.00 & 84.19 & 100.00 & 96.61 & 35.75 & 74.02 & 80.58 & 84.41 & 90.96 & 98.05 & 85.87 \\
            
            SnapKV & 40\% & 100.00 & 100.00 & 95.99 & 100.00 & 98.48 & 49.70 & 81.12 & 92.98 & 96.81 & 93.05 & 99.24 & 91.58 \\
            Ada-KV & 40\% & 100.00 & 100.00 & 96.93 & 99.66 & 97.94 & 32.82 & 80.50 & 91.21 & 62.85 & 92.14 & 98.96 & 86.64 \\
            Champion-Snap & 40\% & 100.00 & 100.00 & 98.29 & 100.00 & 99.33 & 47.52 & 81.37 & 93.85 & 89.12 & 93.10 & 98.50 & 91.01 \\
            
            SnapKV & 60\% & 100.00 & 100.00 & 99.11 & 100.00 & 99.05 & 81.87 & 80.86 & 92.61 & 97.76 & 93.96 & 99.06 & 94.93 \\
            Ada-KV & 60\% & 100.00 & 100.00 & 99.30 & 99.66 & 99.12 & 47.38 & 81.51 & 94.68 & 81.33 & 93.87 & 99.30 & 90.56 \\
            Champion-Snap & 60\% & 100.00 & 100.00 & 99.56 & 100.00 & 98.71 & 87.97 & 80.77 & 92.39 & 96.26 & 94.04 & 99.24 & 95.36 \\
            
            SnapKV & 80\% & 100.00 & 100.00 & 100.00 & 100.00 & 100.00 & 94.28 & 80.88 & 94.01 & 98.53 & 93.36 & 99.24 & 96.39 \\
            Ada-KV & 80\% & 100.00 & 100.00 & 99.91 & 99.66 & 99.60 & 69.03 & 81.29 & 95.70 & 95.09 & 95.46 & 99.35 & 94.10 \\
            Champion-Snap & 80\% & 100.00 & 100.00 & 100.00 & 100.00 & 98.71 & 91.53 & 80.99 & 94.05 & 97.68 & 93.96 & 99.43 & 96.03 \\
            
            \midrule
            PyramidKV & 20\% & 100.00 & 100.00 & 85.18 & 100.00 & 99.69 & 29.17 & 76.87 & 83.52 & 64.29 & 88.06 & 98.10 & 84.08 \\
            Champion-Pyramid & 20\% & 100.00 & 100.00 & 88.89 & 100.00 & 99.67 & 29.63 & 78.24 & 82.11 & 65.13 & 87.45 & 97.92 & 84.46 \\
            
            PyramidKV & 40\% & 100.00 & 100.00 & 96.88 & 100.00 & 99.69 & 47.22 & 81.84 & 92.95 & 84.56 & 92.12 & 98.88 & 90.38 \\
            Champion-Pyramid & 40\% & 100.00 & 100.00 & 95.15 & 100.00 & 97.55 & 49.39 & 79.82 & 91.71 & 85.75 & 90.54 & 99.11 & 89.91 \\
            
            PyramidKV & 60\% & 100.00 & 100.00 & 99.56 & 100.00 & 99.05 & 81.87 & 80.86 & 92.61 & 97.76 & 93.96 & 99.06 & 94.98 \\
            Champion-Pyramid & 60\% & 100.00 & 100.00 & 99.23 & 100.00 & 97.78 & 79.03 & 81.12 & 94.43 & 97.50 & 93.06 & 99.43 & 94.69 \\
            
            PyramidKV & 80\% & 100.00 & 100.00 & 100.00 & 100.00 & 100.00 & 94.28 & 80.88 & 94.01 & 98.53 & 93.96 & 99.24 & 96.45 \\
            Champion-Pyramid & 80\% & 100.00 & 100.00 & 100.00 & 100.00 & 99.08 & 96.34 & 81.08 & 94.71 & 98.71 & 93.66 & 99.43 & 96.64 \\
            \bottomrule
        \end{tabular}
    }
    \caption{Detailed RULER subtask results on Mistral-7B-Instruct-v0.3 under different KV cache retention ratios, including Ada-KV.}
    \label{tab:ruler_mistral_detail}
\end{table*}

\begin{table*}[t]
    \centering
    \scriptsize
    \setlength{\tabcolsep}{3.2pt}
    \resizebox{\textwidth}{!}{
        \begin{tabular}{llcccccccccccc}
            \toprule
            Method & Ratio & S1 & S2 & S3 & MK1 & MK2 & MK3 & MQ & MV & CWE & FWE & VT & Avg. \\
            \midrule
            FullKV & 100\% & 100.00 & 100.00 & 100.00 & 100.00 & 100.00 & 100.00 & 100.00 & 100.00 & 100.00 & 99.70 & 100.00 & 99.97 \\
            \midrule
            SnapKV & 20\% & 100.00 & 100.00 & 100.00 & 100.00 & 100.00 & 44.15 & 100.00 & 100.00 & 99.52 & 97.78 & 100.00 & 94.68 \\
            Ada-KV & 20\% & 100.00 & 100.00 & 98.02 & 100.00 & 100.00 & 46.57 & 99.96 & 99.96 & 99.83 & 97.29 & 99.90 & 94.68 \\
            Champion-Snap & 20\% & 100.00 & 100.00 & 100.00 & 100.00 & 100.00 & 57.96 & 100.00 & 100.00 & 100.00 & 98.15 & 100.00 & 96.01 \\
            
            SnapKV & 40\% & 100.00 & 100.00 & 100.00 & 100.00 & 100.00 & 68.27 & 100.00 & 100.00 & 100.00 & 97.81 & 100.00 & 96.92 \\
            Ada-KV & 40\% & 100.00 & 100.00 & 96.34 & 100.00 & 100.00 & 76.43 & 100.00 & 99.92 & 99.91 & 98.81 & 99.97 & 97.40 \\
            Champion-Snap & 40\% & 100.00 & 100.00 & 100.00 & 100.00 & 100.00 & 95.46 & 100.00 & 100.00 & 100.00 & 99.06 & 100.00 & 99.50 \\
            
            SnapKV & 60\% & 100.00 & 100.00 & 100.00 & 100.00 & 100.00 & 98.33 & 100.00 & 100.00 & 100.00 & 98.77 & 100.00 & 99.74 \\
            Ada-KV & 60\% & 100.00 & 100.00 & 99.47 & 100.00 & 100.00 & 98.38 & 100.00 & 100.00 & 99.94 & 99.58 & 99.97 & 99.76 \\
            Champion-Snap & 60\% & 100.00 & 100.00 & 100.00 & 100.00 & 100.00 & 100.00 & 100.00 & 100.00 & 100.00 & 99.39 & 100.00 & 99.94 \\
            
            SnapKV & 80\% & 100.00 & 100.00 & 100.00 & 100.00 & 100.00 & 100.00 & 100.00 & 100.00 & 100.00 & 99.70 & 100.00 & 99.97 \\
            Ada-KV & 80\% & 100.00 & 100.00 & 100.00 & 100.00 & 100.00 & 100.00 & 100.00 & 99.96 & 100.00 & 99.75 & 100.00 & 99.97 \\
            Champion-Snap & 80\% & 100.00 & 100.00 & 100.00 & 100.00 & 100.00 & 100.00 & 100.00 & 100.00 & 100.00 & 99.70 & 100.00 & 99.97 \\
            
            \midrule
            PyramidKV & 20\% & 100.00 & 100.00 & 94.62 & 100.00 & 100.00 & 39.83 & 100.00 & 100.00 & 98.16 & 98.11 & 100.00 & 93.70 \\
            Champion-Pyramid & 20\% & 100.00 & 100.00 & 97.21 & 100.00 & 100.00 & 50.96 & 99.38 & 99.80 & 99.72 & 96.28 & 100.00 & 94.85 \\
            
            PyramidKV & 40\% & 100.00 & 100.00 & 94.25 & 100.00 & 100.00 & 60.57 & 100.00 & 100.00 & 99.71 & 98.11 & 100.00 & 95.69 \\
            Champion-Pyramid & 40\% & 100.00 & 100.00 & 94.32 & 100.00 & 100.00 & 92.56 & 100.00 & 100.00 & 99.91 & 98.44 & 100.00 & 98.66 \\
            
            PyramidKV & 60\% & 100.00 & 100.00 & 100.00 & 100.00 & 100.00 & 98.02 & 100.00 & 100.00 & 100.00 & 98.77 & 100.00 & 99.71 \\
            Champion-Pyramid & 60\% & 100.00 & 100.00 & 99.24 & 100.00 & 100.00 & 99.78 & 100.00 & 100.00 & 100.00 & 99.37 & 100.00 & 99.85 \\
            
            PyramidKV & 80\% & 100.00 & 100.00 & 100.00 & 100.00 & 100.00 & 100.00 & 100.00 & 100.00 & 100.00 & 99.70 & 100.00 & 99.97 \\
            Champion-Pyramid & 80\% & 100.00 & 100.00 & 100.00 & 100.00 & 100.00 & 100.00 & 100.00 & 100.00 & 100.00 & 99.70 & 100.00 & 99.97 \\
            \bottomrule
        \end{tabular}
    }
    \caption{Detailed RULER subtask results on Llama3.1-70B-Instruct under different KV cache retention ratios, including Ada-KV.}
    \label{tab:ruler_llama31_70b_detail}
\end{table*}

\begin{table*}[t]
    \centering
    \scriptsize
    \setlength{\tabcolsep}{3.2pt}
    \resizebox{\textwidth}{!}{
        \begin{tabular}{llllrrrrrrc}
            \toprule
            Model & Plug-in & Benchmark & Pairs & W & T & L & Win\% & Mean $\Delta$ & Median $\Delta$ & Wilcoxon $p$ \\
            \midrule
            Llama3.1-70B & Champion-Snap & LongBench & 64 & 44 & 12 & 8 & 68.8 & +0.763 & +0.440 & $<0.001$ \\
            Llama3.1-70B & Champion-Pyramid & LongBench & 64 & 39 & 12 & 13 & 60.9 & +0.648 & +0.265 & $<0.001$ \\
            Llama3.1-8B & Champion-Snap & LongBench & 64 & 28 & 0 & 36 & 43.8 & $-0.031$ & $-0.115$ & 0.55 \\
            Llama3.1-8B & Champion-Pyramid & LongBench & 64 & 27 & 5 & 32 & 42.2 & +0.010 & $-0.050$ & 0.48 \\
            Mistral-7B & Champion-Snap & LongBench & 64 & 25 & 6 & 33 & 39.1 & $-0.350$ & $-0.055$ & 0.18 \\
            Mistral-7B & Champion-Pyramid & LongBench & 64 & 25 & 2 & 37 & 39.1 & $-0.177$ & $-0.250$ & 0.11 \\
            Llama3.1-70B & Champion-Snap & RULER & 44 & 7 & 37 & 0 & 15.9 & +1.032 & 0.000 & 0.016 \\
            Llama3.1-70B & Champion-Pyramid & RULER & 44 & 9 & 31 & 4 & 20.5 & +1.064 & 0.000 & 0.17 \\
            Llama3.1-8B & Champion-Snap & RULER & 44 & 18 & 22 & 4 & 40.9 & +0.547 & 0.000 & 0.003 \\
            Llama3.1-8B & Champion-Pyramid & RULER & 44 & 13 & 23 & 8 & 29.5 & +0.200 & 0.000 & 0.20 \\
            Mistral-7B & Champion-Snap & RULER & 44 & 18 & 13 & 13 & 40.9 & +0.026 & 0.000 & 0.79 \\
            Mistral-7B & Champion-Pyramid & RULER & 44 & 15 & 13 & 16 & 34.1 & $-0.045$ & 0.000 & 0.70 \\
            \bottomrule
        \end{tabular}
    }
    \caption{Paired delta statistics for ChampionKV plug-ins. Paired per-unit deltas are Champion minus baseline at (task, budget) granularity for LongBench (16 tasks $\times$ 4 budgets $=64$) and at (subtask, ratio) granularity for RULER ($11\times4=44$). W/T/L count pairs with positive/zero/negative deltas; the two-sided Wilcoxon signed-rank test is over nonzero deltas. Because many RULER subtasks are saturated (ties), the win rate alone understates positive gains; the test uses the sign and magnitude of nonzero deltas.}
    \label{tab:paired_delta_stats_app}
\end{table*}

\clearpage
\section{Appendix: Detailed Information of Benchmarks}
\label{sec:detailed_benchmarks}

To comprehensively evaluate the performance retention capabilities of ChampionKV in long-context inference, we employed two comprehensive benchmarks: LongBench~\cite{bai2024LongBench} and RULER~\cite{hsieh2024ruler}. The following details the specific sub-datasets and task mechanisms used in our experiments.

\subsection{LongBench Benchmark}

LongBench encompasses a variety of real-world long-context application scenarios. In our experiments, we focus on the 16 English datasets used in our evaluation, covering six task domains:

\begin{itemize}
    \item \textbf{Single-Doc QA:} This task requires the model to accurately locate and extract facts from a single long-form source. We selected \textbf{NarrativeQA (NrtvQA)} \cite{kovcisky2018narrativeqa}, which is based on long books and movie scripts; \textbf{Qasper} \cite{dasigi2021dataset}, which is based on academic papers in NLP; and \textbf{MultiFieldQA-en (MF-en)} \cite{bai2024LongBench}, which covers heterogeneous texts such as legal documents and government reports. These tasks heavily rely on the model's ability to deeply parse local long texts.
    
    \item \textbf{Multi-Doc QA:} This task requires the model to establish logical connections across multiple scattered documents and perform multi-hop reasoning. We utilized \textbf{HotpotQA} \cite{yang2018hotpotqa}, \textbf{2WikiMultihopQA (2WikiMQA)} \cite{ho2020constructing}, and \textbf{Musique} \cite{trivedi2022musique}. The model must integrate fragmented clues across paragraphs within a long context to deduce the final answer.
    
    \item \textbf{Summarization:} This task evaluates the model's comprehensive understanding and information compression capabilities over global long contexts. The test set includes \textbf{GovReport} \cite{huang2021efficient}, which focuses on the summarization of long government reports; \textbf{QMSum} \cite{zhong2021qmsum}, which focuses on query-based summarization of meeting transcripts; and \textbf{MultiNews} \cite{fabbri2019multi}, which focuses on comprehensive summarization of multiple news reports.
    
    \item \textbf{Few-shot Learning:} This task assesses the model's in-context learning capabilities using prior examples within a long context. We included \textbf{TREC} \cite{li2002learning} for text classification, \textbf{TriviaQA} \cite{joshi2017triviaqa} for reading comprehension, and \textbf{SAMSum} \cite{gliwa2019samsum} for daily dialogue summarization. The model must perform accurate pattern matching based on multiple few-shot examples provided within the context.
    
   \item \textbf{Synthetic Tasks:} These tasks are designed to precisely measure long-text localization capabilities through artificially constructed patterns. Two synthetic tasks, \textbf{PassageCount (PCount)} and \textbf{PassageRetrieval-en (PRe)}, are included in LongBench \cite{bai2024LongBench}. PassageCount identifies and counts the total number of unique paragraphs after deduplication, while PassageRetrieval-en requires the model to retrieve the original paragraph from a set of Wikipedia paragraphs based on a given summary.
    
    \item \textbf{Code Completion:} This task evaluates the model's semantic understanding in cross-file or extremely long codebase environments. The \textbf{Lcc} dataset is derived from the original Long Code Completion dataset \cite{guo2023longcoder}, while \textbf{RepoBench-P (RB-P)} \cite{liu2024repobench} is designed for repository-level code auto-completion and requires aggregating information from code across files. The model is required to predict the next line of code.
\end{itemize}

\subsection{RULER Benchmark}

Compared to the foundational Needle-in-a-Haystack (NIAH) test, RULER~\cite{hsieh2024ruler} provides a more challenging and flexibly configured evaluation for long-context information retrieval and aggregation. The RULER experiments in this paper specifically include the following 11 subtasks:

\begin{itemize}
    \item \textbf{Single NIAH (S1, S2, S3):} Basic single-target information retrieval tasks. The model is required to accurately locate and extract a uniquely hidden piece of key information, namely the Needle, from an extremely long context composed of a large amount of irrelevant background text, namely the Haystack.
    
    \item \textbf{Multi-Keys NIAH (MK1, MK2, MK3):} Advanced retrieval tasks. Multiple different Needles are embedded in the context, but the query only asks the model to retrieve a specific one. This task examines the model's precise addressing capability in the presence of multiple strong distractors.

    \item \textbf{Multi-Values / Multi-Queries NIAH (MQ, MV):} Multiple related key-value pairs are scattered throughout the context. The MV task requires the model to retrieve and aggregate all values corresponding to a specific key, while the MQ task requires the model to simultaneously answer multiple given queries. Both tasks heavily test the model's ability to capture and integrate multi-point information within long texts.

    \item \textbf{Word Extraction (CWE, FWE):} This category includes Common Words Extraction (CWE) and Frequent Words Extraction (FWE). The model must find the core vocabulary with the highest frequency of occurrence in an extremely long vocabulary sequence generated by a specific statistical distribution, testing the model's awareness of global statistical distributions in long-context information.
       
    \item \textbf{Variable Tracking (VT):} Multiple complex variable binding statements, such as $X_2 = X_1$ and $X_3 = X_2$, are inserted into the long text. The model needs to track the assignment chain and return all variable names that ultimately point to the same value, greatly challenging long-distance logical chain tracking capabilities.
    
\end{itemize}

\end{document}